\documentclass[12pt]{report}
\usepackage{amsmath}
\usepackage{amsfonts}
\usepackage{amssymb}
\usepackage{breqn}
\usepackage{listings}
\usepackage[T1]{fontenc}
\usepackage[
    a4paper,
    margin=3cm,
    bindingoffset=1cm,
    lines=25,
    includeheadfoot
]{geometry}
\usepackage{graphicx}
\usepackage{booktabs}
\usepackage{pifont}
\usepackage{microtype}
\usepackage{textcomp}
\usepackage{setspace}
\usepackage[binary-units,per-mode=symbol]{siunitx}
\usepackage[english]{babel}
\usepackage[capitalise]{cleveref}
\crefname{section}{Sect.}{Sect.}
\crefname{figure}{Fig.}{Fig.}
\crefname{table}{Tab.}{Tab.}
\crefname{equation}{Eq.}{Eq.}
\Crefname{table}{Table}{Tables}
\Crefname{equation}{Equation}{Equations}
\crefname{equation}{Equation}{Equations}
\def\figname{\csname cref@figure@name\endcsname\xspace}
\def\tabname{\csname cref@table@name\endcsname\xspace}
\def\secname{\csname cref@section@name\endcsname\xspace}
\def\eqname{\csname cref@equation@name\endcsname\xspace}
\def\eqpname{\csname cref@equation@name@plural\endcsname\xspace}
\usepackage{xspace}
\usepackage{makecell}
\usepackage{hyphenat}
\usepackage{listings}
\usepackage{hyphenat}
\usepackage{multirow}
\usepackage{float}
\usepackage{verbatim}
\usepackage[ragged,flushmargin]{footmisc}
\usepackage{csquotes}

\usepackage{todonotes}
\usepackage{xcolor} 
\definecolor{light-gray}{gray}{0.97} 
\definecolor{codegreen}{rgb}{0,0.6,0}

\usepackage{todonotes}
\presetkeys{todonotes}{inline}{}

\definecolor{codebg}{rgb}{0.95,0.95,0.95}
\definecolor{rulecolor}{rgb}{0.5,0.5,0.5}
\usepackage[backend=biber,style=ieee,doi=false,isbn=false,mincitenames=1,maxcitenames=4]{biblatex}

\renewcommand{\thead}[1]{\ensuremath{H}\xspace}

\usepackage{xstring}
\usepackage{xparse}
\usepackage{acro}  
\NewDocumentCommand\acrodef{mO{#1}mG{}}{\DeclareAcronym{#1}{short={#2}, long={#3}, #4}}

\acrodef{3GPP}{3rd Generation Partnership Project}
\acrodef{4G}{4$^{\text{th}}$ Generation}
\acrodef{5G}{5$^{\text{th}}$ Generation}
\acrodef{5GV2X}[5G-V2X]{5$^{\text{th}}$ Generation Cellular V2X}
\acrodef{ACI}{Adjacent Channel Interference}
\acrodef{ACC}{Adaptive Cruise Control}
\acrodef{AGC}{Automatic Gain Control}
\acrodef{AI}[AI]{Artificial Intelligence}
\acrodef{AoI}{Age of Information}
\acrodef{API}{Application Programming Interface}
\acrodef{AWGN}{Additive White Gaussian Noise}
\acrodef{BER}{Bit Error Probability}
\acrodef{CACC}{Cooperative Adaptive Cruise Control}
\acrodef{CITS}{Cooperative Intelligent Transport Systems}
\acrodef{CSI}{Channel State Information}
\acrodef{CTL}{ Certificate Trust Lis}
\acrodef{CRL}{Certificate Revocation List}
\acrodef{CSMA}{Carrier-Sense Multiple Access}
\acrodef{CV2X}[C-V2X]{Cellular V2X}
\acrodef{D2D}{Device-to-Device}
\acrodef{DoS}{Denial of Service}
\acrodef{DES}{Discrete Event Simulation}
\acrodef{eNB}{eNodeB}
\acrodef{ETSI}{European Telecommunications Standards Institute}
\acrodef{FCC}{Federal Communications Commission}
\acrodef{FDMA}{Frequency-Division Multiple Access}
\acrodef{FSM}{Finite State Machine}
\acrodef{gNB}{gNodeB}
\acrodef{GUI}{Graphical User Interface}
\acrodef{GPS}{Global Position System}
\acrodef{HARQ}{Hybrid Automatic Repeat reQuest}
\acrodef{HiL}{Hardware-in-Loop}
\acrodef{IBC}{Identity-based Cryptography}
\acrodef{IC}{In Coverage}
\acrodef{IDS}{Intrusion Detection System}
\acrodef{ITS}{Intelligent Transport Systems}
\acrodef{IPsec}{IP Security}
\acrodef{IEEE}{Institute of Electrical and Electronics Engineers}
\acrodef{k-NN}{k-Nearest Neighbors}
\acrodef{JCAS}{Joint Communication and Sensing}
\acrodef{WLAN}{Wireless Local Area Network}
\acrodef{LOS}[LoS]{Line of Sight}
\acrodef{LSTM}{Long Short-Term Memory}
\acrodef{LTE}{Long Term Evolution}
\acrodef{LTEV2X}[LTE-V2X]{Cellular V2X}
\acrodef{LuST}{Luxembourg SUMO Traffic}
\acrodef{MAC}{Medium Access Control}
\acrodef{MDS}{Misbehavior Detection System}
\acrodef{MEC}{Mobile Edge Computing}
\acrodef{MI}{Mutual Information}
\acrodef{MIMO}{Multiple Input Multiple Output}
\acrodef{ML}[ML]{Machine Learning}
\acrodef{MLO}{Multi-Link Operation}
\acrodef{MUMIMO}[MU-MIMO]{Multi-User \ac{MIMO}}
\acrodef{MPC}{Model Predictive Control}{short-indefinite={an}}
\acrodef{NN}{Neural Network}
\acrodef{NR}{New Radio}
\acrodef{OFDM}{Orthogonal Frequency-Division Multiplexing}
\acrodef{OFDMA}{Orthogonal Frequency-Division Multiple Access}
\acrodef{OoC}{Out of Coverage}
\acrodef{PC5}{PC5}
\acrodef{PDR}{Packet Delivery Ratio}
\acrodef{PDF}{Probability Density Function}
\acrodef{PHY}{Physical Layer}
\acrodef{PKI}{Public Key Infrastructure}
\acrodef{PMF}{Probability Mass Function}
\acrodef{PSK}{Phase Shift Keying}
\acrodef{QAM}{Quadrature Amplitude Modulation}
\acrodef{QOS}[QoS]{Quality of Service}
\acrodef{QPSK}{Quadrature Phase Shift Keying}
\acrodef{RNN}{Recurrent Neural Network}
\acrodef{RU}{Resource Unit}
\acrodef{RL}{Reinforcement Learning}
\acrodef{FL}{Federated Learning}
\acrodef{RSU}{Road Side Unit}
\acrodef{SAE}{Society of Automotive Engineering}
\acrodef{SBSPS}{Sensing-Based Semi-Persistent Scheduling}
\acrodef{SC-FDMA}{Single Carrier-Frequency Division Multiple Access}
\acrodef{SINR}{Signal to Interference plus Noise Ratio}{short-indefinite={an}}
\acrodef{SIR}{Signal to Interference Ratio}
\acrodef{SNR}{Signal to Noise Ratio}
\acrodef{SUMO}{Simulation of Urban Mobility}
\acrodef{STA}{Station}
\acrodef{SVM}{Support Vector Machine}
\acrodef{TDMA}{Time-Division Multiple Access}
\acrodef{UE}{User Equipment}
\acrodef{URLLC}{Ultra-Reliable Low-Latency Communication}  
\acrodef{Uu}{User to Network Interface}
\acrodef{V2V}{Vehicle to Vehicle}
\acrodef{V2X}{Vehicle to Everything}
\acrodef{V2I}{Vehicle to Infrastructure}
\acrodef{VANET}{Vehicular ad-hoc Network}
\acrodef{VeReMi}{Vehicular Reference Misbehavior}
\acrodef{VN}{Vehicular Network}
\acrodef{VLC}{Visible Light Communication}
\acrodef{VRU}{Vulnerable Road Users}
\acrodef{WHD}{Weighted Hamming Distance}
\acrodef{ZOH}{Zero Order Hold}
\acrodef{VAE}{Variational Autoencoder}
\acrodef{GMM}{Gaussian Mixture Model}
\acrodef{OBU}{On-board Unit}
\acrodef{TraCI}{Traffic Control Interface}

\DeclareFieldFormat{sentencecase}{#1} 
\DeclareFieldFormat{titlecase}{#1} 
\DeclareFieldFormat{titlecase}{#1} 
\usepackage{xpatch}
\xpatchbibmacro{textcite}{\addspace}{\addnbspace}{}{}
\xpatchbibmacro{Textcite}{\addspace}{\addnbspace}{}{}
\DefineBibliographyStrings{english}{
	andothers = et~al\adddot\addspace
}

\graphicspath{{./images/}} 

\begin{document}
    \newgeometry{left=20mm,right=20mm,top=25mm,bottom=20mm}
\begin{titlepage}
    \begin{center}
        
        \begin{figure}[t]
            \centering
            \includegraphics[width=72.4mm,height=30mm]{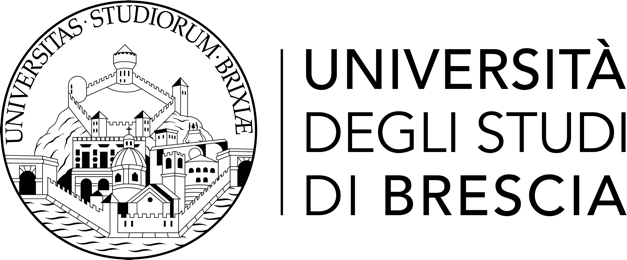}    
        \end{figure}

        \vspace*{10mm}
        
        {\fontsize{17}{17}\fontfamily{lmss}\selectfont
            DIPARTIMENTO DI INGEGNERIA DELL'INFORMAZIONE\\
        }

        \vspace*{10mm}

        {\fontsize{17}{17}\fontfamily{cmss}\selectfont
            Corso di Laurea Magistrale\\
            in Ingegneria Informatica\\

        }    
        
        \vspace*{20mm}

        {\fontsize{20}{20}\fontfamily{lmss}\selectfont 
            Tesi di Laurea\\[3mm]
            \textbf{A neural-network based anomaly detection system and a safety protocol to protect vehicular network} \\[5mm]
            \textbf{
            Sistema di rilevamento delle anomalie basato su rete neurale e protocollo di sicurezza per proteggere la rete veicolare} \\[3mm]
        }

    \end{center}

    \vfill

    \begin{flushleft}
        {\fontsize{17}{17}\fontfamily{lmss}\selectfont 
            \textbf{Relatori:} Prof. Renato Lo Cigno \\ 
            ~~~~~~~~~~~~~ Dott. Lorenzo Ghiro
        }
        
    \end{flushleft}

    \vspace*{5mm}

    \begin{flushright}
        {\fontsize{17}{17}\fontfamily{lmss}\selectfont 
            Laureando:\\
            Marco Franceschini\\
            Matricola n. 727808\\
        }

    \end{flushright}

    \vspace*{5mm}

    \rule{0.95\textwidth}{0.4pt}
    \begin{center}
    {\fontsize{17}{17}\fontfamily{lmss}\selectfont 
        Anno Accademico 2023/2024
    }
    \end{center}

\end{titlepage}
\restoregeometry

    \setstretch{1.1}

    \section*{Abstract}

The increasing need of road efficiency, driving safety, and sustainable transport relies on \ac{CITS} developments in modern transportation, with a stress on the C o Communications. 
By enabling vehicles to communicate with each other, exchanging real-time data through an ad-hoc network, \ac{CITS} can improve road safety and efficiency. Some of the challenging tasks in vehicular networks include the protection and the guarantee of correctness of the exchanged data. The common practice to deal with those tasks involves utilizing a Public Key Infrastructure  to authorize vehicles using the network. Despite this, it doesn't mean that an authorized vehicle cannot misbehave. To achieve driving safety, a standard solution is using a Misbehavior Detection System that can detect malfunctions leading to wrong, missing, or misleading messages in the network by analyzing the data exchanged in every message. 

This thesis presents a simple Machine Learning framework based on Long-short Term Memory neural networks, a standard approach in the referring literature for misbehavior detection. The detection system must be trained off-line, as usually done, on a standard dataset, and for this purpose, the VeReMi dataset has been chosen. 
Many works focus only on the off-line model, trying to reach the best scores on the whole dataset. This thesis aims to introduce the simple detection framework at run-time on a completely different scenario, compared to the one on which the dataset was built, suitable to be executed on single vehicles while they communicate. 
The detector is validated within a platooning application, showing that, if properly used, the predictions can prevent practically all accidents caused by misbehavior. 
In particular, the detector's predictions are provided to a simple defense protocol that dismantles the platoon if a message is detected to be anomalous. 

The results show that the detection system and the defense protocol can save almost all the accidents due to misbehavior conditions in the simulated set. In particular, detecting a general misbehavior condition is precisely predicted, with an excellent accuracy score. Although those results, the detection system cannot identify each kind of misbehavior by assigning a label to each one, with reasonable accuracy results, due to very different traffic conditions while running on-line. This suggests that designing a more complex defense protocol with personalized responses for each misbehavior, usable in all traffic conditions, is impossible, even with more sophisticated Artificial Intelligence solutions. Despite this, the developed system gives a solid basis to work on it, and with millions of data and more misbehavior while training, it may be introduced into real-world vehicles as a \ac{CITS} technology, enhancing driving safety and making cooperative driving systems safer.

\newpage
\section*{Sommario}
\acresetall
La crescente esigenza di efficienza, sicurezza stradale e trasporto sostenibile, insieme ai problemi di congestione del traffico e urbanizzazione, trova una possibile soluzione nei recenti sviluppi delle tecnologie \ac{CITS}, con particolare riferimento alla C di Comunicazioni. Attraverso la comunicazione tra veicoli, scambiando dati in tempo reale tramite una rete veicolare ad-hoc, le tecnologie \ac{CITS} possono migliorare la sicurezza e l'efficienza stradale. Alcune delle principali sfide nell'utilizzo di una rete ad-hoc includono la protezione e la garanzia della correttezza dei dati scambiati. La pratica standard per gestire tali problemi prevede l'utilizzo di una Public Key Infrastructure per autorizzare i veicoli consentendogli di accedere alla rete. Nonostante ciò, non vuol dire che un veicolo autorizzato non possa trasmettere messaggi anomali. Per garantire la totale sicurezza di guida anche da questo punto di vista, la soluzione standard è l'utilizzo di un Misbehavior Detection System in grado di rilevare malfunzionamenti che portano a messaggi errati o mancanti analizzando i dati scambiati in ogni messaggio trasmesso.

Questa tesi presenta un semplice framework di Machine Learning basato su una rete neurale di tipo
Long-short Term Memory (LSTM), un approccio standard
nella letteratura di riferimento per i sistemi di rilevamento di malfunzionamenti. Il sistema, basato su rete neurale, dev'essere addestrato
off-line, come di solito avviene, su un dataset di riferimento, a questo scopo, è stato scelto il dataset VeReMi. 
Molti lavori di ricerca si focalizzano solo sul modello off-line, cercando di raggiungere risultati ottimi esclusivamente sul dataset. Questa tesi mira a introdurre il semplice  framework sviluppato a run-time su
uno scenario completamente diverso, rispetto a quello su cui è stato costruito il dataset,
adatto per essere eseguito su ogni singolo veicolo mentre comunica con i vicini.
Il sistema è convalidato da un'applicazione di platooning, dimostrando che, se
utilizzate correttamente, le previsioni del sistema possono prevenire praticamente tutti gli incidenti causati da
messaggi anomali o malfunzionamenti. In particolare, le previsioni del sistema vengono fornite a un semplice protocollo di difesa che smantella il platoon se un messaggio viene rilevato come anomalo.

I risultati mostrano che il framework e il protocollo di difesa possano salvare quasi
tutti gli incidenti dovuti a condizioni di misbehavior nel set simulato. In particolare, il rilevamento di una condizione di misbehavior generale è predetta con precisione, con un'accuratezza eccellente. Nonostante questi risultati, il sistema non è in grado identificare ogni tipo di misbehavior assegnando una label a ciascuno, con risultati di accuratezza ragionevoli, a causa di condizioni di traffico molto diverse durante l'esecuzione
on-line. Ciò suggerisce che progettare un protocollo di difesa più complesso con
risposte personalizzate per ogni tipo di misbehavior, utilizzabile in tutte le condizioni di traffico, è impossibile, anche con soluzioni di intelligenza artificiale più sofisticate di quella progettata. Nonostante ciò, il sistema sviluppato fornisce una solida base su cui lavorare e, con l'introduzione di milioni di dati e di più tipologie di misbehavior durante la fase di training, potrebbe essere installato nei veicoli del mondo reale come tecnologia \ac{CITS}, migliorando la sicurezza e rendendo i sistemi di guida cooperativa più sicuri.

    \tableofcontents
    
    \acresetall
\chapter{Introduction}
\label{ch:intro}
\ac{CITS} represent a significant advance in modern transportation, potentially transforming how vehicles and infrastructure interact on the road \cite{sjoberg2017cooperative}. 
As urbanization and traffic congestion continue to increase, the necessity of creating more efficient, safe, and sustainable transport systems has never been greater.
The significance of \ac{CITS} lies in its ability to improve road safety,  traffic congestion, and decrease emissions. 
By enabling vehicles to communicate with each other and with infrastructures, \ac{CITS} applications can help prevent accidents by sending timely warnings, optimizing traffic flow through smart intersection management, and reducing fuel consumption through features such as platooning\footnote{A group of vehicles travels closely at high speeds, with the lead vehicle controlling the speed and direction. This reduces fuel consumption, increases road capacity, and improves safety by minimizing human error.}, eco-driving support\footnote{Vehicles receive real-time information about traffic conditions, speed limits, and road gradients to optimize fuel efficiency and reduce emissions.},  intersection management\footnote{Vehicles and traffic lights communicate to optimize traffic flow, reducing wait times and collisions.}, emergency vehicle preemption\footnote{Emergency vehicles can communicate with traffic lights to request priority at intersections, but also directly with other vehicles coordinating their movements, reducing response times and enhancing safety for both the emergency vehicle and other road users.}, and many others. 
Studies \cite{eu2020pedestrians, eu2022cleanair, eu2024roadfatalities} suggest that the implementation of \ac{CITS} could result in a 50\% reduction in road fatalities and a 20\% decrease in fuel consumption and greenhouse gas emissions. 
Indeed, given the prevalence of human errors in vehicular accidents, it is not difficult to predict that the road accidents reduction can even approach 100\%, as it is today in well managed railways. 
With their offer of benefits in safety, efficiency, and sustainability, \ac{CITS} are a cornerstone of the future of transportation. 
As these technologies continue to develop, they promise to reshape our transportation networks and significantly improve the quality of life in urban areas and beyond.

Road users and traffic management systems need to exchange real-time data to achieve \ac{CITS}'s goals of enhancing driving safety and efficiency of vehicles, which requires efficient and secure wireless networks. 
\ac{VANET}, \ac{V2V}, \ac{V2I}, \ac{V2X} are all keywords pointing the the same concept: \ac{CITS} is enabled by communication technologies. 
Only a high level of collaboration can ensure a dynamic and responsive transportation network that can be adapted to change road conditions. 
Vehicles in a \ac{VANET} are equipped with a wireless interface, creating a dynamic network that can be accessed without additional overhead. 
This network can also include infrastructure components called \ac{RSU}, which are sparsely positioned along the road. 
The resulting network, which includes sparse infrastructure, is known as a \ac{VN}. 
A form of \ac{VN} can also be based on cellular technologies (\ac{CV2X}), which privileges the infrastructure over direct communications; however, the key concepts remain the same and many situations require in any case direct, ad-hoc, on-demand communications. 
In this ad-hoc network, which is highly dynamic and heterogeneous, there are some challenging tasks to protect the integrity of the data and guarantee its correctness. 
For instance, in \ac{CITS}, the typical approach involves utilizing \ac{PKI} to exclusively provide key material and certificates to authorized vehicles and entities. 
Unauthorized entities are barred from the system as their messages lack valid signatures. The state of the art of \ac{PKI} for \ac{CITS} is discussed in detail
in \cref{s:securityreview}. 
However, if a vehicle possesses valid key material and starts to misbehave inside the \ac{VN}, this approach cannot defend against it. 
The word \textit{misbehavior} doesn't include only the attackers that intentionally corrupt communications and information, but also cars that produce anomalies due to genuine errors, e.g., a \ac{GPS} malfunction, that can bring to a misbehavior inside the \ac{VN}. 
A \ac{MDS} \cite{van2018survey} can be used to address the work of recognizing the misbehavior of authorized vehicles.

An \ac{MDS} in a \ac{VN} is designed to identify actions that compromise the integrity and safety of data exchanges within the network. 
Unlike traditional \acp{IDS} in IT, which focus on identifying unauthorized access or disruptions to the network, the \acp{MDS} analyze the behavior and the content of the messages exchanged between vehicles and infrastructure. These systems look for inconsistencies or signs of malicious intent, trying to identify attackers or anomaly vehicles, which could cause dangerous situations on the road. 
This approach is particularly suited to the dynamic nature of \ac{VN}.

\section{Thesis Goals} 
\label{s:goal}
To achieve \ac{CITS}’s objectives, with a specific emphasis on driving safety, a cryptographic defense based on \ac{PKI} is not sufficient. It also requires a system capable of identifying malfunctions in authorized vehicles.
The literature review about those \acp{MDS} is presented in detail in \cref{s:misbeh_literature}. 
Analyzing the works made about those kinds of systems, it emerges that the researchers are moving to \ac{AI} based solutions, focusing in particular on \ac{NN} models. 
To train an \ac{MDS} based on an \ac{AI} solution, a dataset containing misbehavior data and the exchanged messages in the \ac{VN} is mandatory; the larger and more comprehensive the dataset is, the better. 
Then one can imagine solutions that keep learning from real traffic, but an initial base is mandatory to create a reliable system that analyze the exchange of messages  in the real world to identify anomalies. 
Since many different models have been developed over the years, maintaining a standard dataset is a good practice to compare the results with the previous works and to increase the complexity of the same dataset while introducing new attacks and data. 
The \ac{VeReMi} dataset \cite{van2018veremi, kamel2020veremi}, a standard reference for evaluating misbehavior detection mechanisms for \acp{VANET} in many different works, is used for this thesis, and it is indeed one of the very few dataset available for research in this area. 
The original dataset includes several simple attacks. 
The purpose of this dataset is not only to establish a basis for comparing detection methods but also to act as a foundation for developing more complex attacks and find the relative countermeasures.  

The literature analysis reveals that many works focus only on creating a good \ac{MDS} trained on the whole \ac{VeReMi} dataset, but they evaluate it only offline, using for the testing part of the same \ac{VeReMi} dataset, failing to evaluate the \ac{MDS} on-line\footnote{The term on-line will be used through all the thesis to refer to vehicles that are normally running into the traffic, e.g., on-line detection means a detection computed for every single vehicle that is running in a real-world scenario.}, while the vehicles are running, exchange messages and behave on the road based on the information they receive.
Obviously, when we talk about on-line testing we refer to a simulation environment, as experiments with real vehicles are still impossible in this field: indeed, also the \ac{VeReMi} dataset is obtained with extended simulation experiments. 
The idea of this thesis is to create a simple \ac{MDS} that is \ac{NN}-based but still trained on the standardized \ac{VeReMi} dataset while selecting some attacks that can be easily replicated in a real-world scenario. 
The core idea is to recreate the misbehavior, based on those present in the \ac{VeReMi} dataset and used to train the model, in a completely different scenario compared to the Luxembourg City used to create the dataset and then make the \ac{MDS} work on-line installed on every single car, while the vehicles are moving through the traffic and exchanging messages.

Since the majority of the works focus only on the offline training, there's not only a lack in on-line evaluations of the trained \ac{MDS}, but also a lack in a consequent usage of the predictions of the \ac{MDS} to try to save as much vehicles as possible. A protocol that can coordinates the prediction of a vehicle, or a group of that, is necessary if the objective is to use the detection system in the real life, and also to evaluate how many accidents is able to save.
This is why, along with the \ac{MDS}, the idea is to develop a protocol that can work based on the predictions of the detection system, trying to save the vehicles exposed to the misbehaving one by properly changing their behavior and, eventually, return to autonomous driving (i.e., based only on local sensors) or human driving. 

Given that a good detection system that could be implemented in the vehicle industry has to work under every condition, the thesis goal is to make the \ac{MDS} work on every single vehicle through the road in a not centralized way, in significantly different scenarios, e.g., urban or highway, evaluate its accuracy in detect misbehavior, and with the introduction of the protocol, show how many vehicles could be saved from an accident, improving the driving safety. 
To evaluate the system on-line, a simulator tool is used and presented in \cref{ch:simuenv}, where the \ac{NN} is included in the vehicles communication subsystem and evaluate messages as they arrive to the vehicle, thus on-line. 
In particular, the simulators used include OMNeT++, a discrete event simulator used to emulate the \ac{VN} in which the messages are exchanged, and \ac{SUMO}, a microscopic mobility simulator that is used to emulate the vehicles dynamics in traffic, urban and not.

\section{Thesis Structure}
\label{s:structure}

After analyzing the state of the art about this topic in \cref{ch:related}, the first step to reach the goals defined is the definition of an \ac{MDS} with an \ac{NN} model, trained over the \ac{VeReMi} dataset properly adapted and manipulated to fit the network's input. 
Some \ac{ML} metrics, like accuracy, precision, and recall, are used to evaluate the offline training and validation of the model. The definition, implementation and evaluation of the \ac{MDS} are discussed in \cref{ch:misbeh}.
After that, the simulation's instruments are presented in detail in \cref{ch:simuenv}. While, the innovative idea concerns the validation of the developed \ac{MDS} at run-time, going ahead compared to the simple training and testing on the same dataset. The modification of the simulation instruments, allowing the usage of an \ac{NN} at run-time including the prediction on the incoming messages, is discussed in the \cref{ch:method}. Always in \cref{ch:method} the focus is moved on the implementation of the misbehavior selected through the dataset, into the simulation structure at run-time, and on the design and development of the innovative defense protocol that aims to work on the prediction made by the \ac{MDS}. In the end, the set of simulations defined in \cref{ch:sim} is executed, with all the parameters used explained, trying to validate the framework presented on-line, a test performed only a few times in the literature.
The objective of the simulation's results is evaluate they with some metrics to understand if the \ac{MDS} can be used on every vehicle, how it can be improved and against how many misbehavior it results to be effective along with the defense protocol. 

Based on the accuracy and the other metrics of the detection between the messages that are misbehavior or not, an on-line evaluation of the \ac{MDS} performances is made in \cref{s:res_s}, for what concerns the detection of a general misbehavior, defined as \textit{single} label\footnote{The word single label is used along all the thesis to define that from the prediction results, i.e., the label prediction between 0 (regular) and 8 (dataReplay), it only matters when the predicted label is 0 or different from it. This brings to a definition of only one single label that contains all the misbehavior.}, and in \cref{s:res_m} for what concerns the punctual identification of each kind of misbehavior studied, defined as \textit{multiple} label\footnote{The word multiple label is used with the meaning of a correct identification of all the misbehavior labels, i.e., from 1 (contsPos) to 8 (dataReplay), defining multiple labels for misbehavior.}. A high accuracy is needed to reach a positive result, particularly between misbehavior and not detection, i.e., the single label. In a real-world scenario, at least an accuracy between three and five nines (99.999\%) is required, but considering that the dataset has only a few entries, compared to the millions of data needed to reach that value, the results are evaluated from this perspective. Another important value is how many false positives are predicted instead of not misbehavior messages. This is very important if it's considered that the protocol is applied on the first message that is detected as misbehavior, and having a model that predicts many false anomalies could bring to a continued stop of the exchanging of the messages through the \ac{VN}.
In addition to the predictions accuracy and a false positive metric, a gain that represents how many accidents are saved with the developed detection system is presented in \cref{s:gain_acc}. This value has to be as high as possible, close to 100\%, to be reliable in the real world.

    \chapter{Related work}
\label{ch:related}

This literature review chapter aims to provide a comprehensive overview of the existing research and developments in security and privacy within \ac{VN}, as well as the application of \ac{AI} techniques for \ac{MDS}. The chapter is divided into two main sections. The first section reviews the state-of-the-art in security and privacy management for \acp{VN}, covering standardization efforts, secure communication protocols, privacy-preserving techniques, and challenges associated with the deployment of secure vehicular communication systems. The second section focuses on the literature surrounding \acp{MDS}, particularly those utilizing \ac{AI} and \ac{ML} methodologies, which are pivotal for identifying and mitigating malicious activities and ensuring the robustness of \ac{VN}. By synthesizing the findings from these domains, this chapter lays a solid foundation for understanding the current landscape of \ac{VN} security and misbehavior detection. It also identifies gaps in the existing research, thereby setting the stage for developing advanced protocols and systems to enhance the security and reliability of future \acp{VN}.

\section{Security and Privacy in Vehicular Networks}
\label{s:securityreview}

\subsection{Standardization}
\label{s:standard}
The first step in analyzing the state-of-the-art about security and privacy in \acp{VN} is to examine the current status of the standardization process.

Various standards have been already established related to this topic, particularly by the \ac{ETSI}. The recent document \cite{ETSITS103601V211} (V2.1.1 – March 2024) details the requirements
and the protocols for the distribution of security credentials in the \ac{ITS} framework. The specifications are technology-agnostic
(i.e., they to apply both to \ac{ITS}-G5 and \ac{CV2X}) and refer to several different
communications profiles. 
The most important definitions in this document involve:
enrollment credentials and authorization tickets for vehicles and \ac{RSU} to ensure the authenticity and integrity of the communicating entities; a \ac{CTL} distribution service that helps ITS stations to verify the trustworthiness of certificates and, similarly, the \ac{CRL} distribution service.
Security protocols are also defined to manage credentials, including mechanisms for handling message retries and errors effectively.
While, the \ac{ETSI} standardized document for trust and privacy management \cite{ETSITS102941V221} (V2.1.1
- November 2022) is mainly focused on security, trust, and privacy management
for \ac{ITS}. This standard covers a broad range
of protocols and procedures to ensure that vehicular communications systems
\ac{V2X} are secure, and users’ privacy is protected. 
In particular, a key component is the implementation of a \ac{PKI} and privacy enhancement technologies, such as certified pseudonyms and secure credential management, to minimize personal data transmission and ensure user privacy. Data security mechanisms employ encryption, decryption, digital signatures, and secure broadcasting to prevent unauthorized data access and to ensure message integrity, and they also employ a secure architecture with well-defined security layers, policies, and procedures, including the use of \ac{IPsec} for network layer security. Threat and risk management involves identifying potential threats like eavesdropping, identity theft, tampering, and \ac{DoS} attacks, along with risk assessment procedures and countermeasures.
Also, the standard \cite{ETSITS102940V111} (V1.1.1 - June 2012) and its following release \cite{ETSITS102940V211} (V2.1.1 - July 2021) are focused on secure layered architecture and threats,
basically describing something very similar to the previously presented standard.
Finally, the \ac{ETSI}'s standards \cite{ETSITS103097V211} provide specifications for secure data structures used in ITS, detailing security headers and certificate formats essential for maintaining data integrity and security within \acp{VN}.

\subsection{Privacy management}
Privacy management in \acp{VN} involves strategies and technologies designed to protect the identity and location information of vehicles and their occupants while ensuring secure communication. After the standardization, significant advancements have been made to enhance privacy.
The key strategy is pseudonymity \cite{petit2015pseudonym}, where vehicles use temporary identifiers that do not reveal the real identity of the vehicle's owner. Vehicles obtain pseudonyms from a trusted authority and use them to sign messages within the network. These pseudonyms are changed periodically to prevent tracking. This method employs various cryptographic techniques, including \ac{PKI}, \ac{IBC}, and group signatures. \ac{PKI} manages public key encryption and digital signatures, \ac{IBC} reduces overhead by using the vehicle's identifier as its public key, and group signatures allow vehicles to sign messages as members of a group without revealing their specific identities, ensuring strong anonymity.

Following the \ac{ETSI} standardization \cite{ETSI_TR_103415} (V1.1.1 - April 2018), several pseudonym change strategies have been standardized and implemented to enhance privacy. These strategies include fixed intervals, adding randomness to intervals, silent periods after pseudonym changes, vehicle-centric changes based on behavior, density-based changes in high-density areas, mix-zones where all vehicles change pseudonyms simultaneously, collaborative changes where vehicles synchronize pseudonym changes, cryptographic mix-zones, and pseudonym swaps between vehicles. These strategies are evaluated based on metrics such as privacy, user-centric performance, safety, and cost.
Research \cite{forster2017evaluation} has demonstrated that effective pseudonym change strategies significantly enhance privacy, particularly in urban scenarios. In highway scenarios, the effectiveness is somewhat reduced due to more predictable vehicle movements. To maintain service continuity and quality, it has been suggested to announce pseudonym changes in some studies like \cite{feiri2015changes}, facilitating collaborative pseudonym changes and increasing overall system privacy.

Introducing the \ac{5G} technologies has further improved privacy management \cite{basta20225g}. Enhanced location tracking precision and improved clock synchronization allow for more effective pseudonym changes, coinciding with specific events or zones to enhance privacy without disrupting communication. The \ac{5G} service-based architecture modularizes network functions, enabling security measures and privacy settings in different network slices, thus enhancing overall privacy management.
These advancements post-standardization represent a significant step forward in managing privacy in \ac{VN}. They ensure that user data remains secure while maintaining the integrity and efficiency of the communication systems.

\subsection{Studied attacks on Cooperative Adaptive Cruise Control}
\label{s:caccattack}
\ac{CACC} \cite{dey2015review} systems enable efficient transportation by allowing vehicles to communicate and behave cooperatively. However, this dependence on communication exposes vehicles to various security threats due to the necessary use of a network.
Some studied attack models in \ac{CACC} systems include jamming, data injection, and sensor manipulation. Jamming can be regular or reactive, disrupting communication and affecting message reception. Data injection involves sending falsified or altered data packets, which can be countered by an \ac{MDS}. Sensor manipulation involves direct manipulation of vehicle sensors to feed incorrect data, which can be mitigated through secure communication protocols. 

Various \ac{CACC} controllers have been analyzed in the research work \cite{van2017analyzing} for their vulnerabilities to these attacks. The constant spacing controller aims to maintain a fixed distance between vehicles but is particularly vulnerable to jamming and data injection attacks. The Ploeg controller \cite{ploeg2011design}, designed to degrade in communication failures, predicts the acceleration of the preceding vehicle when data transmission fails, maintaining some level of control and safety. The consensus controller \cite{ren2005consensus, zhu2015necessary, liu2021consensus}, using information from all vehicles in the platoon, shows resilience against data injection attacks but requires a higher degree of communication fidelity. But all the simulations made underline the importance of incorporating advanced \acp{MDS} or designing controllers that can maintain some level of functionality even when faced with sophisticated attacks.

Research on the constant space version of \ac{CACC} \cite{wolf2020securing} tried to mitigate the false data injection attack with some strategies like a direct fallback to \ac{ACC}, the use of a suspiciousness parameter, and the modification of the control input. In this case, a platoon of 8 vehicles with a sinusoidal speed pattern, implemented by the leader, is simulated, and the attacker try to inject some false speed data. A simple \ac{MDS} is configured to recognize attacks by comparing acceleration data from different vehicles and determining inconsistencies. For sure, under this type of simple attack, the fallback to an \ac{ACC} will bring the loss of all the good properties of a \ac{CACC}, in particular, the one with constant spacing. In the end, the good result of this study is that the researchers tried to create a kind of protocol to manage the detection of misbehavior. 
Another proposed approach against the false data injection attack could be the one inspected in \cite{cunningham2023datainjection}, in which the developers created a secure \ac{CACC} designing from scratch the control system, trying with the help of an \ac{NN}-based \ac{MDS}, to intercept and eliminate all the possible attacks about data injection. This particular research presents a complex environment of simulations using both Matlab and a \ac{HiL} approach, with the setup of the created algorithm on a golf cart.

\subsection{Achievement}
From a deep analysis of the literature about security and privacy, it emerges that standardization has made a lot of progress in that field by designing a secure architecture from a cryptographic point of view and some good approaches that could be used as privacy management, in order to maintain security and safety in the \ac{VN} environment, but also remove some possible problems about privacy that a driver could require. Furthermore, by analyzing some other attacks, in particular, focusing on cooperative systems, it emerges that, even with this kind of security already standardized, a safe system against some possible attacks, like false data injection or possible misbehavior of the communication systems of some vehicles on the road, could be necessary. In particular, integrating an \ac{MDS} on board could be the best solution to defend against that kind of problem.

\section{Misbehavior detection systems literature}
\label{s:misbeh_literature}

As the previous section said, traditional security mechanisms are sometimes insufficient because they only exclude outsider attackers who lack the proper cryptographic keys. 
However, insiders can still pose a threat since they might possess valid credentials, and the vehicles may still misbehave due to malfunctioning or malicious behavior. 
To address these security gaps, the authors of \cite{van2018survey} introduce and analyze various misbehavior detection mechanisms that are capable of identifying insider attacks or malfunctioning based on the analysis of behavioral patterns. 
Here follows an overview of the types of misbehavior detection mechanisms analyzed:

\begin{itemize}
    \item \textbf{Behavioral detection:} Identifies anomalies by monitoring network nodes' behavior, such as message frequency and adherence to protocol norms. An example is a \textit{watchdog} that checks packet forwarding without needing to understand message content.
    \item \textbf{Trust-based detection:} Creates a reputation system where nodes earn and update trust scores based on their actions, often using voting mechanisms.
    \item \textbf{Consistency-based detection:} Compares data from different nodes about the same event to find discrepancies indicating falsified information.
    \item \textbf{Plausibility-based detection:} Verifies if reported data, like speeds or positions, align with expected behavior models and physical constraints.
\end{itemize}

These mechanisms have strengths but also some limitations, especially against
an attacker who uses sophisticated strategies to mimic normal behaviors.
So future researches in \ac{CITS} focus on enhancing security and efficiency through hybrid detection mechanisms and advanced \ac{AI}. Key areas include improving scalability, addressing new vulnerabilities, and balancing security with privacy. 

The survey faced up in \cite{van2018survey} prepared the field for new research topics, and one of the main ones could be the usage of \ac{ML} and \ac{AI} techniques to help and design some misbehavior detection algorithms. 
In \cite{boualouache2023survey} is presented a comprehensive overview of how \ac{ML} techniques are utilized to enhance misbehavior detection in \acp{VN}. In the survey is highlighted that while \ac{ML} offers significant advantages for security, challenges such as adapting to dynamic environments and ensuring privacy remain. Here are the primary \ac{ML} methods discussed:

\begin{itemize}
    \item \textbf{Supervised learning:} Trains models using labeled data, subdivided into classification and regression.
    \item \textbf{Unsupervised learning:} Identifies patterns in data without labels, useful for anomaly detection and clustering similar objects, which helps discover new misbehavior in \acp{VN}.
    \item \textbf{Reinforcement learning:} Adapts by continuously improving decisions based on feedback, dynamically adjusting security measures.
    \item \textbf{Deep learning:} Utilizes layered \acp{NN} to process large data volumes and extract complex features.
    \item \textbf{Transfer learning:} Transfers knowledge from one problem to a related one, enhancing efficiency and scalability.
\end{itemize}

The first important comparison is between an \ac{NN} approach against other types of algorithms. In particular two studies \cite{matousek2018robust, matousek2019detecting} compare the same simulation structure firstly with \ac{ML}-based algorithms unsupervised and semi-supervised, like \ac{k-NN}, One-class Support Vector Machine and Isolation Forest, and then with an \ac{NN} approach with algorithms like Autoencoder Replicator Neural Networks and \ac{LSTM} Networks. To compare those different approaches, the papers use the same simulation environment, based on the \ac{LuST} scenario that is designed to replicate a realistic
urban traffic environment based on the city of Luxembourg. In particular, the aggressive behavior class
is inserted into the \ac{LuST} scenario, and the vehicles that belong to
this class are characterized by significantly altered driving parameters compared to normal ones, including acceleration and deceleration, minimum gap in terms of space left between vehicles, maximum speed, generally higher, speed factor and speed deviation, impatience and sub-lane models that generally are represented from
vehicles more aggressive with lane changes and maneuvers. 

The results show that, although \ac{k-NN} has good results, approaches based on \acp{NN} tend to have better performances. This is why, from now on, a set of research about \acp{MDS} that are \ac{NN}-based will be analyzed. In particular, the focus will be on papers that use the \ac{VeReMi} dataset \cite{kamel2020veremi}, which will be analyzed later to compare the performances of the algorithms and the strong and weak points of the simulation infrastructure.

\subsection{Long Short-Term Memory systems}
\label{s:LSTM}
The first and most common approach to a misbehavior detection system based on an \ac{NN} is to use \ac{LSTM} to manage the data sequence. In particular, in \cite{hsu2021deep}, the \ac{VeReMi} dataset is manipulated in a way to obtain sequences of 20 vectors with a sliding window of size 10, labeled as malicious or not. In this case, a \ac{SVM} classifier is also used to extract 11 more features, which include behavioral deviation, location plausibility, velocity information, and comprehensive information. The model is trained on 90\% of the sequences labeled as normal (non-misbehaving vehicles) and the experimental results show that the combined \ac{LSTM} model, followed by an \ac{SVM} classifier, achieves high detection accuracy.

Another approach that includes time sequences is \cite{liu2022misbehavior}; the data are heavily manipulated in this work. Unnecessary labels such as sender ID and message ID were removed, and only the X and Y coordinates of vehicle position and speed were retained for training. Moreover, Ground Truth files and vehicle communication log files were combined to create a comprehensive dataset that includes both actual and claimed vehicle data. In the end, duplicate messages and unnecessary indexes were deleted using Matlab scripts. Like the previous approach, the input is a sequence of messages sorted by communication time, capturing the temporal dependencies of vehicle position and speed. The Softmax activation function was used in the output to classify the data into six categories (one normal vehicle type and five attacker types). The \ac{LSTM} model achieved an accuracy of around 90\% on the validation set and 88\% on the test set.

Also, in \cite{alladi2023deep} was used an approach \ac{LSTM}-based and the creation of a time sequence of 20 data points, and each data entry was created containing label, send time, pseudo ID, and X, Y coordinates of position and velocity.

\subsection{Reinforcement Learning}
\label{s:RL}
\acp{LSTM} represent only the simplest approach to the problem. More complicated solutions, always working on the same dataset, were adopted over the years. In \cite{sedar2022reinforcement} the researchers present a novel
approach utilizing \ac{RL} to enhance misbehavior detection
in \ac{V2X} communications. The \ac{NN} used in the \ac{RL} model is structured around a Q-learning framework tailored for time-series data analysis in \ac{V2X} scenarios. The network includes a \ac{LSTM} layer,
which is crucial for capturing temporal dependencies in the data. This \ac{LSTM}
layer feeds into a fully connected layer that outputs Q-values, representing
the potential utility of actions taken by the network in given states. The model employs an $\epsilon$-greedy strategy for exploration and exploitation, enhancing its ability to
learn from new data and adjust to evolving conditions effectively. The results of the simulations demonstrate the high efficacy of the \ac{RL}-based misbehavior detection model. The model achieved a remarkable recall of 99.70\% and an F1 score of 98.45\%, indicating a superior
ability to accurately identify misbehavior without a high rate of false positives.

\subsection{Federated Learning}
Another complex solution is presented in \cite{marmol2024federated} where a \ac{FL} approach is used. In particular, this solution presents the possibility of using a cloud-based solution, thanks to the use of some local deep learning techniques like \acp{VAE} and \acp{GMM}. Each vehicle uses its locally refined \ac{VAE}, after being provided from a centralized \ac{RSU}, to monitor incoming data for potential
misbehavior. The likelihood function from the \ac{GMM} first assesses if new data
points conform to the expected distributions of benign behavior. Points that raise suspicions based on the \ac{GMM} likelihood are then processed through the
\ac{VAE}, where a significant reconstruction error indicates potential misbehavior. This complex solution is certainly interesting, but to achieve an on-line detection in which each vehicle can predict independently from a centralized solution, \ac{FL} doesn't represent the correct approach to reach this goal.

\subsection{On-line simulations approach}
All the solutions presented from now only do an offline approach by training and testing their model on the dataset, leaving out the possibility of simulating the misbehavior detection in a real-world scenario. Right now, the focus is moved on works that present some simulations results in addition to creating a misbehavior detector \ac{ML}-base. The first example presented is \cite{bouchouia2023multi}, in which a model that leverages a combination of \ac{ML} techniques and \ac{RL} for effective misbehavior detection is used. This example provides simulations using CARLA, SUMO, and ARTERY simulators, and in particular, it tried to replicate the scenario used to create the \ac{VeReMi} dataset. At a certain point, various types of misbehavior are injected into the simulation environment, such as false data injection, message tampering, and \ac{DoS} attacks. Consequently, a real-time detection based on the detector is used to predict the misbehavior and make proper decisions. 

Another important example is the research made in \cite{hawlader2021intelligent}. This example, in particular, defines a setup of a simulation environment using OMNeT++ and SUMO in which every vehicle, equipped with a \ac{ML} algorithm selected from a pool of six possibilities, has to face up some misbehavior like constant position attack, constant offset position attack, random position attack, random offset attack, eventual stop attack. In particular, when a vehicle intercepts a misbehavior, it has to report the detection or the accident to a centralized trusted authority.

\subsection{Achievement}
What is possible to see, from a deep analysis of some papers about \acp{MDS} based on an \ac{NN} approach, is that the literature is full of possible models that, in particular, work well on the \ac{VeReMi} dataset, properly created to work on misbehavior or attacks. What is missing in a lot of works about this subject is firstly the definition of a set of simulations that are independent from the \ac{VeReMi} dataset and not centralized, in a way to see if the detection system can truly work well in a different environment and on every vehicle, and then the definition of a safe protocol to manage the detection in a proper way. 
By analyzing the literature, as anticipated in \cref{s:goal}, the purpose set for this thesis is not to define the best \ac{NN} system to detect all the attacks in the \ac{VeReMi} dataset, but to define a good misbehavior detector based on the models used in the analyzed works and the data manipulation that they made, that can work well on some particular attacks of the dataset that can be easily replicated and adapted to a new environment, e.g., a highway. Then try to do an on-line detection that is not centralized but independent in every vehicle and different from the simulation scenarios used to create the dataset. Moreover, to fit the lack of a defense application, the last thesis goal is to define a simple management protocol, based on the \ac{MDS} predictions, that can avoid accidents caused by the emulated misbehavior and bring driving safety in line with \ac{CITS} objectives.

    \chapter{Misbehavior detection system}
\label{ch:misbeh}

The first step to reach the goals set for this thesis is the creation of an \ac{MDS}. 
This system is intended to detect anomalies that can cause the vehicle, or those near it, to behave improperly during the traffic flow while using cooperative driving systems based on the \ac{V2X} communication. 
An example of an anomaly could be the malfunction of the \ac{GPS}. 
From the review in \cref{s:misbeh_literature}, it emerged that the recent works focus on systems based on an \ac{NN} approach. 
Along with the \ac{MDS} implementation, a standard and well-formed dataset to train an \ac{NN} for this purpose has to be chosen. Still from the literature review, in \cref{s:misbeh_literature}, many \ac{ML}-based model developed used the \ac{VeReMi} dataset, a standard for what concerns misbehavior detection. 
Considering \ac{VeReMi} as the basis for the thesis \ac{MDS} implementation, the data have to be manipulated to fit the input of a selected \ac{NN}, in particular, making feature engineering (\cref{s:feat_eng}) and embedding (\cref{s:embedding}) choices. Once the \ac{NN} has been selected (\cref{s:nn}) and trained on the properly modified data, a portion of the dataset is used to validate the model offline, understand its performances and what it has to be expected from the further insertion in the on-line simulations.

\section{Dataset}
The most used dataset in the works analyzed is the \ac{VeReMi} dataset \footnote{The dataset can be accessed through the GitHub URL, in which it's possible to find the download link with all the folders containing the log files for each scenario: \url{https://github.com/josephkamel/VeReMi-Dataset} } \cite{van2018veremi, kamel2020veremi}.
The \ac{VeReMi} dataset was created to overcome the challenges of reproducibility and comparability in research on vehicular misbehavior detection. 
Before its creation, studies in this domain relied on individually designed simulation scenarios, which allowed for customization but made it difficult to compare the effectiveness of different detection mechanisms. 
By providing a common reference dataset, \ac{VeReMi} facilitates the comparison of results across different studies, promoting a more rigorous scientific approach to evaluating misbehavior detection systems.

The dataset was generated using the \ac{LuST} scenario, an open-source synthetic traffic scenario validated with real data. 
This scenario simulates traffic conditions in a subsection of Luxembourg City, covering an area of 1.61 $km^2$ with a peak vehicle density of 67.4 vehicles per $km^2$. The simulations are conducted using VEINS, integrating OMNeT++ for network simulation and SUMO for traffic simulation.

The \ac{VeReMi} dataset comprises message logs from simulated vehicular communications and ground truth files that document each vehicle's behavior, including those simulating malicious behavior. 
The dataset includes several misbehavior scenarios, with data collected from multiple simulation runs under varying vehicle and attacker density conditions. Key components of the dataset include:

\begin{itemize}
    \item \textbf{Message Logs:} These logs contain detailed records of periodic messages exchanged between vehicles.
    \item \textbf{Ground Truth Files:} These files provide the actual positions and behaviors of vehicles. This information is crucial for validating the detection mechanisms being tested.
    \item \textbf{Attack Scenarios:} The dataset includes various predefined attack types, such as constant position attacks, random position attacks, constant offset attacks, random offset attacks, and eventual stop attacks. Each attack type has specific parameters and behaviors that are systematically applied during the simulations.
\end{itemize}

The dataset is presented in 2 different versions. In particular, in the first one \cite{van2018veremi}, the simulations were made with the parameters presented in \cref{tab:simulation_parameters}.
\begin{table}[H]
\centering
\scriptsize
\begin{tabular}{|l|l|l|}
\hline
\textbf{Parameter}                & \textbf{Value}                      & \textbf{Notes}                           \\ \hline
Mobility                          & SUMO LuST (DUA static)              &                               \\ \hline
Simulation start                  & (3,5,7)h                            & controls density                         \\ \hline
Simulation duration               & 100s                                &                                          \\ \hline
Attacker probability              & (0.1, 0.2, 0.3)                     & attacker with this probability           \\ \hline
Simulation Area                   & 2300,5400-6300,6300                 & various road types                       \\ \hline
Signal interference model         & Two-Ray Interference                & VEINS default                            \\ \hline
Obstacle Shadowing                & Simple                              & VEINS default                            \\ \hline
Fading                            & Jakes                               & VEINS default                            \\ \hline
Shadowing                         & Log-Normal                          & VEINS default                            \\ \hline
MAC implementation                & 802.11p                             & VEINS default                            \\ \hline
Thermal Noise                     & -110dbm                             & VEINS default                            \\ \hline
Transmit Power                    & 20mW                                & VEINS default                            \\ \hline
Bit rate                          & 6Mbps                               & VEINS default (best reception)           \\ \hline
Sensitivity                       & -89dBm                              & VEINS default                            \\ \hline
Antenna Model                     & Monopole on roof                    & VEINS default                            \\ \hline
Beaconing Rate                    & 1Hz                                 & VEINS default                            \\ \hline
\end{tabular}
\label{tab:simulation_parameters}
\caption{Simulation parameters of the first version of the \ac{VeReMi} dataset \cite{van2018veremi}.}
\end{table}

In this version of the dataset, only a few attacks are presented. Through the years, the newest version of the dataset was created \cite{kamel2020veremi}. Something has been changed; in particular, in this new version, the simulations are made throughout the day, but then 2 different time intervals are selected: between 7 a.m. and 9 a.m., which is considered high density, and 2 p.m. and 4 p.m., which is considered low density. In all the simulations, the percentage of vehicle that are misbehavior is 30\%.

\subsection{Misbehavior presented}

The most recent version of the dataset includes 19 types of malfunctions, anomalies, or attacks. 
For what concerns misbehavior, these are the main ones:

\begin{itemize}
    \item \textbf{Position malfunctions}: are usually a result of a positioning
    system failure (e.g., \ac{GPS}). These failures affect the longitude and latitude fields of the safety messages and could
    manifest as one of these four use cases:
    \begin{enumerate}
        \item The position is constant throughout the simulation: $Pos_{t+1} = Pos_t$
        \item The position is random at every time-step, considering an interval through all the surface of simulation: $Pos = U[min, max]$
        \item A constant offset is added to the real position: $Pos_t = Pos_t + off$
        \item A random offset is added to the real position: $Pos_t = Pos_t + U[offsetMin, offsetMax]$
    \end{enumerate}

    \item \textbf{Speed Malfunctions}: this could be the result of an \ac{OBU}
    error or a physical sensor failure. The speed malfunction is generated similarly to the previously described
    position malfunction. This results in a Constant, Random,
    Constant Offset and Random Offset modification of the
    $V_x$ and $V_y$ fields.

    \item \textbf{Delayed Messages}: could be a result of a large network
    overhead or a low-cost or slow on-board processing unit.
    These messages contain all the correct data and required
    information but are sent with delay $\Delta_t$ from reality.
\end{itemize}

While the attacks introduced in the \ac{LuST} scenario are:

\begin{itemize}
    \item \textbf{\ac{DoS} attacks}: consists of a vehicle sending messages with
    a frequency higher than the limit set by the corresponding the
    \ac{IEEE} or \ac{ETSI} standards.
    \item \textbf{\ac{DoS} Random}: are \ac{DoS} attacks with all the message
    fields set to random values. It could be a strategy to flood
    the network and prevent genuine messages from being
    broadcast. This attack could also be executed in Sybil mode, with the attacker changing its identity on every 
    message sent to avoid detection.
    \item \textbf{Data Replay}: sends information previously
    received from a specific target neighbor. The replayed information is signed with the attacker’s certificate. It could
    be executed in Sybil mode, with the attacker changing its
    identity on every new chosen target to avoid detection.
    \item \textbf{Disruptive attacks}: are an information replay of previously received data from random neighbors. It could also
    be a strategy to flood the network and prevent genuine
    messages from being broadcast. This attack could also
    be executed in Sybil and \ac{DoS} modes.
    \item \textbf{Eventual Stop}: are attacks where a vehicle simulates a
    sudden stop by freezing the position values and setting
    the speed values to null.
    \item \textbf{Traffic congestion Sybil}: is an attack to create 
    fake traffic congestion. The attacker generates a grid of
    fake vehicles in a chosen position by maintaining a new
    identity and a correct message frequency for each fake
    vehicle.
\end{itemize}

\subsection{Selected Anomalies}
\label{s:selected-mis}
In this research, as explained i \cref{s:goal}, the choice is not to create an \ac{NN} on all the scenarios available, but a good amount of cases, from those presented in the previous section, that can be easily replicated and studied in a real-world scenario, i.e., the simulation environment that is explained in \cref{ch:simuenv}.
In particular, the chosen misbehavior are \textit{position malfunctions}, in constant, random, and random offset position scenarios, \textit{speed malfunctions}, in random and random offset scenarios, \textit{eventual stop}, \textit{disruptive attacks} and \textit{data replay}.
The malfunctions are selected because, in a real-world scenario, these kinds of anomalies can truly happen, e.g., \ac{GPS} malfunctions that can transmit wrong positions or \ac{OBU} anomalies. The other misbehavior selected can be replicated in simulation without many problems, allowing an on-line study of the \ac{MDS} implemented on every vehicle in different real-world scenarios, e.g., urban or highway.

\begin{figure}[H]
    \centering
    \includegraphics[width=1\linewidth]{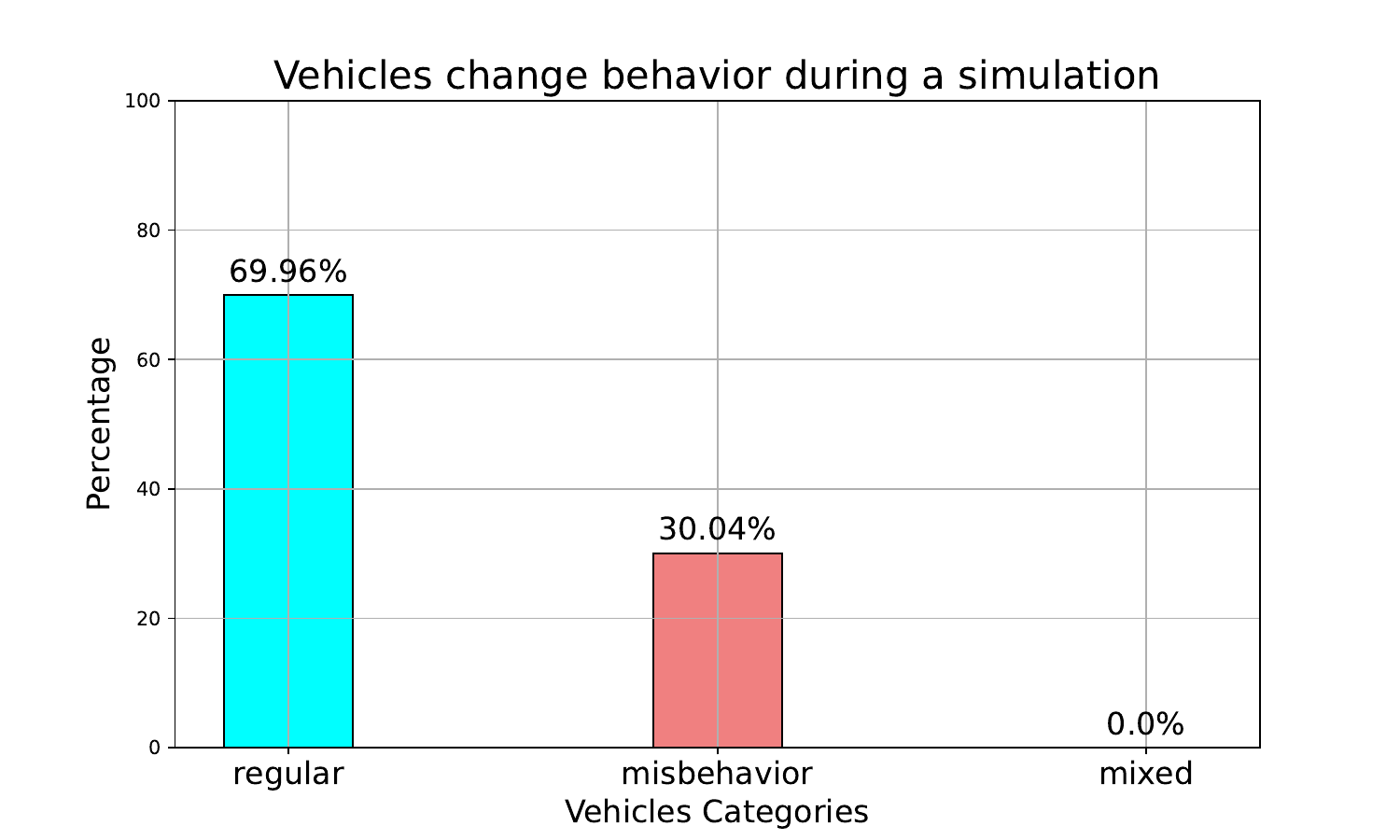}
    \caption{How many vehicles change their type (regular or misbehavior) during the \ac{VeReMi} constant position offset simulation}.
    \label{fig:constoff-cat}
\end{figure}

It's important to underline that the \textit{constant offset} scenario was not used from the position and speed malfunctions. 
In fact, during the \ac{VeReMi} dataset simulations, as shown in \cref{fig:constoff-cat}, every vehicle can be regular\footnote{From now on, every vehicle that is running without sending misbehavior messages is called \textit{regular}}, which means that every message sent is without anomalies, or misbehavior, which means that every message sent is with the specified anomaly.
However, no one can be mixed\footnote{A vehicle that changes its misbehavior state while a simulation is running, starting, or stopping to misbehave during a simulation.}, and this behavior is clearly not representative of a real-world scenario, since in the world there's not a beginning or an end of a simulation. Moreover, since the thesis simulations aim to represent a real-world scenario and detect misbehavior while a simulation is running, every vehicle could be considered mixed, as defined. Given all the misbehavior beginning while the simulation is running, it is sufficient to train the model on the \textit{random offset scenario} to intercept also the \textit{constant offset}. In fact, at the beginning of the misbehavior, the first \textit{constant offset} message, that is the one that determines the meaningful prediction of the \ac{MDS} and the application of the protocol further defined in \cref{ch:method}, is seen as a \textit{random offset} since the message includes any offset compared to the previous one. So both random and constant offset can be called more in general offset.
Since training the \ac{NN} on the \textit{random offset} scenario is sufficient to intercept every offset malfunction, even if it's constant, to avoid some noise that can be introduced from the \textit{constant offset} scenario while training the model, that is particularly difficult to learn, the decision is not to include this scenario in the training. To simplify the nomenclature, the \textit{random offset} scenario is from now called \textit{offset}. 

Another scenario excluded from the training and the on-line evaluation is the \textit{constant speed}. This scenario can introduce a lot of noise during the on-line prediction, considering that in a real-world scenario for some particular \ac{VN} goals, like platooning, the purpose is to reach a constant speed, so in a simulation environment that aims to maintain a constant speed, it doesn't make any sense to train a model that, when the speed is constant, returns a misbehavior label. Moreover, even if the vehicles are oscillating and not constant in speed, sending at a certain time a message with a constant speed misbehavior, with a value close to the ones between the vehicles oscillates, doesn't create some particular problem on the other platoon's vehicles, in particular, no accidents happens. So, as before, to avoid some noise during the on-line simulations, the decision is to delete the \textit{constant speed} scenario from the model training.

\subsection{Feature engineering}
\label{s:feat_eng}
\begin{figure}[H]
    \centering
    \includegraphics[width=1\linewidth]{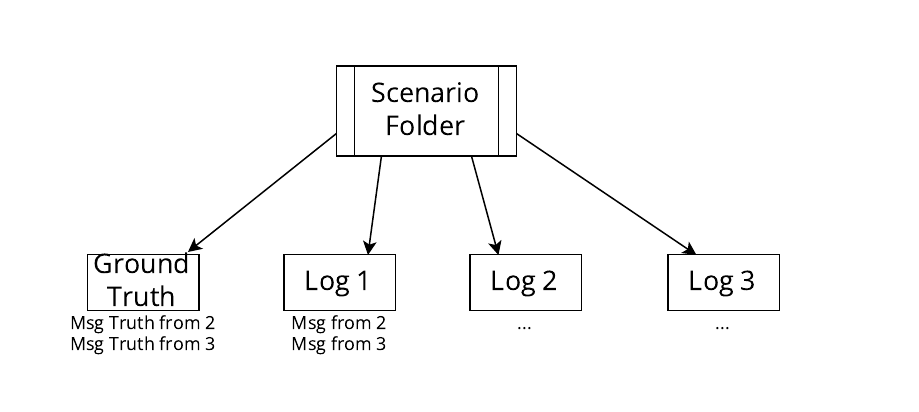}
    \caption{Subfolders' structure of the \ac{VeReMi} dataset}
    \label{fig:folder-veremi}
\end{figure}

The data in the extended version of the \ac{VeReMi} are presented as a folder for every scenario, which can be between 7 a.m and 9 a.m. or 2 p.m and 4 p.m., containing a \textit{GroundTruth} file, or more than one, and a \textit{Log} file for each vehicle in the simulation. 
In particular, each log file contains all the messages received from the other vehicles in the simulation, which are then represented in the Ground Truth JSON with the real data. 
So, referring to the example in \cref{fig:folder-veremi}, if the message from 2 is misbehavior with false position data, the truth message from 2 contains the real position of the vehicle 2 for that specific message.

\begin{figure}[H]
    \centering
    \includegraphics[width=1\linewidth]{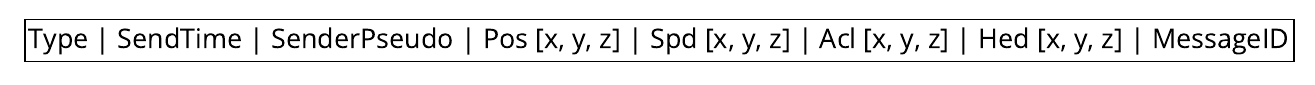}
    \caption{Structure of a message in the log files}
    \label{fig:msg-struct}
\end{figure}

The data structure of every message in each \ac{VeReMi} log file is composed of the features presented in \cref{fig:msg-struct}. 
In particular, from the feature \textit{type}, the only interesting value for this thesis is type 3, which stands for messages received from other vehicles. 
The \textit{SendTime} and \textit{SenderPseudo} represent the time at which the vehicle enumerated as the \textit{senderPseudo} feature is transmitted. 
While the \textit{Postion}, \textit{Speed}, \textit{Acceleration}, and \textit{Heading} features are written as a 3D vector, in which the z coordinate is always 0, as SUMO does not consider elevation. 
The last important feature is the \textit{MessageID}, which identifies the message compared to the others in the simulation.
 
While scanning all the files, a feature representing the vehicle receiving the message was added since each log file represents a vehicle. This feature is called \textit{rx}, which stands for receiving vehicle.
Furthermore, the position and speed features were extracted in new features \textit{posx, posy, spdx, spdy} so that it's possible not to use the vectors anymore. 
For the acceleration, the Euclidean norm of the x, y vector was computed, but to reconstruct the sign of the vector, it was necessary to compare the angle computed with the vehicle's heading. The angle of the acceleration and heading was computed as follows:
\begin{lstlisting}[language=python]
def calculate_horizontal_angle(x, y):
    angle = math.degrees(math.atan2(y, x))
    angle = angle % 360
    return angle
\end{lstlisting}
While the comparison between the heading angle and the acceleration angle to assign the sign is like this:
\begin{lstlisting}[language=python]
df['acl'] = df.apply(
    lambda row: -row['acl'] if abs(row['hed'] - row['acl_angle']) > 90 
    else row['acl'], axis=1)
\end{lstlisting}
So, if the difference is greater than 90 degrees, the acceleration is considered negative since it's in the opposite direction of the vehicle's direction.

\subsection{Labels}
\label{s:lab}
The most important feature to be recovered from the dataset and needed to train the \ac{NN} is the \textit{label} that represents a kind of misbehavior.
The choice is between using a \textit{single} label model, in which the label feature can be only two values: 0 for the regular vehicles and 1 for the misbehavior ones, or using a \textit{multiple} label model that aims to not only intercept the misbehavior but also identify they. Since the protocol, presented in \cref{ch:method}, uses only the prediction between misbehavior or not, the feature label can be \textit{single}. Considering that this protocol is straightforward and many modifications can be applied, particularly exploiting the differences between the various misbehavior adopting different responses based on the prediction, a \textit{multiple} label is chosen. Since to save a vehicle on the road with the thesis protocol, the \textit{single} label is enough, the metrics results are presented in \cref{ch:sim} both for the \textit{single} and \textit{multiple} label evaluation, in particular considering all the misbehavior label of the \textit{multiple} option as a general misbehavior for the \textit{single} case.
In particular, the \textit{multiple} labels assigned are:
\begin{itemize}
    \item 0: regular
    \item 1: constant position
    \item 2: random position
    \item 3: random position offset
    \item 4: random speed
    \item 5: random speed offset
    \item 6: eventual stop
    \item 7: disruptive
    \item 8: data replay
\end{itemize}

\begin{figure}[H]
    \centering
    \includegraphics[width=1\linewidth]{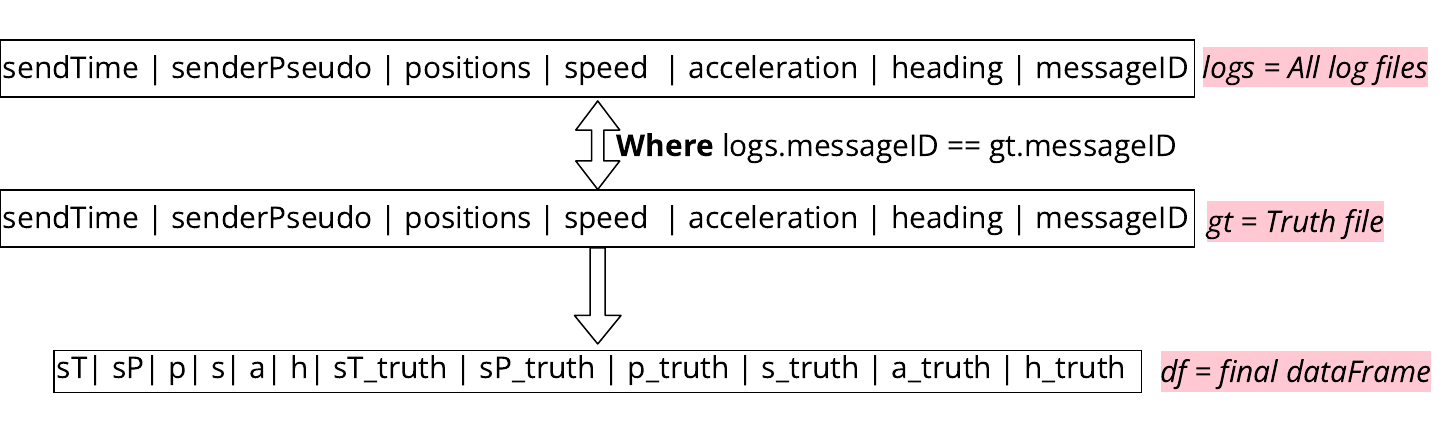}
    \caption{Representation of the data merging between the log files and the ground truth.}
    \label{fig:merge}
\end{figure}

\begin{figure}[H]
    \centering
    \includegraphics[width=0.8\linewidth]{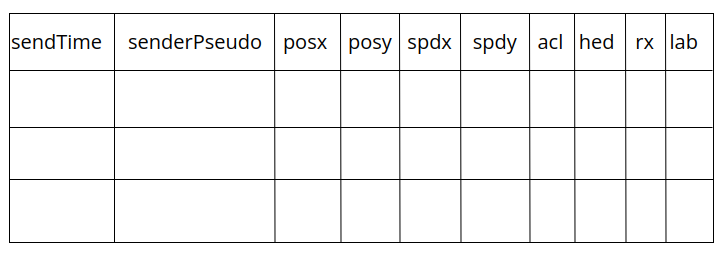}
    \caption{Structure of the final data frame saved for each scenario type}
    \label{fig:df-struct}
\end{figure}

With this kind of feature, it's possible to consider a supervised learning solution for offline training. 
So, to build this feature, all the messages in the log files are merged with the respective \textit{messageID} of the Ground truth, as shown in \cref{fig:merge}.  
Given a full data frame containing entries with all the truth data and the data transmitted for the same message, it is possible to compare all the message's features and assign the relative label of the misbehavior analyzed if one is different. 
Here is the pseudocode of the comparison:
\begin{lstlisting}
foreach f in features:
    if f != f_truth:
        df['lab'] <- label
    else:
        df['lab'] <- 0
\end{lstlisting}

By merging and modifying all the files for a specific misbehavior, it is possible to obtain a unique pandas data frame and then save it into a CSV file called with the scenario name, e.g., "ConstPos\_1416". The structure of the final file is presented in \cref{fig:df-struct}. 

\subsection{Embedding}
\label{s:embedding}

As said in \cref{s:goal}, to emulate the on-line evaluation, a structure of simulators is used, in particular involving \ac{SUMO}, a time-driven simulator used to emulate the urban mobility that orders chronologically the events. Given the mobility simulation ordered by time, also the messages exchanged between vehicles will consequently be time ordered, particularly by looking into the \textit{sendTime} field of every message. Since also the \ac{VeReMi} dataset simulations act in the same way, an important choice is how to deal with the sequence of messages ordered by time, since the sequentiality is important while training the model and while using it on a vehicle. 
In \cref{ch:related}, some works present a possible approach by defining a sliding window dimension containing a message sequence. In particular, in \cite{hsu2021deep}, a sliding window of size 10 is used. By following some research papers, the choice is to use a window of messages and not compute the prediction on every message received. This could also be a good approach that can help the \ac{NN} in learning the misbehavior patterns provided. In particular, the choice is a jumping window\footnote{A jumping window moves in non-overlapping steps, processing distinct and separate chunks of data.} of dimension 5, since in the \ac{VeReMi} simulations, considering the beacon time presented in the \cref{tab:simulation_parameters}, this dimension means 5 seconds, but in the thesis simulations, as presented in \cref{ch:sim}, the beacon time is smaller to achieve particular vehicular network's goals, like platooning. 
So, considering a beacon time of 0.1 s, a jumping window of dimension 5 will mean waiting for 0.5 s to compute a prediction. 
It's reasonable to identify an anomaly in less than 0.5 seconds, considering that a misbehavior, anomaly, or even an attack will not generally cause an accident in that time. 

\begin{figure}[H]
    \centering
    \includegraphics[width=0.8\linewidth]{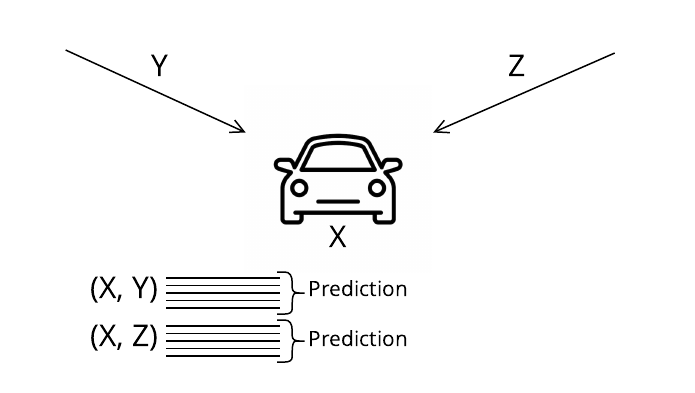}
    \caption{Group every 5 messages received from the same vehicle}
    \label{fig:msg-grouping}
\end{figure}

Since the prediction is made with a jumping window of 5 messages, the \ac{NN} has to be trained of data with the same structure. 
It's logical that the sequence of 5 messages has to be provided by the same vehicle since a sequence of data from different vehicles along the road doesn't make sense. 
Considering the final data frame obtained, shown in \cref{fig:df-struct}, the data entries have to be grouped by sending and receiving vehicles, as shown in \cref{fig:msg-grouping}, to create logical sequences that can be later divided to compose the jumping windows.

The final result has to be an X vector in which each entry is composed of a sequence of five messages grouped by the receiving and the sending vehicle and a parallel y vector that has to maintain the labels of the respective sequence.
To obtain the final X and y vectors ready to be used to train the \ac{NN}, the first step is grouping:
\begin{lstlisting}[language=python]
grouped = df.groupby(['rx', 'senderPseudo'])
\end{lstlisting}
after this line of code, grouped contains all the unique groups in the scenario selected between the receiving and the sending vehicle. It's important to underline that a group can contain a number of messages not divisible by 5, so after ordering the messages of a group by time
\begin{lstlisting}[language=python]
group = group.sort_values(by='sendTime')
\end{lstlisting}
a cut of the exceeding messages was computed in order to achieve divisibility by 5:
\begin{lstlisting}[language=python]
if len(group) % time_steps != 0:
    take = - (len(group) % time_steps)
    group = group.iloc[:take]
\end{lstlisting}

Given a group of 5 messages, the next step is to choose which label from the five messages to assign to the entire group. 
In particular, since during the \ac{VeReMi} simulations, a vehicle in the scenarios selected cannot be mixed, the consequent choice is to use the label of the last of the 5 messages to categorize all the groups.

\begin{figure}[H]
    \centering
    \includegraphics[width=1\linewidth]{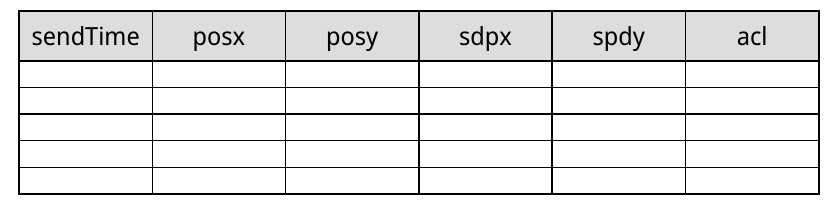}
    \caption{Structure of an entry of the X vector}
    \label{fig:X-struct-init}
\end{figure}

At this point, a data point of the X vector, after the feature engineering and the previously explained embedding step, results in having the features and the size shown in \cref{fig:X-struct-init}. 
Notice that the \textit{rx} and \textit{senderPseudo} columns were removed since it doesn't make any sense to learn the number of the vehicle that is showing an anomaly or the vehicle that is receiving it, but also the \textit{heading} column was removed, since in the simulations further explained the heading is not an important feature. 
It has to be underlined that to import the model in real-world scenarios different from this thesis, it could be useful to still introduce the \textit{heading} column during the training. Given this particular entry, the associated label will be shown at the same index in the y vector. 

\subsection{Population differences}
Given the vector processed, some other problems have to be faced. The \ac{VeReMi} dataset was created based on the \ac{LuST} scenario, an urban scenario created on the city of Luxembourg map. At the same time, if the result that this thesis aims to achieve is a universal misbehavior detector\footnote{An \ac{MDS} that can be used while driving on every kind of road and in every road condition.}, a population bias problem must be deleted somehow. It's enough to consider that the thesis simulation scenario is highway-based. The speed and acceleration data could be different from an urban scenario. Another problem also involves the position features; in fact, the dataset is simulated in the Luxembourg coordinates system, while the test's simulations could be different, or if other researchers wanted to use the model in a completely different environment, the coordinates have to be transformed in an absolute way. The last significant difference is that during the simulation, sometimes a jumping window containing information not every 1 second (the beacon time used in the dataset) appears. In this case, it's important to consider a time delta between the messages. To delete those biases, some solutions are adopted:

\begin{itemize}
    \item \textbf{Coordinates system}: To manage this problem, a solution that significantly changes the embedding of the input data is used. Considering the first message received from the group of five as a reference, all the sequential four messages are changed by evaluating a difference between the message considered and the first. This solution is adopted for the \textit{position, speed} and \textit{acceleration} features to obtain the delta of those dataset columns. In this way, those features now represent the evolution in the four steps considered, given the first message as a reference and not an absolute position anymore. The \textit{speed} and \textit{acceleration} features are also transformed since, as an input of the \ac{NN}, it could be good to present the delta evolution for all the data.
    \item \textbf{Population bias}: Given the previous solution, the speed and acceleration values bias is removed since the evaluation is on the delta and not the absolute value anymore. Another bias on the population could be that the deltas learned are smaller due to the urban scenario used. Still, in this case, a smaller \textit{beacon time} is helpful to balance between the bigger deltas ordinarily present in a highway and the training population. In particular, with a frequency of 10 Hz, the deltas are considered on an interval 10 times smaller than the one used in the \ac{VeReMi}, so this will theoretically give a balance between the two factors.
    \item \textbf{Send time sequence}: The last problem is solved by introducing a feature called \textit{dt}, which stands for \textit{delta time}, that is, the time elapsed between the two sequential messages.
\end{itemize}

\begin{figure}[H]
    \centering
    \includegraphics[width=0.8\linewidth]{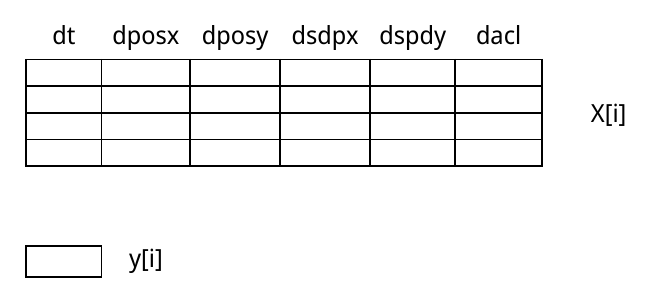}
    \caption{Final X and y structure, ready to be used as the input of the model}
    \label{fig:final-struct}
\end{figure}

So, the final structure of the X and y vectors is presented in \cref{fig:final-struct}, in which the prefix \textit{d} stands for delta. In particular, an entry of the X vector right now comprises those six columns and four rows, while the same index in the y vector is represented by a single cell containing the misbehavior label.

\section{Neural Network}
\label{s:nn}
The choice presented in \cref{s:embedding} is to work with data sequences ordered by time and grouped in a jumping window of five messages. As said in \cref{s:goal}, to evaluate the \ac{MDS} designed on-line, a simulation structure that includes \ac{SUMO}, a time-driven simulator, is used to emulate urban mobility. \ac{SUMO} is a tool in which the operations of a system are represented as a chronological sequence of events. 
Given the analysis made in \cref{ch:related}, the simplest but most effective \ac{NN} model for managing chronological data sequences is a \ac{RNN}. 
In particular, the most common \ac{NN} used is the \ac{LSTM}, which is a type of \ac{RNN} aimed at dealing with the vanishing gradient problem present in traditional \acp{RNN}. 
\acp{LSTM} are considered one of the most advanced models to forecast time series.

\subsection{Long Short-Term Memory}
\begin{figure}[H]
    \centering
    \includegraphics[width=1\linewidth]{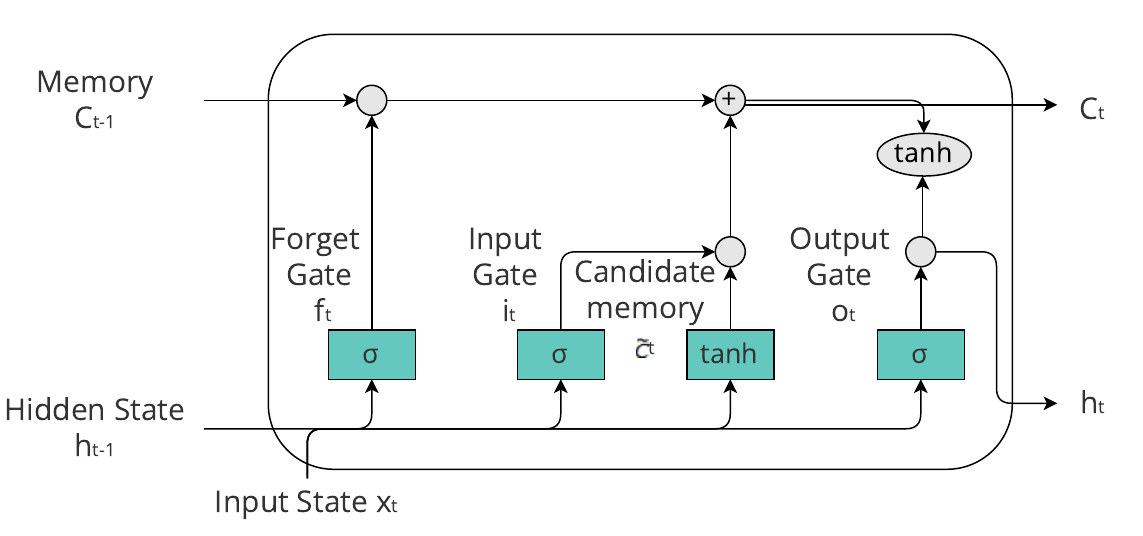}
    \caption{\ac{LSTM} structure}
    \label{fig:lstm-struct}
\end{figure}

The \ac{LSTM} \acp{NN} is a type of \ac{RNN} architecture designed to overcome the limitations of traditional \acp{RNN}, especially in learning and retaining information over long data sequences. 
Introduced by Hochreiter and Schmidhuber in 1997 \cite{hochreiter1997long}, \ac{LSTM} networks are effective at learning long-term dependencies, which makes them suitable for various sequential data tasks. 
The \ac{LSTM} network includes several vital components and formulas, presented in \cref{fig:lstm-struct}, that enable it to process sequential data effectively. 
The Cell State ($C$) is a memory carrying information across different time steps. The Forget Gate ($f$) determines what information to discard from the cell state, calculated as \( f_t = \sigma(W_f \cdot [h_{t-1}, x_t] + b_f) \), in which $\sigma$ is the Sigmoid function, $W_f$ is the weight matrix for the forget weight, $[h_{t-1}, x_t]$ is the concatenation of the previous hidden state and the current input and $b_f$ represents the bias vector for the forget gate. The Input Gate ($i$) decides what new information to add with the formula \( i_t = \sigma(W_i \cdot [h_{t-1}, x_t] + b_i) \), while the Candidate Cell State ($\Tilde{C}_t$) creates new candidate values to be added given by \( \tilde{C_t} = \tanh(W_C \cdot [h_{t-1}, x_t] + b_C) \), in which tanh is the hyperbolic tangent function, which squashes values between -1 and 1. The Output Gate ($o$) decides what information from the cell state to output, using \( o_t = \sigma(W_o \cdot [h_{t-1}, x_t] + b_o) \). Finally, the Hidden State (h), which is the output based on the cell state, is computed as \( h_t = o_t * \tanh(C_t) \).

\ac{LSTM} networks are well-suited for sequential data and time series because they can remember information for long periods and effectively manage the vanishing gradient problem. Traditional \acp{RNN} often struggle with learning from long sequences because gradients tend to vanish or explode during backpropagation. However, the design of \acp{LSTM}, with their cell state and gating mechanisms, mitigates this issue by allowing the network to capture long-range dependencies in the data. Sequential data, such as time series, requires the model to retain information from previous steps to make accurate predictions. \acp{LSTM} excel in tasks where the order and context of the data points are critical, as they dynamically control what information to retain or forget, enabling them to model complex temporal patterns more effectively than traditional \acp{RNN}. Ultimately, the robustness and flexibility of \acp{LSTM} make them a good choice for tasks involving sequential and temporal data, including anomaly detection.

\subsection{Structure}
\begin{figure}[H]
    \centering
    \includegraphics[width=1\linewidth]{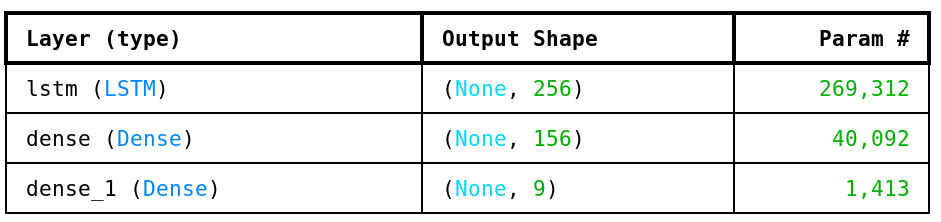}
    \caption{\ac{NN} model structure used for the \ac{MDS}.}
    \label{fig:nn-struct}
\end{figure}

Given the data vector X, composed of a group of four rows and six columns that represent the features, after inserting a first \textit{Input layer} that specifies the size of the training vector, the decision is to put an \ac{LSTM} layer of dimension 256, followed by a Dense layer of dimension 156, with \textit{relu} activation function, and a finale Dense layer of dimension 9, that has the \textit{softmax} as an activation function since it has to be the output layer and classifies the input messages through all the available labels. Here is the Python code used to create this model: 
\begin{lstlisting}[language=python]
model = Sequential([
    Input(shape=(4, 6)),
    LSTM(256, return_sequences=False),
    Dense(156, activation="relu"),
    Dense(9, activation="softmax")
])
\end{lstlisting}
The solution is simple but enough to deal with the misbehavior pattern provided. The \ac{NN} model's summary is presented in \cref{fig:nn-struct}. 

Following, the model is compiled with the \textit{AdamW} compiler, an optimizer that extends the \textit{Adam} (Adaptive Moment Estimation), which is widely used in training deep learning models.  Introduced by Loshchilov and Hutter in their paper \cite{loshchilov2017decoupled}, \textit{AdamW} ensures consistent and effective regularization by applying weight decay directly to the weights, separate from the adaptive learning rate mechanism used in Adam. This approach addresses the suboptimal regularization in Adam, where the learning rate and adaptive updates influence the effective weight decay. As a result, AdamW provides better generalization and reduces overfitting, making it a preferred choice for training deep learning models. Here is the compiled code:
\begin{lstlisting}[language=python]
model.compile(AdamW(learning_rate=0.001),
          loss='sparse_categorical_crossentropy',
          metrics=['accuracy']
        )
\end{lstlisting}
here the \textit{learning rate} selected, after trying different values, is 0.001, while the loss function is the \textit{sparse categorical crossentropy} since the model is dealing with different labels.

\subsection{Training}
\label{s:training}
Before doing the training, more work has to be done. 
In particular, the dataset has to be balanced because otherwise, label 0, which represents the regular vehicles, has a huge number of messages compared to the other. 
Balancing the dataset when using \ac{LSTM} networks or any \ac{ML} model is generally a good practice because it addresses the issue of class imbalance, which can significantly affect model performance. \ac{LSTM} networks, like other \ac{ML} models, learn patterns based on the data they are trained on. Suppose the dataset is imbalanced, meaning one class has significantly more samples than others. In that case, the \ac{LSTM} is more likely to predict the majority class simply because it appears more frequently in the training data. An imbalanced dataset can also cause the \ac{LSTM} to generalize poorly because it might not learn the underlying features of the minority class as effectively as those of the majority class. In the end balancing the dataset before training an \ac{LSTM} can lead to improved model performance, especially when dealing with tasks where the minority class is of significant importance \cite{alkharabsheh2022comparison, ward2021data}.

In the \ac{VeReMi} dataset all the simulations are just replications of the same, all the misbehavior labels have the same number of messages. The decision is to take random messages from regular messages, i.e., label 0, in a double quantity with respect to a misbehavior label. This bias is introduced to take advantage of the regular label and reduce the possibility of predicting misbehavior instead of an regular message in a false positive way since it is more probable that a vehicle is not misbehaving compared to any kind of anomaly presented; the goal to avoid the false positive prediction is explained in \cref{s:structure}.

Another processing adopted for the data is scaling, in particular using the Standard Scaler, a pre-processing technique used to transform the features of a dataset so that they have a mean of zero and a standard deviation of one \cite{burges1998tutorial}. Using the Standard Scaler helps normalize the feature values, leading to improved performance, stability, and convergence in various \ac{ML} algorithms. By ensuring that features contribute equally and consistently, the Standard Scaler enhances the overall effectiveness and reliability of the models.
\begin{lstlisting}[language=python]
scaler = StandardScaler()
scaler.fit(X)
\end{lstlisting}
The scaler is also exported using the \textit{pickle} library in a way to later be used in the simulation environment to scale the data while the simulation is running.

The last step before effectively training the model is dividing the dataset between the train and the test set. To evaluate the offline training, 33\% of the dataset is used to compose the validation set, which is used to validate the model trained on the remaining 77\%. In this way, it's possible to obtain a full report composed of all the metrics useful in the \ac{ML} environment to evaluate a model.
\begin{lstlisting}[language=python]
X_train, X_test, y_train, y_test = train_test_split(X_diff_scaled,
y, test_size=0.33, random_state=42)
\end{lstlisting}

The training of the previous explained \ac{NN} with the processed dataset is executed with a \textit{batch_size} of 64, 100 \textit{epochs} and also the \textit{EarlyStopping} callback as it follows:

\begin{lstlisting}[language=python]
es = EarlyStopping(
    monitor='val_accuracy', 
    min_delta=0.001, 
    patience=10,
    verbose=1,  
    restore_best_weights=True 
)
history = model.fit(X_train, y_train, epochs=100, batch_size=64,
validation_data=(X_test, y_test), callbacks=[es])
\end{lstlisting}

\section{Performance}
\label{s:perform}

To evaluate the performance of the detector on the offline testing (the validation set), four \ac{ML} metrics are used, which are:

\begin{itemize}
    \item \textbf{Accuracy}: is a measure of the overall correctness of the model. It is the ratio of correctly predicted observations to the total observations. The formula is:
    \begin{equation}
        \text{Accuracy} = \frac{TP + TN}{TP + TN + FP + FN}
    \end{equation}
    in which \textit{TP} stands for true positives, \textit{TN} for true negatives, \textit{FP} represents the false positives, while \textit{FN} the false negatives
    \item \textbf{Precision}: is a measure of the correctness of positive predictions. It is the ratio of correctly predicted positive observations to the total predicted positive observations. The formula is: 
    \begin{equation}
        \text{Precision} = \frac{TP}{TP + FP}
    \end{equation}
    \item \textbf{Recall}: also known as Sensitivity or True Positive Rate, is a measure of the ability of the model to find all the relevant cases within a dataset. It is the ratio of correctly predicted positive observations to all observations in the actual class.
    The formula is:
    \begin{equation}
        \text{Recall} = \frac{TP}{TP + FN}
    \end{equation}

    \item \textbf{F1-Score}: is the harmonic mean of Precision and Recall, providing a single metric that balances both concerns. It is particularly useful when a balance between Precision and Recall is required. The formula is:
    \begin{equation}
        \text{F1-Score} = 2 \times \frac{\text{Precision} \times \text{Recall}}{\text{Precision} + \text{Recall}}
    \end{equation}
\end{itemize} 

\begin{figure}[H]
    \centering
    \makebox[\textwidth][c]{\includegraphics[width=1.1\linewidth]{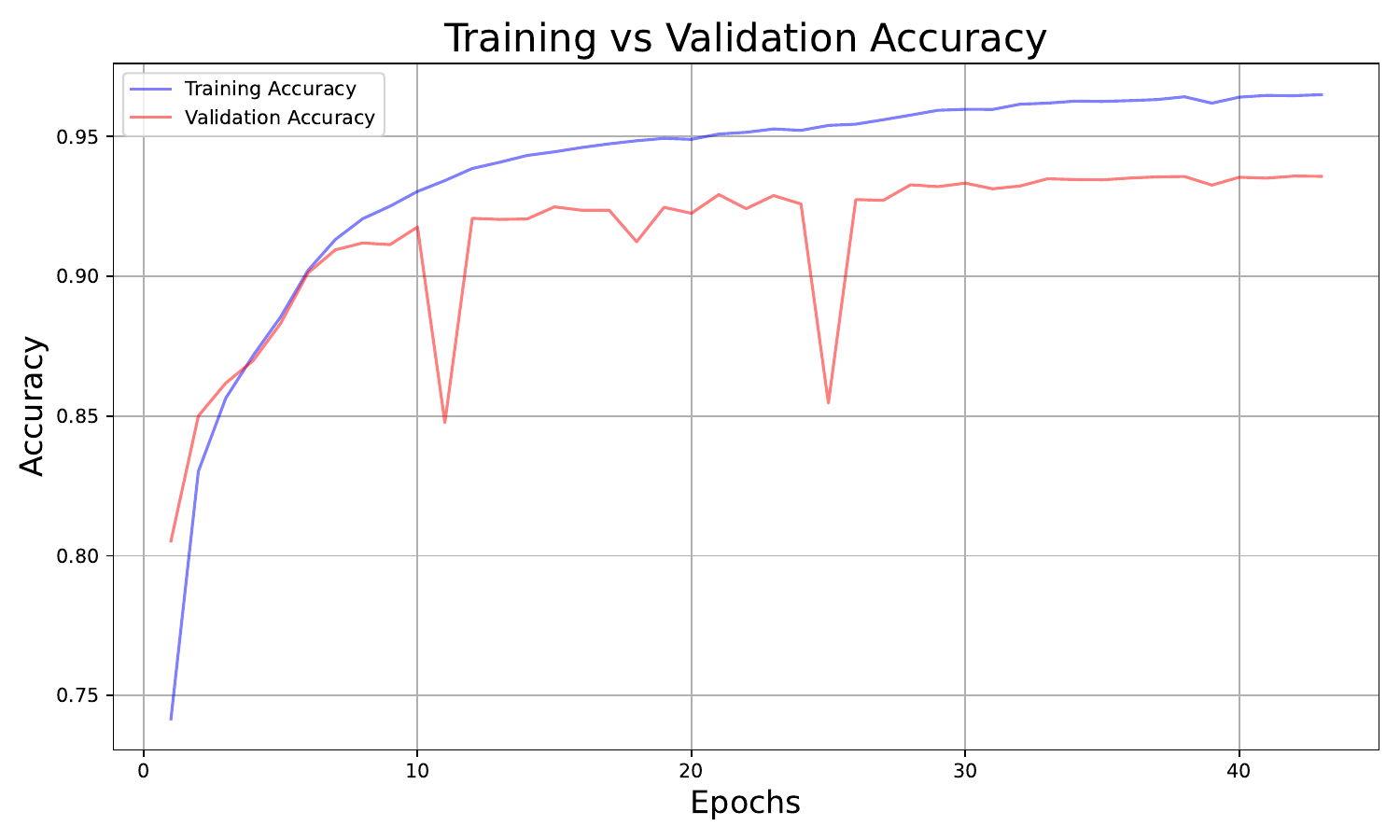}}
    \caption{Learning accuracy, comparing the test one vs the validation one.}
    \label{fig:acc}
\end{figure}

In particular, the accuracy of the test set used to train the model is compared with the accuracy of the validation set for each training epoch. The results in \cref{fig:acc} show that the validation accuracy reaches almost 94\% on the whole dataset. The accuracy value has to be intended as the general correctness in differentiating all the nine misbehavior labels.

\begin{figure}[H]
    \centering
    \makebox[\textwidth][c]{\includegraphics[width=1.2\linewidth]{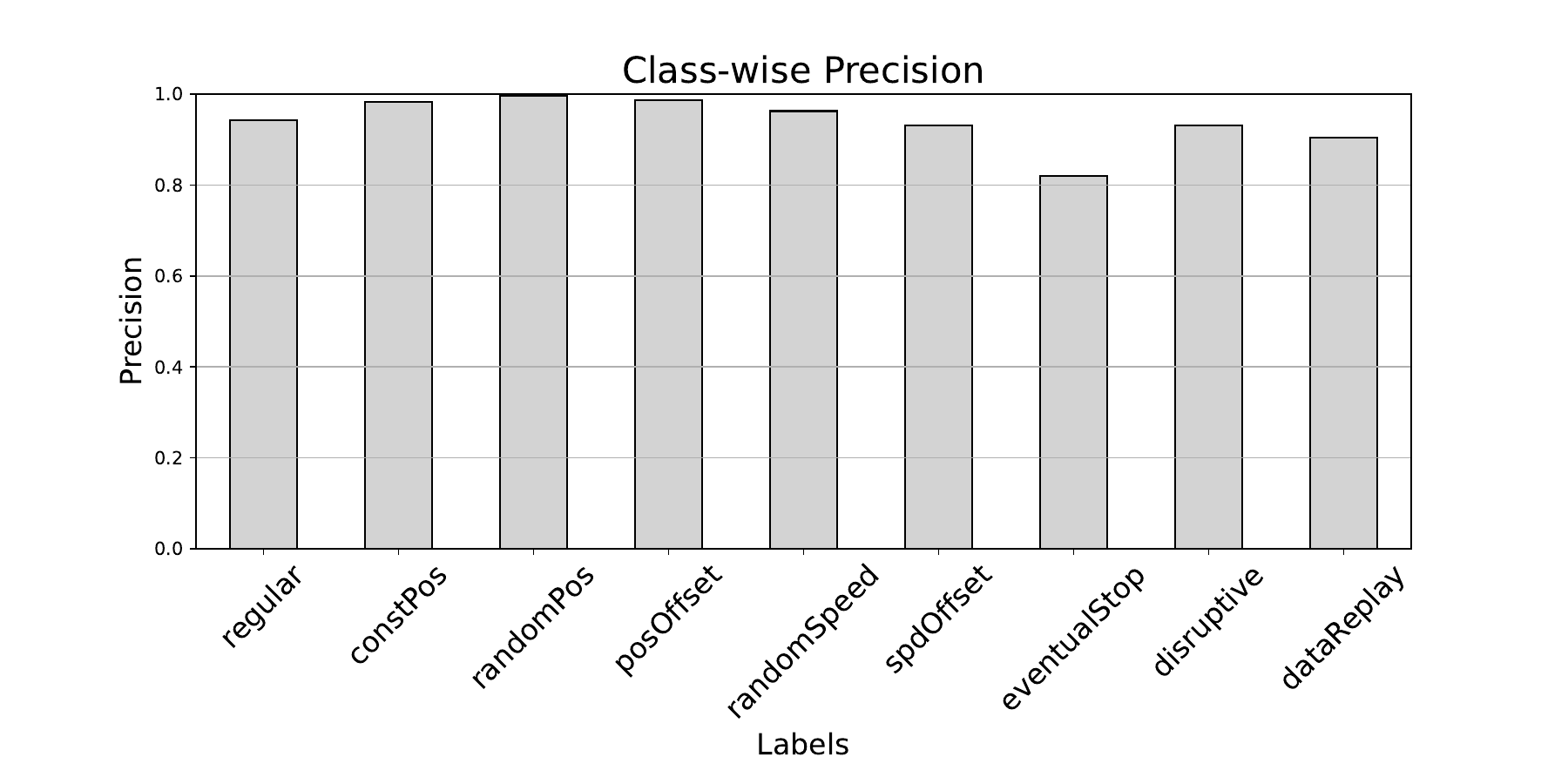}}
    \caption{Precision metric evaluated on every label of the dataset.}
    \label{fig:prec}
\end{figure}

\begin{figure}[H]
    \centering
    \makebox[\textwidth][c]{\includegraphics[width=1.2\linewidth]{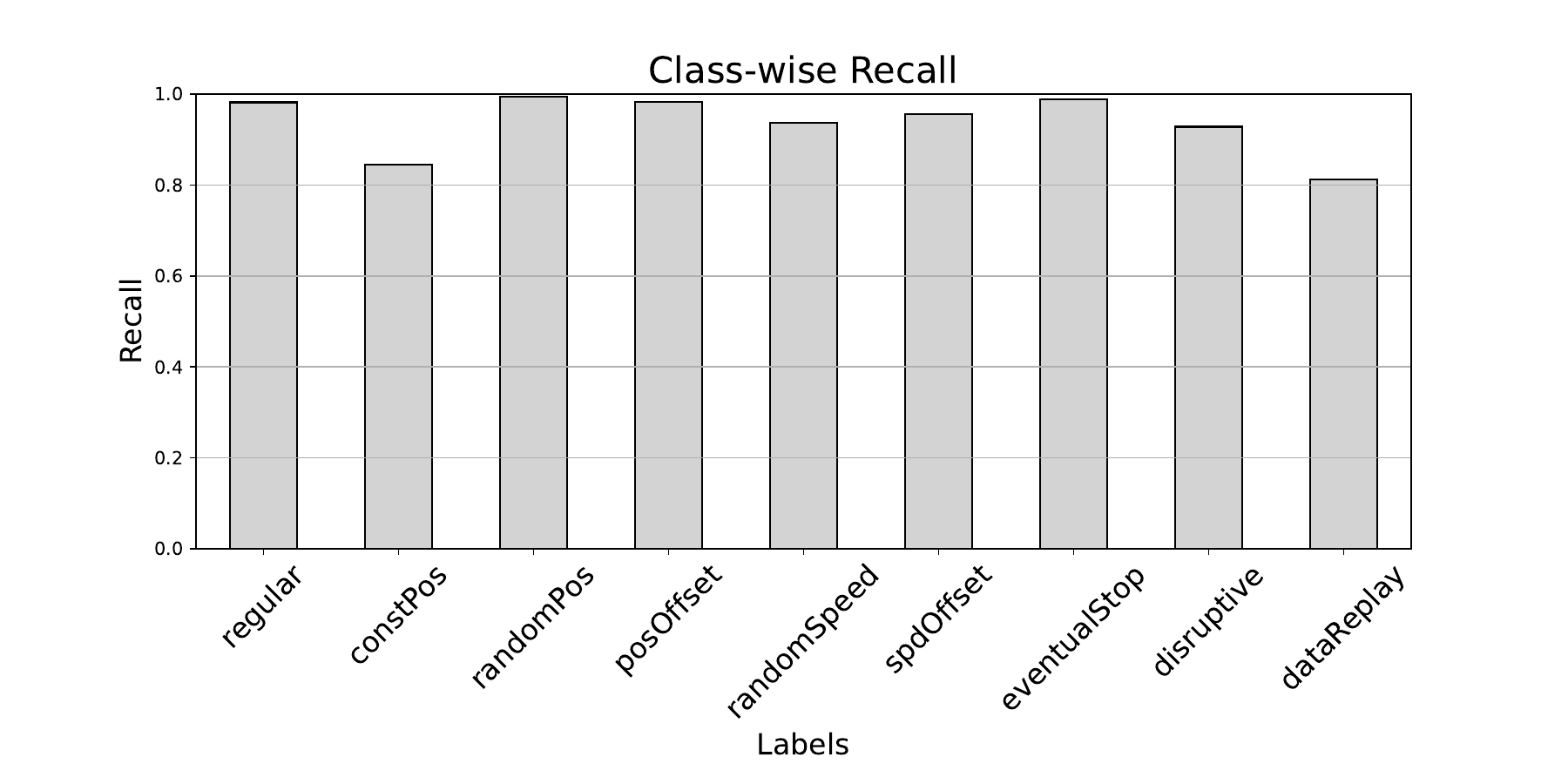}}
    \caption{Recall metric evaluated on every label of the dataset.}
    \label{fig:recall}
\end{figure}

All the metrics are evaluated on each label for what concerns the Precision, Recall, and F1-Score. For the Precision metric, in \cref{fig:prec}, it's possible to see how the positive predictions work better for the labels \textit{constPos}, \textit{randomPos}, and \textit{posOffset} that stand for all the \textit{position malfunction}. While the worst one is the label of the \textit{eventual stop} misbehavior.
As for what concerns the True Positive Rate, in \cref{fig:recall}, there's a generally good average, with the worst cases being the constant position one and the data reply.

\begin{figure}[H]
    \centering
    \makebox[\textwidth][c]{\includegraphics[width=1.2\linewidth]{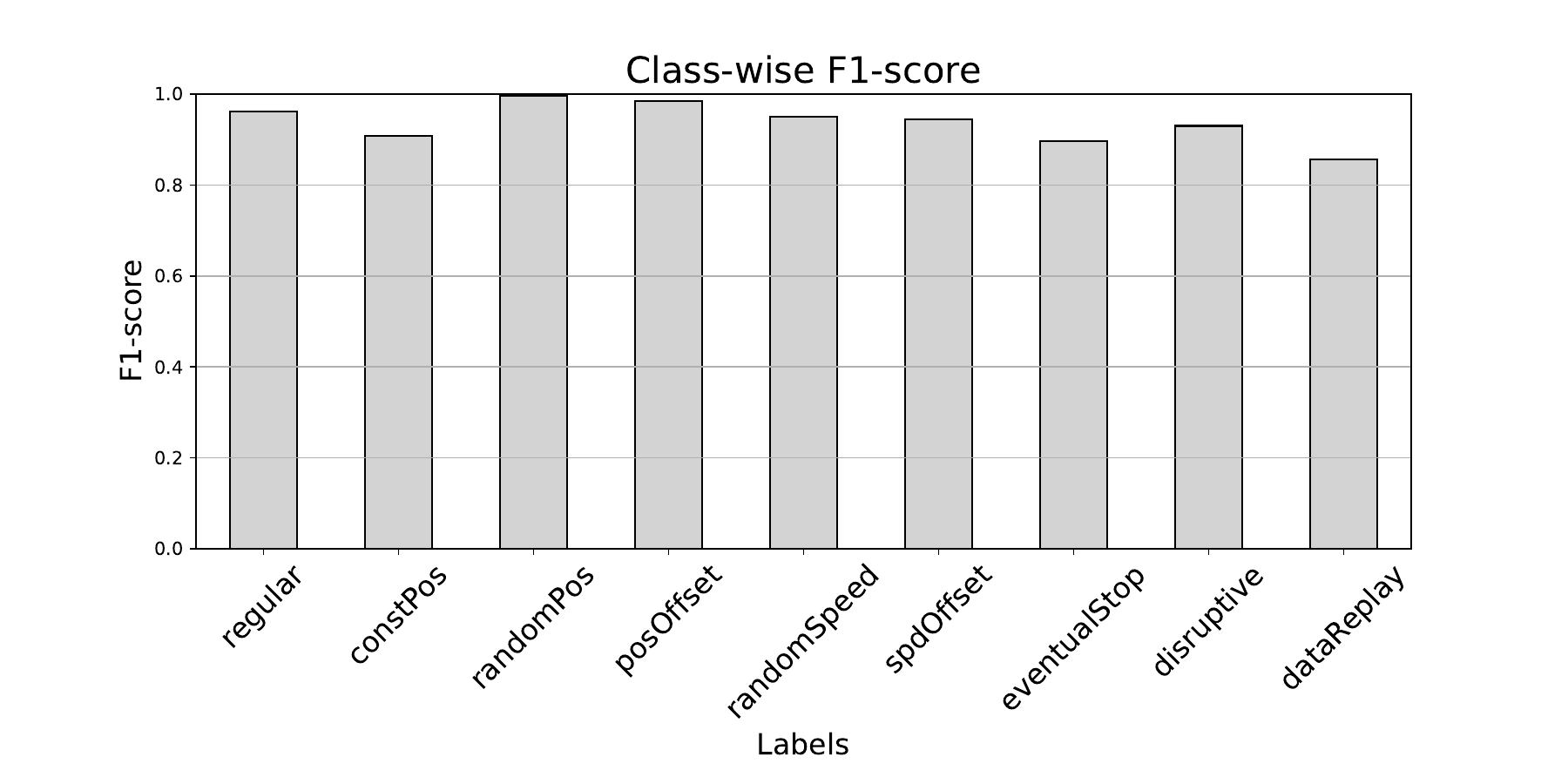}}
    \caption{F1-Score metric evaluated on every label of the dataset.}
    \label{fig:f1-score}
\end{figure}

The F1-Score, in \cref{fig:f1-score}, represents an average balance between precision and recall, and it's possible to see how every label has an F1-Score over 80\%, with the worst case that coincides with the data replay attack. This could be logical if it's considered that the data replay is similar in idea to the disruptive attack, which has a higher F1-Score.

The most important result is that the label of the not-misbehaving vehicles (regular) has a very good average score on every metric. This is meaningful if we consider that the protocol that is designed in \cref{s:protocol} works with the single label, so when it has a prediction that is not regular it starts to work and it's important that the regular label is the best one detected to let the traffic flow as much as possible.

\begin{figure}[H]
    \centering
    \makebox[\textwidth][c]{\includegraphics[width=1.3\linewidth]{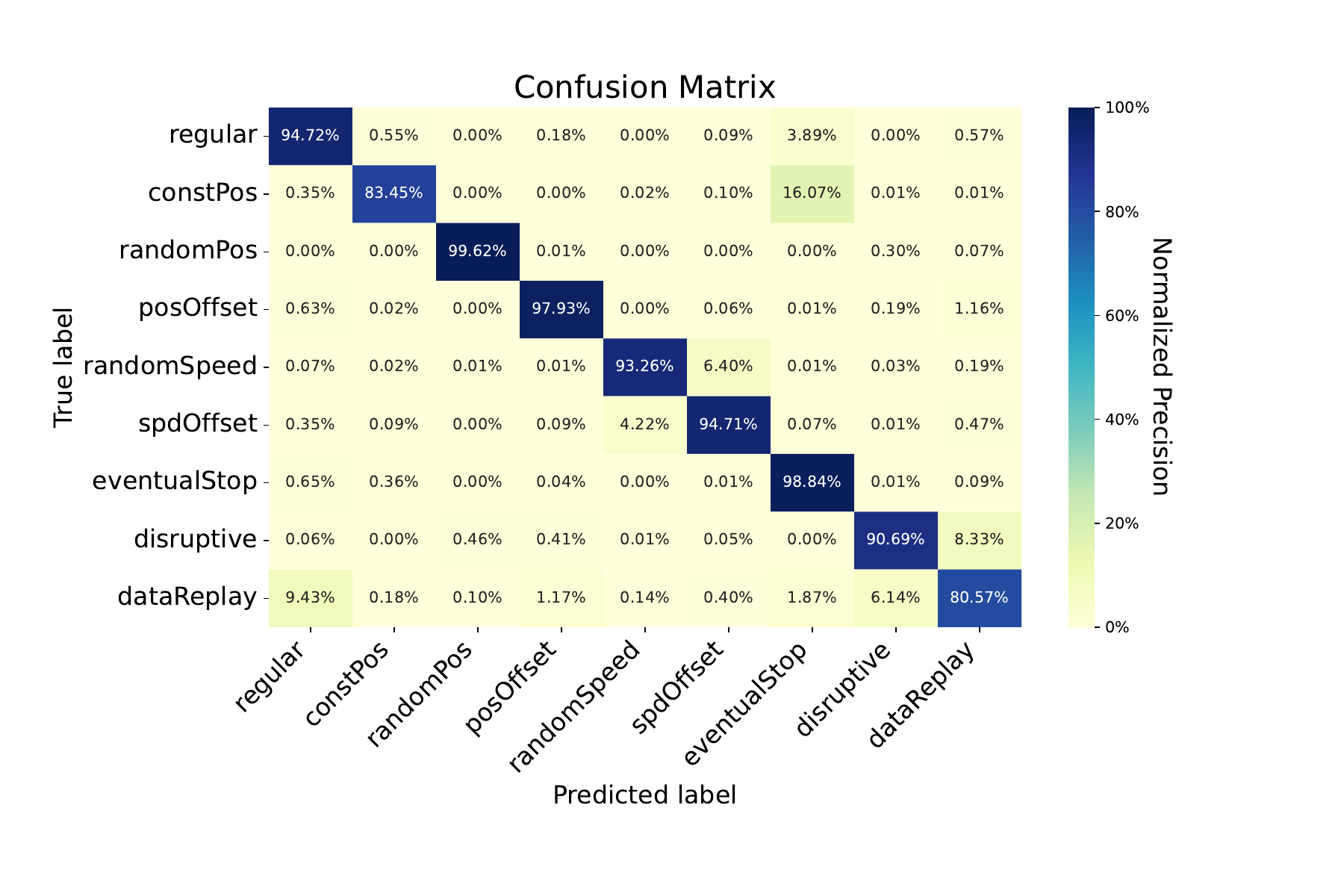}}
    \caption{Confusion matrix normalized by row of the model trained on each label.}
    \label{fig:conf-matrix}
\end{figure}

The last plot shown, in \cref{fig:conf-matrix}, is meaningful to understand better how the misbehavior are predicted using multiple labels. In particular, this plot
reports a standard confusion matrix with values normalized by the sum of predictions per each category (i.e., per each row), leading to a metric that we call "Normalized Precision". In the figure, it's possible to see how the regular label has a very good prediction, and every label is generally well predicted. The biggest problems are with between the labels \textit{constPos} and \textit{eventualStop}, which are more than sometimes confused; since the deltas used to train the network are logically very close, it's normal that this confusion happens, but it's not a problem during the simulation, because in the thesis environment is enough to intercept any kind misbehavior to start the protocol. Another logical confusion is between data replay and disruptive. As it's said those attacks are logically similar, but another time, for the simulation is not a problem. The only problem in the simulations could be between \textit{dataReplay} and the \textit{regular} label, which, compared to the other one, shows some confusion problems.

    \chapter{Simulation environment}
\label{ch:simuenv}
The simulations are realized thanks to the coordinated use of different tools. In particular, a well-known simulator infrastructure is used to emulate urban traffic and the \ac{VN}.
OMNeT++ \cite{omnet++} is a well-known discrete event simulator that can replicate a network. 
In particular, it can provide a model of block programming similar to the ISO/OSI stack, so it's particularly easy to design cross-layer network protocols. For the traffic, \ac{SUMO} \cite{sumo} is chosen because it can be easily coupled to OMNeT++ through Veins \cite{sommer2011bidirectionally}, then extended by PLEXE \cite{segata2014plexe} to provide all the platooning functionalities.
Thanks to this structure, it is possible to implement a set of \ac{VN} simulations, including the possibility of designing some particular misbehavior. Also, the design of a management protocol can be implemented across those tools.

\section{SUMO}
\ac{SUMO} \cite{sumo} is an open-source, highly scalable microscopic traffic simulator developed in Germany at the Transport Research Institute within the German Aerospace Center. It allows for the simulation of traffic composed of individual vehicles moving through a road network. The simulation addresses a wide range of traffic management topics and is purely microscopic, meaning that each vehicle is explicitly modeled, has its route, and moves individually through the network. The simulations are deterministic by default, but there are various options to introduce randomness.

SUMO includes many modules and programs to prepare and execute a traffic simulation. The two most used programs are sumo and sumo-gui. The former allows for simulating without visualization, using only the command line, and is often used together with \ac{TraCI}; the latter allows traffic simulation and visualization through a graphical interface.

\ac{TraCI} provides access to an ongoing road traffic simulation to external applications, allowing them to retrieve the values of simulated objects and manipulate their behavior on-line. The data exchange between SUMO and OMNeT++ utilizes this interface and is based on a TCP connection that encapsulates messages composed of sets of commands and responses. \ac{TraCI} is essential in simulations as it allows assigning a different controller for each vehicle, setting and changing their parameters, and even creating complex scenarios with preset accelerations and decelerations.

\section{Veins}

\begin{figure}[H]
    \centering
    \includegraphics[width=1\textwidth]{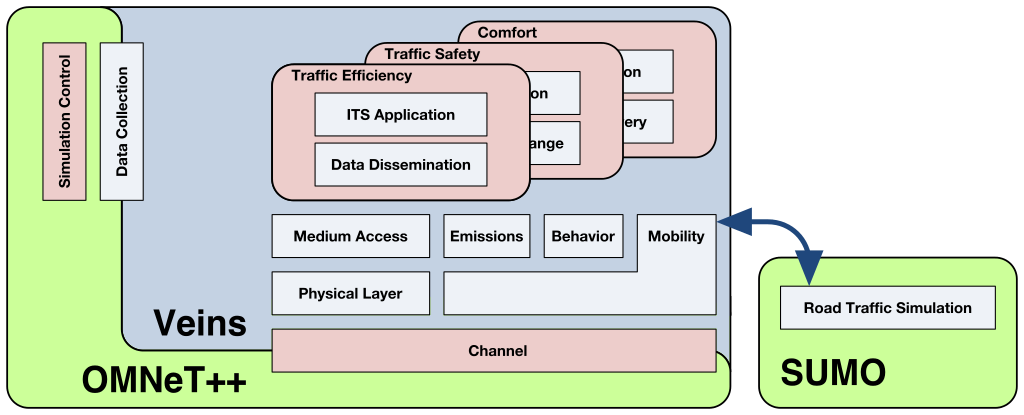}
    \caption{Coupling between OMNeT++ and \ac{SUMO} through Veins (https://veins.car2x.org).}
    \label{fig:veins-arch}
\end{figure}

Veins \cite{sommer2011bidirectionally} extends OMNeT++ by providing a vehicular communication stack based on \ac{IEEE} 802.11p. It can couple the network and mobility simulator, as shown in \cref{fig:veins-arch}, by creating a network node in OMNeT++ for each vehicle in \ac{SUMO}. Each node is associated with a stack that includes an \ac{IEEE} 802.11p network interface, a beaconing protocol, and one or more applications running on it.
Veins replicates the movement for the corresponding node in OMNeT++ whenever a vehicle moves. The two simulators, network and traffic, are extended with a module dedicated to communication. During the simulation, the two modules exchange commands through a TCP connection. The set of commands and responses exchanged is based on \ac{TraCI}, the interface exposed by \ac{SUMO}.

OMNeT++ is an event-based simulator that schedules each node's movements at regular intervals. This approach fits well with \ac{SUMO}, which advances simulation time in discrete steps. The modules integrated into OMNeT++ and \ac{SUMO} act as a buffer, collecting all commands between two-time steps and ensuring synchronization between the two simulations.
Veins, using \ac{TraCI}, queries \ac{SUMO} about the traffic state, asking for information such as the number of vehicles, their positions, and speeds for each of them, and can modify the traffic dynamics by changing a vehicle's route or its acceleration.

\section{Plexe}
PLEXE \cite{segata2014plexe, segata2022multi-technology} is an extension of Veins that enables realistic simulations and studies of platooning. Utilizing Veins' capabilities to simulate both the network and traffic, PLEXE integrates all the necessary components to study car platoons, such as control system models and maneuvers for forming and maintaining platoons.

PLEXE extends the interaction between SUMO and OMNeT++ via \ac{TraCI}, allowing vehicle data to be sent to all other cars in a platoon, thus enabling platooning protocols and applications. This allows the \ac{CACC} in \ac{SUMO} to be powered using data received from vehicles via Veins.
Modifications and extensions are required from two perspectives to enable car platoon simulations: on the \ac{SUMO} side, regarding the control system models and platoon maneuvers, and on the Veins side, for implementing platooning protocols and applications.

\subsection{SUMO control models}
\ac{SUMO} features several vehicle movement models designed to mimic human driver behavior. Examples include IDM and Krauss. The primary modification introduced by PLEXE is the new control model called CC (Cruise Controllers), which is based on Krauss. This movement model was specifically created to implement cooperative driving control systems. For example, Cruise Control and \ac{ACC} are part of this model.

The version of \ac{SUMO} used includes, in addition to CC and \ac{ACC}, the PATH \cite{rajamani2000path} and Pleog \cite{ploeg2011design} models. During the simulation, choosing which system controls the vehicle is possible. Indeed, through the \ac{TraCI} interface, one can access the models implemented by PLEXE. When a \ac{CACC} is active for a vehicle, it is possible to send it the necessary data, exchanged via IVC from the communication network simulator, OMNeT++.

\subsection{Protocols and applications in Veins}
\label{s:prot-veins}
Regarding Veins, each vehicle is equipped with an IEEE 802.11p network card, a basic protocol for sending messages, and an application for distribution. The idea is to have a protocol layer rooted in the BaseProtocol class, which provides all the basic functionalities to be extended, and a set of subclasses that focus solely on implementing the beaconing strategy.

The same principle is used for the application layer, where BaseApp is responsible for loading simulation parameters and passing data to the \acp{CACC} via \ac{TraCI}. The underlying layers are tasked with deciding a platoon application aspects, such as which vehicle is the leader, the lane a car should transit on, and much more.

\subsection{Defense application}
Given the Plexe/Veins structure, composed of protocols and applications, it's important to underline that the latter is responsible for the reading of each beacon. It's possible to create a new application by extending BaseApp, which is the basic one and that lets the most basic platooning examples. In this thesis, it's necessary to introduce a new logic because, for every five beacons, as explained in \cref{s:embedding}, it's necessary to perform a prediction that can be avoided in case of defense is turned off. In addition to that, it's also necessary to introduce a Finite State machine protocol across the predictions to manage the misbehavior detection during a simulation. So a new application class is created and called \textit{AIDefenseApp}, it extends the basic application, in particular overriding the \textit{onPlatooningBeacon} method, responsible for the beacons' reading.

From the extension of the basic application, it's important to focus on the new parameters defined. In particular, two parameters are defined, the first one is called \textit{AIdefenseEnabled}, and it's a boolean, that defines if the defense protocol is enabled or not, and the second one is \textit{radar} that stands for the usage of the radar information instead of the beacons' ones. In particular, \textit{radar} is always set to false during the beginning of the simulations since the protocol defines the condition on which to turn it on.
\begin{lstlisting}[language=c++]
*.node[*].appl.AIdefenseEnabled = {true, false}
*.node[*].appl.radar = false
\end{lstlisting}

About the protocol, the base one is still used but with the addition of new parameters and functions.
The first new parameter is inserted into the beacons, and it can be turned on during the simulations. This parameter is simply called \textit{warning}, and it stands for when a misbehavior or an attack is detected.
\begin{lstlisting}[language=c++]
bool warning;
\end{lstlisting}
While for what concern the new functions, \textit{setWarning}, to activate the warning into the beacons, and \textit{activeAttack}, to activate misbehavior or an attack, are created.
\begin{lstlisting}[language=c++]
// set the warning in sent beacons
void setWarning(bool misbehaviour);
// set the onAttack variable
void activeAttack(const char* type);
\end{lstlisting}
All the details will be explained later in \cref{ch:method} with the definitions of the protocol and the attacks to be detected.
     \chapter{Evaluation Methodology}
\label{ch:method}

The simulators presented in \cref{ch:simuenv} are mostly written in C++ language. To use the \ac{MDS}, based on a Python trained \ac{NN}, is necessary a link between the simulations environment and the model of the \ac{NN}, saved in Keras \textit{h5} format. The c++ library \textit{pybind11} for the binding between C++ and Python is used to achieve the result presented in \cref{s:nn-bind}.

After allowing the usage of the detector through the simulation structure, to properly evaluate the \ac{MDS}, implemented in \cref{ch:misbeh}, on-line and installed in every vehicle, it is crucial to implement all the misbehavior observed in the simulated scenarios. Considering the significantly different scenario used, presented in \cref{ch:sim}, compared to the dataset's simulations, it's important to underline that, first of all, the data produced are different, e.g., if an extra-urban scenario is considered, the speed and acceleration deltas could be significantly different compared to a urban one. Moreover, the misbehavior's code could not be precisely the same as the dataset, and it's also the purpose of this thesis to deal with misbehavior that could be slightly different from the learning ones. Also, the misbehaving feature interval, like the random offset selected for the offset scenario, could be represented differently, adapting it to the scenario created and logical considerations further explained in \cref{s:misb-def}. 

The last step is then the design of the safe management defense protocol, presented in \cref{s:protocol}. This one is intended to work on the prediction of the \ac{MDS}, and since there's a lack in developing and simulating on-line protocols that work along with \acp{MDS}, it's important to define a solution that could be a standard for future implementation while reaching great results as it stands. 
So the solution presented is slightly simple and works only on the differentiation between misbehavior (all the labels that are not regular) and regular, i.e., what is defined as single label. So, all misbehaving multiple labels are not necessary, but they are still studied to allow future complex protocols that can differentiate every single attack.

\section{Neural Network binding}
\label{s:nn-bind}
As defined in \cref{ch:misbeh}, a keras model and a standard scaler are designed and trained on the \ac{VeReMi} dataset as it is modified in \cref{s:feat_eng} and \cref{s:embedding}, and ready to be used in a simulation. The model is saved in \textit{h5} format, while the standard scaler is in \textit{pkl}. Since the \ac{NN} definition is Python-made, and all the simulation environment is C++ based, it's necessary a binding to use the scaler or the prediction saved at each beacon received. To reach this goal the c++ library \textit{pybind11} is used. In particular, this library is efficient compared to using the \textit{system} library and calling through the system the Python model. Since the prediction is needed every five beacon times, which means 0.5 s considering the further used beacon time, a good efficiency is necessary. 

Starting from the decision to use this particular library, the definition of a class that can maintain the state of the model and the scaler, manage all the predictions, and save all the messages is necessary. In particular, it is defined a new package in Plexe, called \textit{AI}, and into this package, a new c++ class is implemented, called \textit{NNWrapper} that stands for Neural Network Wrapper.

\subsection{Definition and Initialization}
In order to store a message, the first meaningful definition is a struct called \textit{Message} that can contain all the information that is needed to compute the prediction based on the \ac{NN} trained. This structure is defined to save the information of the first message, that is the reference point to compute the normalization of the next four messages, as explained during the model's training in \cref{s:embedding}.
\begin{lstlisting}[language=c++]
struct Message 
{
double dt;
double posx;
double posy;
double spdx;
double spdy
double acl;

Message(double t, double x, double y, double spx, double spy, double a)
    : dt(t), posx(x), posy(y), spdx(spx), spdy(spy), acl(a) {}
};
\end{lstlisting}
In particular, the data saved, as defined in \cref{ch:misbeh}, are position x and y, speed x and y, acceleration, and a delta time called dt. The struct is created along with its constructors in order to create a message with useful data.
The class is defined with some others useful data structures, to save the messages and the manipulated data. 
\begin{lstlisting}[language=c++]
// Map to store the first message of each vehicle
std::map<int, Message> firstMessageMap;

// Map to store the scaled differences for each vehicle
std::map<int, std::map<int, std::vector<double>>> scaledDifferencesMap;
\end{lstlisting}
In particular, a map called \textit{firstMessageMap} is used to save the first messages of the group of five, along with the index of the sender's vehicle, and a map of map called \textit{scaledDifferenceMap} that is used to save all the differences between the messages from the second to the fourth and the first one, also along with the indexes. Those data structures are intended to store the data already scaled in order to speed up the simulation since the computation of the standard scaling is made at each beacon step and not at the end of the receiving of the five messages. 

During the initialization of the application, that is running on a vehicle, the constructor of the \textit{NNWrapper} class is called. In this class, there are three Python objects that are fundamental to be reached in all the class code, and these are the model, the scaler, and the pandas' library object, in order to allow the usage of the data frames to better manage the data before using the scaler. In all the class code, \textit{py} stands for \textit{pybind11}
\begin{lstlisting}[language=c++]
py::object model;
py::object scaler;
py::object pd;
\end{lstlisting}
A path to reach the model and the scaler saved is given, and thanks to the \textit{pybind11} library, the objects are created and saved along with the creation of the class and the import of all the necessary Python libraries. Here is the code of the module imported to use all the Python stuff.
\begin{lstlisting}[language=c++]
py::module_ keras = py::module_::import("tensorflow.keras.models");
py::module_ pickle = py::module_::import("pickle");
py::module_ builtins = py::module_::import("builtins");
\end{lstlisting}

\newpage
\noindent While, in the following code snippet is presented the initialization of the model and the scaler given the \textit{model_path} and the \textit{scaler_path}, using the needed libraries previously imported.
\begin{lstlisting}[language=c++]
// Load the model h5 format
model = keras.attr("load_model")(model_path, py::arg("compile") = false);

// Load the scaler
py::object scaler_file = builtins.attr("open")(scaler_path, "rb");
scaler = pickle.attr("load")(scaler_file);
scaler_file.attr("close")();
\end{lstlisting}

\subsection{Workflow}

\begin{figure}[H]
    \centering
    \includegraphics[width=1\linewidth]{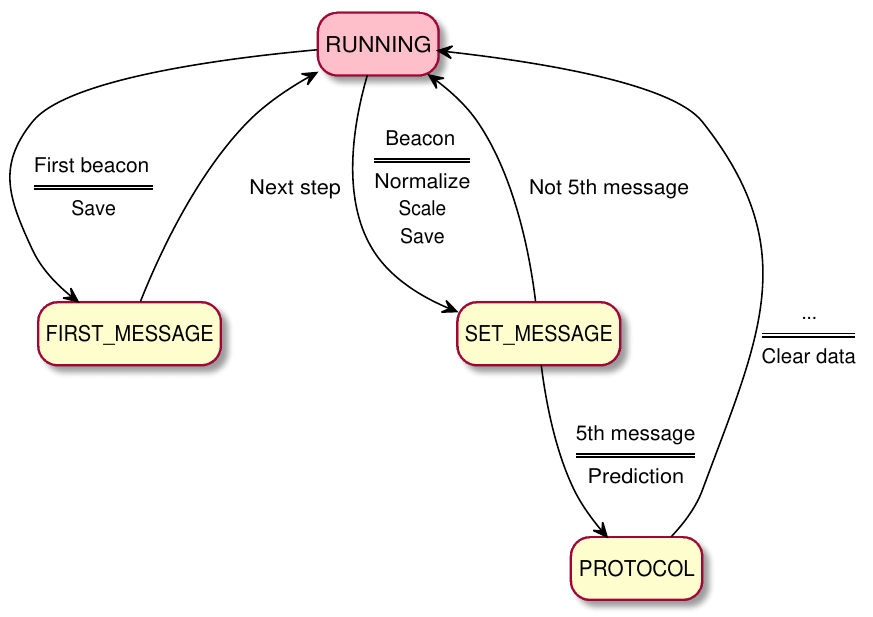}
    \caption{Finite State Machine that represents how during a simulation predictions and scaling are computed, and when the defense protocol is triggered.}
    \label{fig:prediction}
\end{figure}

During the simulation, vehicles exchange beacons with close vehicles on the road or within the same group, such as a platoon. The usefulness of received beacons depends on the control system in use. For example, if the Ploeg control system is used, the front vehicle messages are the only useful beacons. The useful messages received are processed to obtain four scaled and differentiated data entries, which are then used for prediction computation, as defined in \cref{ch:misbeh}. In particular, if the message is the first, it's simply saved into the \textit{firstMessageMap} data structure, and then the simulation proceeds with the next step. 
\begin{lstlisting}[language=c++]
if(firstMessageMap.find(id) != firstMessageMap.end())
{
// not the first message ...
} else {
    Message msg(posx, posy, spd, acl, hed);
    firstMessageMap[id] = msg;
}

\end{lstlisting}
If the message is the second, third, or fourth one, the data normalization (all the operations needed, like computing the deltas) is made, and then the normalized data are scaled thanks to the usage of the Standard Scaler previously bound.   
\begin{lstlisting}[language=c++]
py::object normalized_data = scaler.attr("transform")(input_array);
\end{lstlisting}
Then if the message is the 5th, after the processing of the data, the prediction is computed.
\begin{lstlisting}[language=c++]
py::object prediction = model.attr("predict")(input_array);
\end{lstlisting}
Depending on the label predicted, the protocol works in a precise way as it is defined in \cref{ch:method}.
In \cref{fig:prediction} the working flow previously described is represented.

\section{Misbehavior definition}
\label{s:misb-def}

The misbehavior selected from the \ac{VeReMi} dataset, presented in \cref{s:selected-mis}, are used to train the \ac{MDS}. Each of these anomalies needs to be implemented using simulators to see if the \ac{NN} can detect them in an on-line simulation. In \cref{s:prot-veins}, it is stated that each individual car is equipped with protocols responsible for creating and sending beacons to other vehicles. As any misbehavior is indicated by a variance between the vehicle's on-road behavior and the data it transmits through \ac{V2X} communications, it is imperative to modify the protocols installed on every car in a way to insert all the misbehavior studied. Moreover, it's important to underline that each misbehavior is intended to mimic its respective one from the dataset, but due to the traffic scenario simulated, i.e., a platoon on a highway, as explained in \cref{ch:sim}, some parameters are modified. From now each misbehavior used to train the model is presented along with its implementation into the simulation's protocol.

\begin{itemize}
    \item \textbf{Constant position}: At the time \textit{t} of beginning of the constant position misbehavior, the selected misbehaving vehicle starts to transmit the same position of \textit{t}, here is the formulas, in which $t + i$ is intended as every time step after the start of the misbehavior:
    \begin{equation}
        pos_x[t + i] = pos_x[t]
    \end{equation}
    \begin{equation}
        pos_y[t + 1] = pos_y[t]
    \end{equation}
    Here is the code used to set the \textit{t} position into a function called \textit{activeAttack}, recalled when the simulation reach the time \textit{t}: 
    \begin{lstlisting}[language=c++]
if(strcmp(type, "constPos") == 0)
{
    VEHICLE_DATA data;
    plexeTraciVehicle->getVehicleData(&data);
    posx = data.positionX;
    posy = data.positionY;
}
    \end{lstlisting}
    The significant differences from the dataset misbehavior are that the anomaly start during the simulation and not from the beginning of it, and that the position sent is intended like a sudden stop of the \ac{GPS} functionalities and not a random position.
    \item \textbf{Random position}: When the misbehavior begin, the position features are set to a random value between  $[0 m, 10000 m]$, an interval that is deliberately very broad, to represent a \ac{GPS} completely broken. Here is the formulas:
    \begin{equation}
        pos_x[t + i] = U[0, 10000][m]
    \end{equation}
        \begin{equation}
        pos_y[t + i] = U[0, 10000][m]
    \end{equation}
    and the code used to set the positions:
    \begin{lstlisting}[language=c++]
if(strcmp(attackType, "randomPos") == 0)
{
    std::uniform_int_distribution<> distr(0, 10000);
    posx = distr(gen);
    posy = distr(gen);
}  
    \end{lstlisting}

    \item \textbf{Position offset}: At the time \textit{t} of the misbehavior the sending vehicle starts to create beacons with its actual position plus a random selected offset. Here is the formulas:
    \begin{equation}
    pos_x[t + i] = pos_x[t + i] + U[-10, 10][m]
    \end{equation}
    \begin{equation}
    pos_y[t + i] = pos_y[t + i] + U[-10, 10][m] \end{equation}
    and the code used to set the random offset:
    \begin{lstlisting}[language=c++]
if(strcmp(attackType, "randomOffset") == 0)
{
    std::uniform_int_distribution<> distr(-10, 10);
    offset = distr(gen);
}
    \end{lstlisting}
    The biggest difference from the dataset is that in there it's used an interval between $[-300 m, 300 m]$, for the random version of the offset, and a constant offset of $\Delta_X = 250 m$ and $\Delta_y = -150 m$, while here the offset is significantly smaller. This decision is because the purpose is to represent a small error of the \ac{GPS}, while including all the bigger intervals in the random scenario.

    \item \textbf{Random speed}: Here the concept is the same as the random position scenario, but for the speed features and with an interval of random selection between $[-200 m/s, 200 m/s]$, intentionally extreme speed values to indicate that the \ac{OBU} is not functioning properly, with the formulas like this:
    \begin{equation}
    speed_x[t + i] = U[-200, 200][m/s]
    \end{equation}
    \begin{equation}
    speed_y[t + i] = U[-200, 200][m/s]
    \end{equation}
    and the speed setting as it follows:
    \begin{lstlisting}[language=c++]
if(strcmp(attackType, "randomSpeed") == 0)
{
    std::uniform_int_distribution<> distr(-200, 200);
    spdx = distr(gen);
    spdy = distr(gen);
}        
    \end{lstlisting}
    \item \textbf{Speed offset}: The idea is the same as for the position offset, but with an offset interval between $[-8 m/s, 8 m/s]$, with the formulas as it follows:
    \begin{equation}
    speed_x[t + i] = speed_x[t + i] + U[-8, 8][m/s]
    \end{equation}
    \begin{equation}
    speed_y[t + i] = speed_y[t + i] + U[-8, 8][m/s]   \end{equation}
    and the code like this:
    \newpage
    \begin{lstlisting}[language=c++]
if(strcmp(attackType, "randomOffsetSpeed") == 0)
{
    std::uniform_int_distribution<> distr(-8, 8);
    offset = distr(gen);
}      
    \end{lstlisting}
    \item \textbf{Eventual stop}: This kind of attack is exactly as in the dataset, so all the position, speed and acceleration features are set to 0 once \textit{t} is reached. Here is the code snippet used to represent the misbehavior: 
    \begin{lstlisting}[language=c++]
if(strcmp(type, "eventualStop") == 0)
{
    VEHICLE_DATA data;
    plexeTraciVehicle->getVehicleData(&data);
    spdx = 0;
    spdy = 0;
    posx = 0;
    posy = 0;
    acl = 0;
}
    \end{lstlisting}
    \item \textbf{Disruptive}: This kind of attack is an information replay of previously received data from random neighbor. Since the scenario is significantly different from the urban one, the random neighbour is selected from the indexes of the platoon's vehicles at each step after the misbehaving time \textit{t}, and the selected vehicle's data are re-transmitted into the platoon, but with the misbehaving vehicle's sign. Here is the random index selection code:
    \begin{lstlisting}[language=c++]
std::uniform_int_distribution<> dis(-1, platoonSize-1);
    int randomNum;
    do {
        randomNum = dis(gen);
    } while (randomNum == myPos);
    appl->setReplayIndex(randomNum);
    \end{lstlisting}
    \item \textbf{Data replay}: This attack is similar to disruptive, but for data replay the replay index is selected only at the beginning of the simulation as for disruptive, but it remains the same through all the simulation. While the other misbehavior are similar to the ones presented in \ac{VeReMi} those 2 attacks could be very different, bringing some confusion into the multiple label on-line prediction of the \ac{MDS}.
\end{itemize}

\section{Protocol}
\label{s:protocol}

\begin{figure}[H]
    \centering \includegraphics[width=1\linewidth]{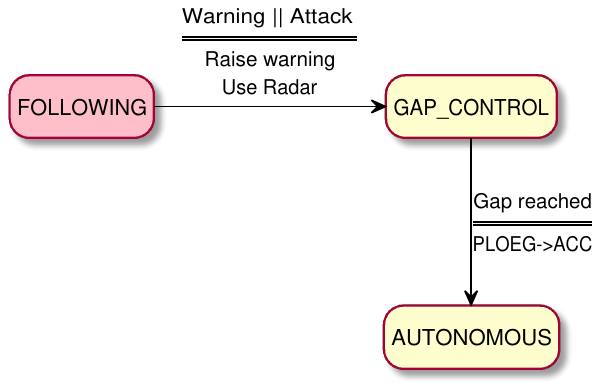}
    \caption{Finite state machine that represents the protocol of management for a misbehavior}
    \label{fig:fsm_prot}
\end{figure}

After implementing misbehavior into the simulations, the \ac{MDS} is ready to work on-line in an attempt to detect them. Upon detecting misbehavior, a protocol is necessary to manage this prediction and try to save as many vehicles as possible on the road. The literature reviewed in \cref{ch:related} lacks a significant development of a defense protocol that can operate based on \ac{MDS}'s prediction. The protocol presented in this section is simple and aims to establish a standard that can be modified in the future.
The protocol works on every vehicle decentralized, that means that the prediction of the \ac{MDS} is made independently from the other ones. In \cref{fig:fsm_prot} is presented a finite state machine that aims to represent the working flow of the protocol.

When a vehicle is operating in traffic and engaging with other cars, its state is called \textit{following}. However, if its \ac{MDS} detects any kind of misbehavior, referred to as the \textit{attack} variable, the vehicle's state changes. For this protocol, it is essential to focus on a single label; a deviation from the regular label triggers the state change. The \ac{MDS} prediction alone is not the sole trigger for the state change. Due to the ability of all nearby vehicles to communicate, a \textit{warning} parameter is embedded in each beacon. If another vehicle detects irregularities in the exchanged messages, it activates the defense protocol. This is critical as a vehicle only processes a fraction of the messages related to its control system and is unable to detect misbehavior from other cars. 

Once the state change is activated, the vehicle raises a warning by setting the \textit{warning} parameter of the beacons to true, signaling ongoing misbehavior, and switches to utilizing radar instead of V2X communications. Since this protocol only distinguishes between misbehavior and regular messages, it does not provide specific responses for each type of misbehavior. Consequently, in the case of severe misbehavior that could result in an accident, it is preferable for the vehicle to utilize radar data instead of communication data for its control system.

\begin{lstlisting}[language=Python, caption={Psuedocode of the protocol procedures}, captionpos=b, label={lst:pseudocode}]
procedure INITIALIZE()
    state <- FOLLOWING
    activeController <- PLOEG
    setControllerGap(headway[PLOEG], distance[PLOEG])
    warning <- False
    useRadar <- False

procedure ONPLATOONBEACON(pb)
    if (state = FOLLOWING) then
        prediction <- predict()
        if (prediction != 0 or pb->warning = True) then
            state <- GAP CONTROL
            warning <- True
            useRadar <- True
            START GAP CONTROL()

procedure GAP REACHED()
    state <- DOWNGRADE
    ctrl <- ACC
    setControllerGap(headway[ACC], distance[ACC])
    
\end{lstlisting}

The protocol aims to use radar data after detecting misbehavior, making support for \ac{CACC} unfeasible due to potential issues with the radar's provision of low-latency, high-precision data for real-time cooperative control. The fundamental concept of the protocol involves transitioning the vehicle's control system from cooperative to semi-autonomous (specifically \ac{ACC}) upon encountering misbehavior. This approach is reminiscent of the method used in reference \cite{segata2022multi-technology}, where the gap between vehicles is increased based on the number of non-functioning communication interfaces until a direct switch of the control system becomes available. Such a procedure is crucial to avoid a sudden transition between control systems with different desired distances, which could lead to an abrupt braking event and result in an accident. Throughout the phase of increasing the gap, the entire process is managed in the \textit{gap control} state, as if all communication interfaces had failed, as detailed in reference \cite{segata2022multi-technology}. Once the required gap is achieved, the control system is prepared to transition to semi-autonomous ACC, with the continued use of radar information during this phase.
The pseudocode of the entire process is detailed in the \cref{lst:pseudocode}. Here the \textit{initialize} procedure has to be intended as the beginning of the simulation, while the \textit{on platoon beacon} is the procedure called every time a beacon is received. Here the management of the group of five messages, explained in \cref{s:embedding}, is deliberately cut, because the focus is on the predicted value and not on how the prediction is computed. Moreover, all the gap control used to allow the fallback to a semi-autonomous system is explained in \cref{s:fallaback}.

\section{Fallback system}
\label{s:fallaback}

Following what it's said in the reference \cite{segata2022multi-technology}, to complete a fallback from a control system to another one, with a larger desired distance, a procedure that progressively increases the inter-vehicle distance is needed. The Gap Control Algorithm, fully presented in \cref{lst:gap}, is specifically designed to ensure the safe and seamless adjustment of the inter-vehicle distance within a platoon, particularly during transitions between different control systems. The algorithm begins with the start gap control procedure, as indicated in \ref{lst:pseudocode}, and by initializing the desired gap, i.e., \textit{gap_t} that stands for target gap, based on the current speed \textit{curSpeed} and specified headway time \textit{curHead}. It then determines whether the gap should be increased or decreased and adjusts the inter-vehicle distance incrementally to avoid uncomfortable or unsafe accelerations. The core of the algorithm involves periodic updates that adjust either the current time headway or the fixed gap distance, depending on the control policy in use. These updates continue until the target gap is achieved, at which point the algorithm verifies that the vehicle has physically reached the desired distance before signaling the end of the procedure. This careful management of gap adjustments helps maintain stability and safety in the platoon, even during transitions between different communication states or control algorithms. The algorithm is particularly valuable in scenarios where vehicles must adapt to varying communication conditions or need to fallback to a semi or full automatic control system to maintain driving safety, ensuring that the platoon can maintain safe distances under failures in the network or misbehavior.

\newpage
\begin{lstlisting}[label={lst:gap}, language=Python, caption={Gap Control Algorithm taken from \cite{segata2022multi-technology} and adapted for this thesis scope}, captionpos=b]
procedure STARTGAPCONTROL(head_t, dt)
    gap_t <- head_t * curSpeed + dt
    increasingGap <- true
    if using time headway then
        curGap <- dt
        curHead <- (curDistance - dt) / curSpeed
        if head_t < curHead then
            increasingGap <- false
    else
        curGap <- curDistance
        curHead <- head_t

    if gap_t < curGap then
        increasingGap <- false
    updateGap()

procedure UPDATEGAP
    delta_h <- delta_g / curSpeed
    gap_t <- head_t * curSpeed + dt
    if isGapControlCompleted() then
        if using time headway then
            curHead <- head_t
        else
            curGap <- gap_t
        setControllerGap(curHead, curGap)
    if isGapReached() then
        gapReached()
        return
    else
        if using time headway then
            curHead <- curHead + sgn(head_t - curHead) * delta_h * delta_t
        else
            curGap <- curGap + sgn(gap_t - curGap) * delta_g * delta_t
        setControllerGap(curHead, curGap)
    schedule(delta_t, updateGap())

procedure ISGAPCONTROLCOMPLETED
    if using time headway then
        if increasingGap then
            return curHead >= head_t
        else
            return curHead <= head_t
    else
        if increasingGap then
            return curGap >= gap_t
        else
            return curGap <= gap_t

procedure ISGAPREACHED
    if increasingGap then
        return curDistance >= gap_t
    else
        return curDistance <= gap_t
\end{lstlisting}

This code snippet provides a clear and structured way to implement the gap control algorithm in a \ac{CACC} system, ensuring that transitions between different states are smooth and safe.
In the pseudocode if a variable has the prefix \textit{cur} it stands for current, while if the suffix is \textit{_t} it stands for target. Moreover there are some variables like \textit{dt} that is a fixed distance offset, the rate at which the headway or gap should be adjusted represented by \textit{delta_t} and \textit{delta_g}. In the end the sign function is defined as $sgn(x) = 1$ if $x\geq0$, $-1$ otherwise.

\section{Gap control dynamic}

\begin{figure}[H]
    \centering \makebox[\textwidth][c]{\includegraphics[width=1.2\linewidth]{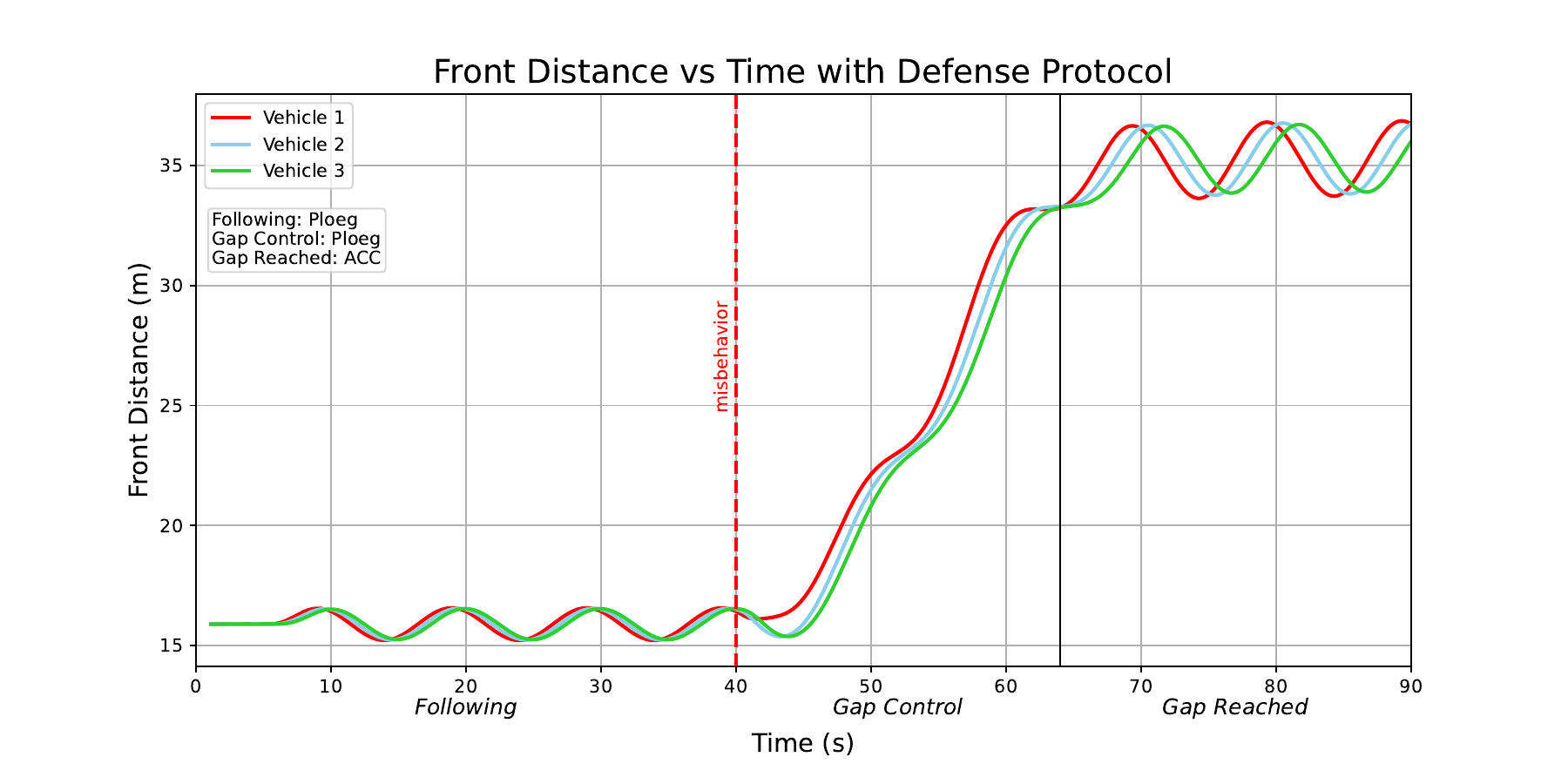}}
    \caption{Front distances vs time of a platoon under a sinusoidal speed pattern of the leader while dismantling after a misbehavior sent from the leader.}
    \label{fig:dynamic-dist}
\end{figure}

\begin{figure}[H]
    \centering \makebox[\textwidth][c]{\includegraphics[width=1.2\linewidth]{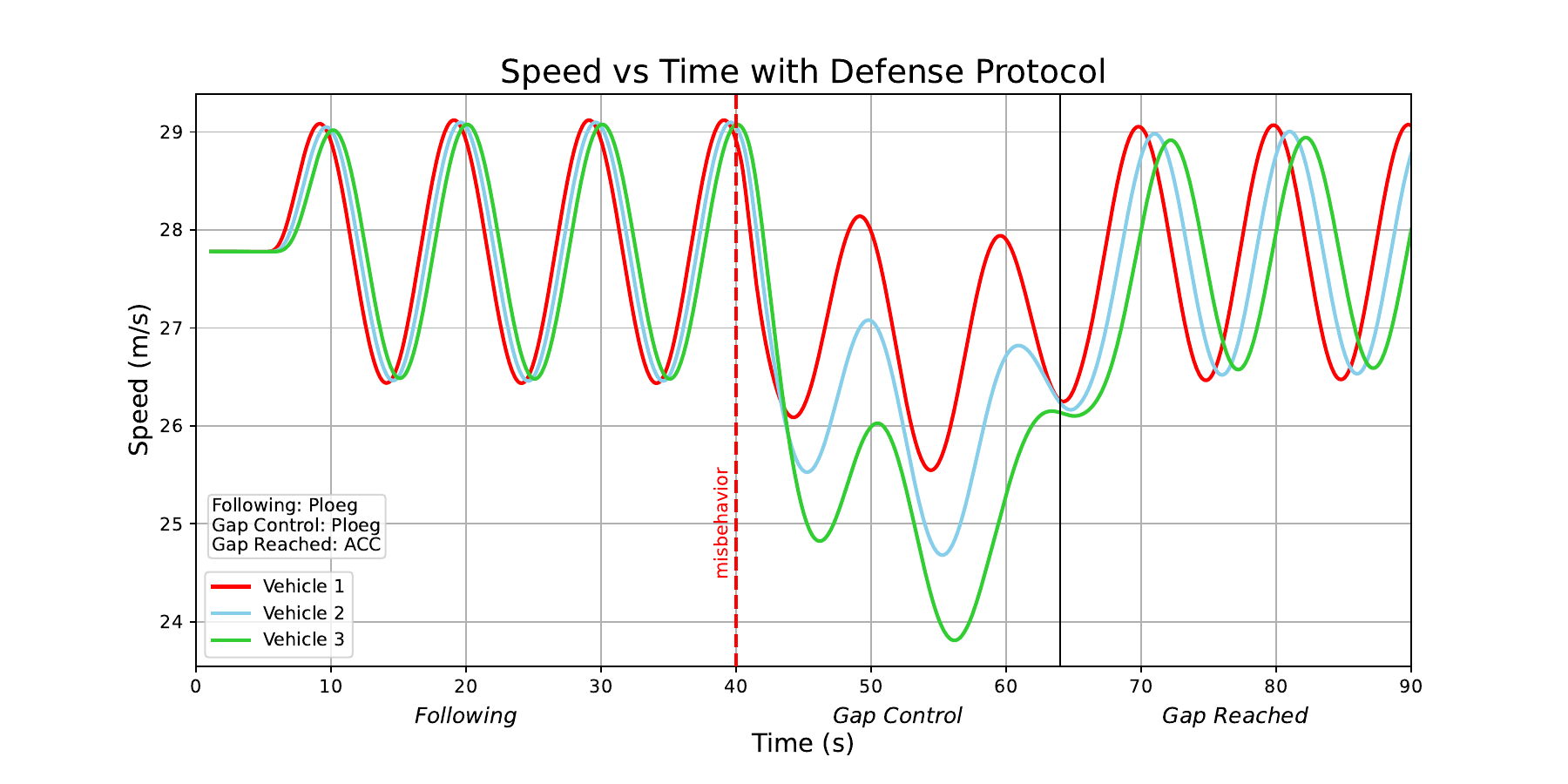}}
    \caption{Speed profiles vs time of a platoon under a sinusoidal speed pattern of the leader while dismantling after a misbehavior sent from the leader.}
    \label{fig:dynamic-spd}
\end{figure}

The fallback dynamic when the protocol starts to work, after a misbehavior detection, is shown in \cref{fig:dynamic-dist} and \cref{fig:dynamic-spd}. In this simulation, taken from the whole set presented in \cref{ch:sim}, the vehicles are using a Ploeg control system while following an oscillating speed pattern of the platoon's leader, used to imitate the normal flow of vehicles on the road, often not constant but turbulent due to traffic conditions. At 40 s of simulation the leader sends a misbehavior that is immediately detected from the \ac{MDS} of the vehicle behind that raises a warning and takes the protocol into action on all the platoon's vehicles.
After the detection it's possible to see how the defense protocol works while dismantling the platoon. \cref{fig:dynamic-dist}  shows the front distances of the following vehicles, showing how during the gap control state the distances are increased, while using the radar that still catches the oscillating leader's movement on the road, catch even when the gap is reached and the control system switched to \ac{ACC}. If instead of semi-autonomous the vehicles became completely autonomous, therefore without using radar, there would be no oscillation. While, in \cref{fig:dynamic-spd} is shown the speed pattern of the following vehicles, showing how after slowing down, they reconstruct the correct leader's pattern with the new semi-autonomous control system. Those plot show how the gap control phase to reach the correct distance to switch the control system uses less than 30 s, increasing the front distances of about 20 m, and without sudden break events.

    \chapter{Simulations}
\label{ch:sim}

The simulations are set on a highway, which differs significantly from the urban scenario based on the City of Luxembourg used in the dataset, in which every vehicle changes its neighbors more rapidly than on a straight highway. This variation, particularly in speed, acceleration, and transmitting vehicles, enables a run-time study of the \ac{MDS} in a new environment along with a saving protocol to assess the system's validity. Furthermore, the \ac{MDS} is validated within a platooning application, showing its use to prevent as much accidents as possible. Platooning is considered one of the most important goals in \ac{CITS} for highway scenarios due to its potential to significantly enhance safety, efficiency, and sustainability in road transportation and also the reference paper \cite{segata2022multi-technology}, on which the safe protocol is built, examined the fallback between control systems in a platoon scenario on a highway, it results to be a good choice on which validate the simulations' set.

\begin{table}[H]
\centering
\begin{tabular}{|l|l|}
\hline
\textbf{Parameter} & \textbf{Value} \\ \hline
Simulation Duration & 120 s \\ \hline
Control System & Ploeg \\ \hline
Scenario & Highway \\ \hline
Beacon Time & 0.1 s \\ \hline
Platoon Sizes & 4, 8, 16 \\ \hline
Leader Speed & 100 Km/h \\ \hline
Oscillation Amplitude & 5 km/h \\ \hline
Oscillation Frequency & 0.1 Hz \\ \hline
Starting Misbehavior Time & U[15 s, 80 s] \\ \hline
Repetition & 100 \\ \hline
\end{tabular}
\caption{Simulation Parameters}
\label{tab:simulation_parameters}
\end{table}

In this scenario, misbehavior are originated from a vehicle within the same platoon consisting of 4, 8, or 16 cars. In particular, each simulation is represented by the misbehavior type used and the platoon's vehicle that starts the misbehavior, that is only one for each simulation. Moreover, every simulation combination, represented by misbehavior ID and misbehavior type, is made with the defense protocol and the \ac{MDS}'s prediction activated and deactivated. So each simulation is uniquely identified by the combination of misbehavior vehicle ID, misbehavior type and defense protocol active or not. Each unique simulation is also repeated 100 times with a different seed and a different misbehavior starting time, randomly taken from a uniform integer distribution between 15 s and 80 s. All the simulation parameters are represented in summary in \cref{tab:simulation_parameters}.

Every simulation has a duration of 120 s, and in all the platoons simulated the control system used is Ploeg for all the vehicles in a platoon, unless for the leader, that is autonomous. An autonomous leader vehicle is required for a platoon, because it can maintain a consistent speed, acceleration, and braking pattern without the variability introduced by human drivers. This consistency is critical for maintaining a stable platoon, where the following vehicles can closely mimic the leader’s actions without experiencing sudden changes.
In particular, the autonomous leader uses a sinusoidal pattern of the speed, with a center value of 100 km/h, an oscillation amplitude of 5 km/h and a frequency of 0.1 Hz. This speed pattern is better through the thesis study, since it's normal on a road to have some little speed changes compared to a continue constant speed.
Moreover, since the simulations are dealing with \acp{CACC} at high speed values, a beacon time smaller than 1 second is crucial because it ensures that the vehicles can react quickly and smoothly to changes in the environment. This is why the beacon time is set to 0.1 s, a commonly used value in vehicular communication systems, particularly in \ac{CITS}.

All the detailed simulations are resumed in \cref{tab:simulation_details}. It has to be noticed that for the platoon size 4 and 8 the misbehavior that begins from the last vehicle ID is not simulated, since Ploeg only take care of the front messages, and if the last vehicle send a misbehaving message, no one would be affected by its misbehavior. While, for the platoon size 16, due to the time demanding simulations and the platoon sizes 4 and 8 that are sufficient for this thesis evaluation, only the leader and the center vehicle, with ID 7, are simulated to send a misbehavior. So, on the platoon size 16 is made only an evaluation to confirm what is seen in the others platoon sizes scenarios. In total the simulations made are 19200.

\begin{table}[H]
\centering
\begin{tabular}{|l|l|l|l|l|l|}
\hline
\textbf{Size} &
\textbf{ID} & \textbf{Misbehavior Type} & \textbf{Defense} & \textbf{Reps} & 
\textbf{Tot} \\ \hline
4 & [0, 1, 2] & \makecell{[constPos, randomPos,\\ posOffset,randomSpeed,\\ spdOffset,eventualStop,\\disruptive, dataReplay]} & [on, off] & 100 & 4800\\ \hline
8 & \makecell{[0, 1, 2, 3,\\ 4, 5, 6]} & \makecell{[constPos, randomPos,\\ posOffset,randomSpeed,\\ spdOffset,eventualStop,\\disruptive, dataReplay]} & [on, off] & 100 & 11200\\ \hline
16 & [0, 7] & \makecell{[constPos, randomPos,\\ posOffset,randomSpeed,\\ spdOffset,eventualStop,\\disruptive, dataReplay]} & [on, off] & 100 & 3200\\ \hline
\end{tabular}
\caption{Simulations summary}
\label{tab:simulation_details}
\end{table}

\section{Results}
\label{s:res}
The results are evaluated from two different perspectives. The most important one involves the prediction accuracy of the single label, presented in \cref{s:res_s}, in which every misbehavior is generally identified with the label \textit{misbehavior}. At the same time, a not misbehaving group of messages is called \textit{regular}. Since the designed protocol, presented in \cref{s:protocol}, works only on the differentiation between misbehavior or not, those are the most important results to be evaluated to catch if the \ac{MDS}, built on the \ac{VeReMi} dataset, works well with the defense protocol in the simulated scenarios. 

The designed protocol is built without a personalized response for each misbehavior, so if the \ac{MDS} identified each label well, the protocol could be modified with specific responses. To evaluate this possibility, the same results are also presented for the multiple labels evaluation, in \cref{s:res_m}, in which the metrics are evaluated by trying to identify every kind of misbehavior. The offline results, in \cref{s:perform}, present good values, but since the \ac{NN} implemented is simple as well as the scenario developed is different from the dataset one, the expected values are not as high as for the single label. Obtaining average good results could mean that the solution could be adopted with an improvement in the \ac{NN}.

The most important value to be shown is the accuracy of the prediction of the misbehavior label. Reminding \ref{s:perform}, the accuracy is a measure of the overall correctness of the model. For the evaluation of the results the formula has been rewritten and has to be intended as it follows:
\begin{equation}
    \text{Accuracy} = \frac{\#Correct\ Predictions}{\#Total}
\end{equation}
For the simulations' set the accuracy is evaluated from the time that a misbehavior begins. In particular, a prediction is considered correct, for the single label, if a misbehavior is predicted within two jumping windows, that means with a reaction time of maximum 1 s. Even if the misbehavior is predicted after those windows the prediction is considered not correct. For the multiple labels, the prediction is considered correct if the first misbehavior prediction within the 2 windows is the respective label of the misbehavior scenario. While, the denominator of the fraction depends on the grouping evaluated, e.g if the results are grouped for each misbehavior type, the accuracy is a value for a specific misbehavior and the total number is all the simulations in the set with that specific misbehavior.
Moreover, the results are shown within a confidence interval with a confidence level of 95\% and its mean, maximum and minimum. The confidence interval gives a range in which expect the true value to fall, and it quantifies how sure to be about that range. Furthermore, the average is not presented as a single value on all the simulations, but to deeply understand what it's happening, and which scenario creates more problems, it is presented first grouping all the simulations by kind of misbehavior, and then also grouping the simulations of a specific platoon size by the vehicle's ID that is sending the misbehavior.

Another important value to be shown, as said in \cref{s:structure}, is the labeling accuracy before that a misbehavior truly happens. This is important to show how many regular labels are confused with the misbehavior one and trigger the protocol before that is really necessary, blocking the normal flow of traffic. In the end, the last value to be analyzed is how many accidents the protocol is able to save.

\subsection{Single label results}
\label{s:res_s}

\begin{figure}[H]
    \centering
     \makebox[\textwidth][c]{\includegraphics[width=1\linewidth]{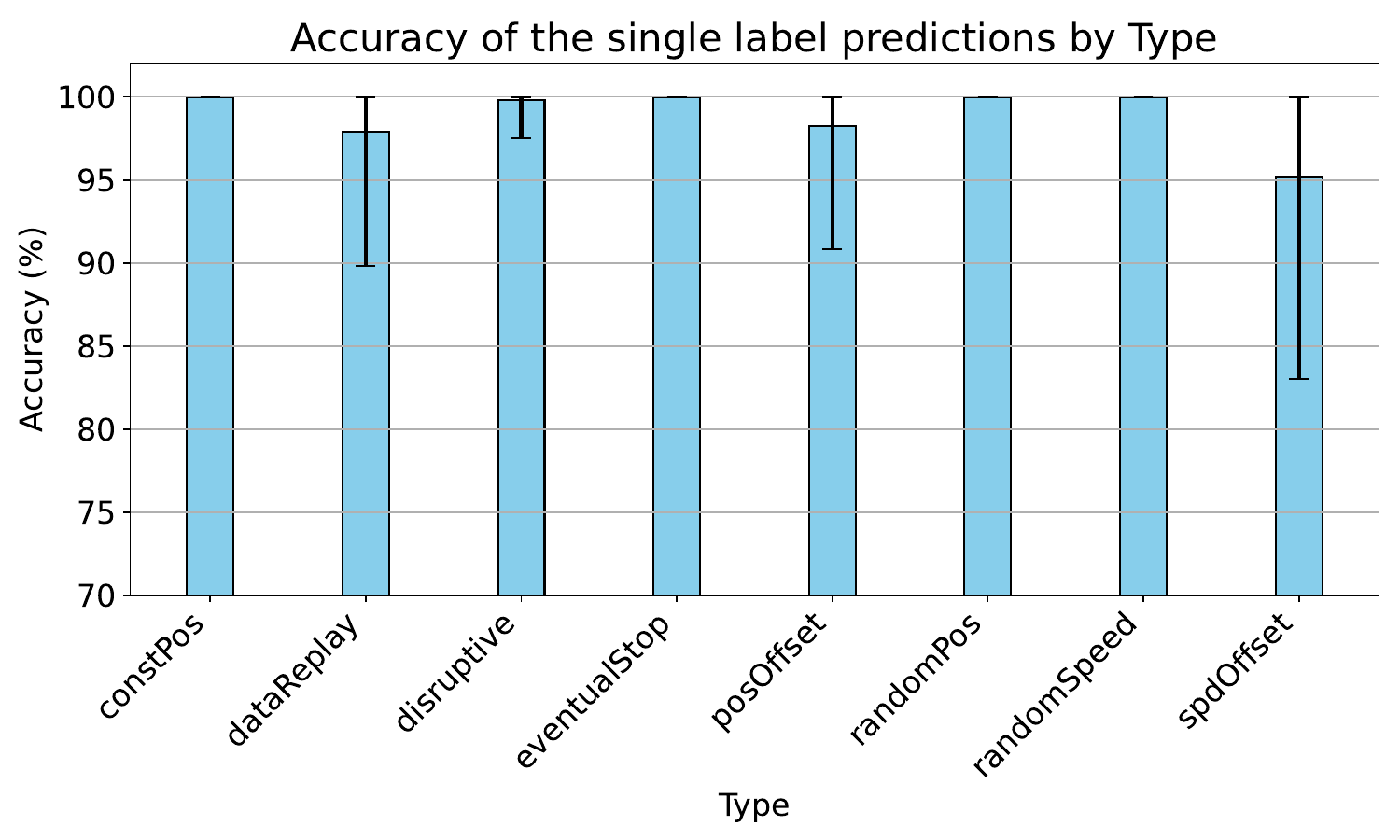}}
    \caption{Prediction accuracy of single label with the confidence interval, evaluated by each kind of misbehavior}
    \label{fig:av_acc_single_type}
\end{figure}

As already said, the single label intends to identify any misbehavior with a unique label that is differentiated from the regular one. In \cref{fig:av_acc_single_type} is shown the average accuracy with the maximum and minimum values of the confidence interval, with a confidence level of 95\%. In particular, this value is shown for each kind of misbehavior, e.g., all the constant position misbehavior, with every misbehaving IDs possible and with every platoon sizes possible, is represented in the "constPos" column. As this section presents the single label results, the prediction is marked as correct when a specific kind of misbehavior begins, if it is identified from the \ac{MDS} as the correct label or as any other kind of misbehavior. The only incorrect predictions are when a misbehavior begins, and it is identified as regular.
\begin{table}[H]
\centering
\begin{tabular}{|l|l|}
\hline
\textbf{Type} &
\textbf{Mean}  \\ \hline
constPos & 100\%\\ \hline
randomPos & 100\%\\ \hline
posOffset & 98.25\%\\ \hline
randomSpeed & 100\%\\ \hline
spdOffset & 95.17\%\\ \hline
eventualStop & 100\%\\ \hline
disruptive & 99.83\%\\ \hline
dataReplay & 97.92\%\\ \hline
\end{tabular}
\caption{Accuracy mean value of the single label grouped by misbehavior type}
\label{tab:mean_single_type}
\end{table}
\noindent In \cref{tab:mean_single_type} is possible to see the punctual average values of \cref{fig:av_acc_single_type} resumed, and it's possible to notice how the average accuracy of many misbehavior are 100\% or close to it. The only misbehavior that seems to create problems, with an average score less than 99\%, are the offsets and data replay. The lower offsets' score could be logical since the deviation from a regular behavior is defined as very small in \cref{s:misb-def}. At the same time, data replay could be the real problem for protocol safety. It has to be noticed that the maximums of every interval still reach 100\%. By doing a mean on the average score, the \ac{MDS} accuracy score on the misbehavior is 98.9\%.

In the following figures, accuracy is shown grouped by each vehicle ID that causes misbehavior for every platoon size available, i.e., 4, 8, and 16. In this case, a column represents the average prediction accuracy, along with the confidence interval, of every kind of misbehavior in a specific platoon size for the particular vehicle ID.

\begin{figure}[H]
    \centering
     \makebox[\textwidth][c]{\includegraphics[width=1\linewidth]{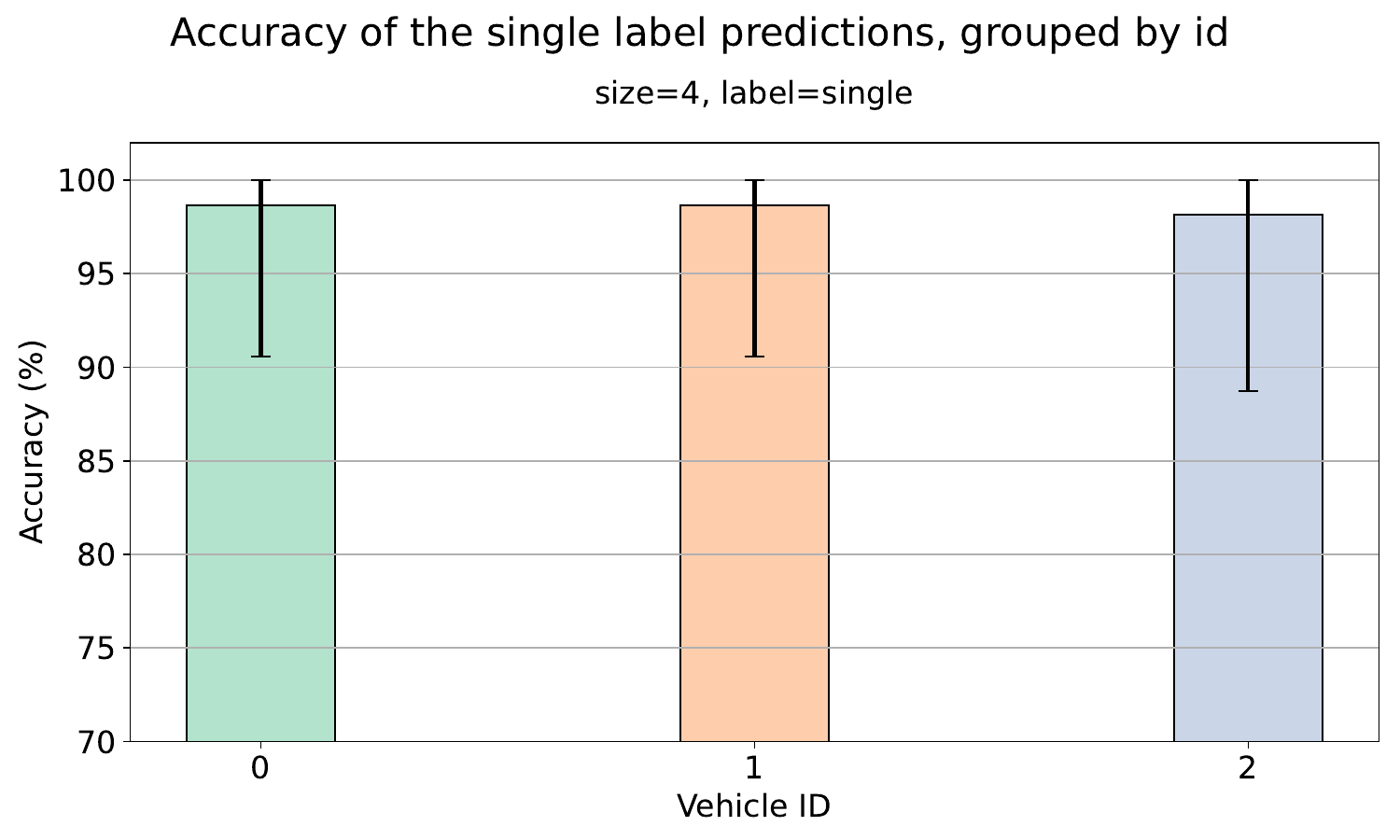}}
    \caption{Prediction accuracy of single label with the confidence interval, evaluated by each vehicle ID in a platoon of size 4}
    \label{fig:av_acc_single_id_4}
\end{figure}

\begin{figure}[H]
    \centering
     \makebox[\textwidth][c]{\includegraphics[width=1\linewidth]{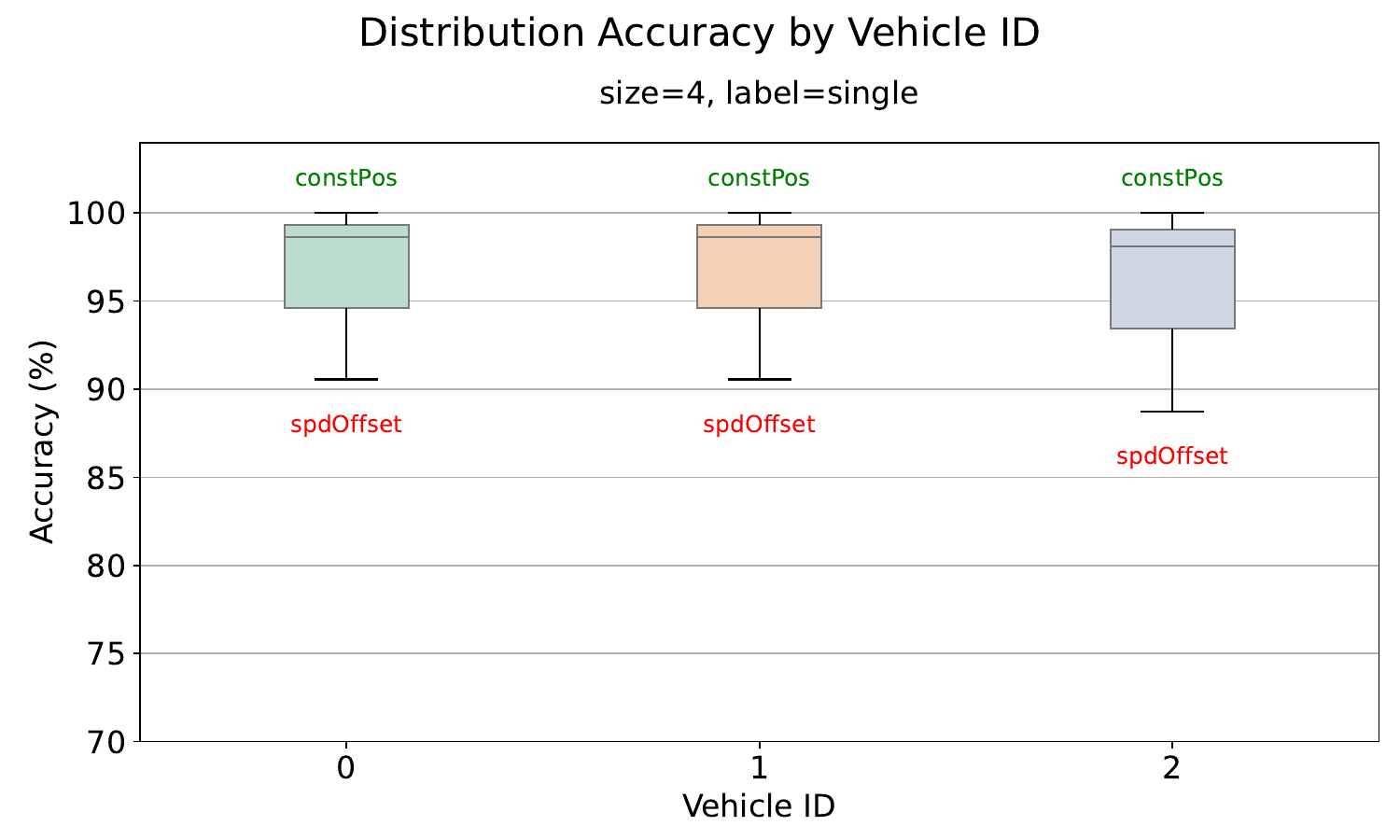}}
    \caption{Prediction accuracy distribution of single label, evaluated by each vehicle ID in a platoon of size 4}
    \label{fig:distr_acc_single_id_4}
\end{figure}

\begin{figure}[H]
    \centering
    \makebox[\textwidth][c]{\includegraphics[width=1\linewidth]{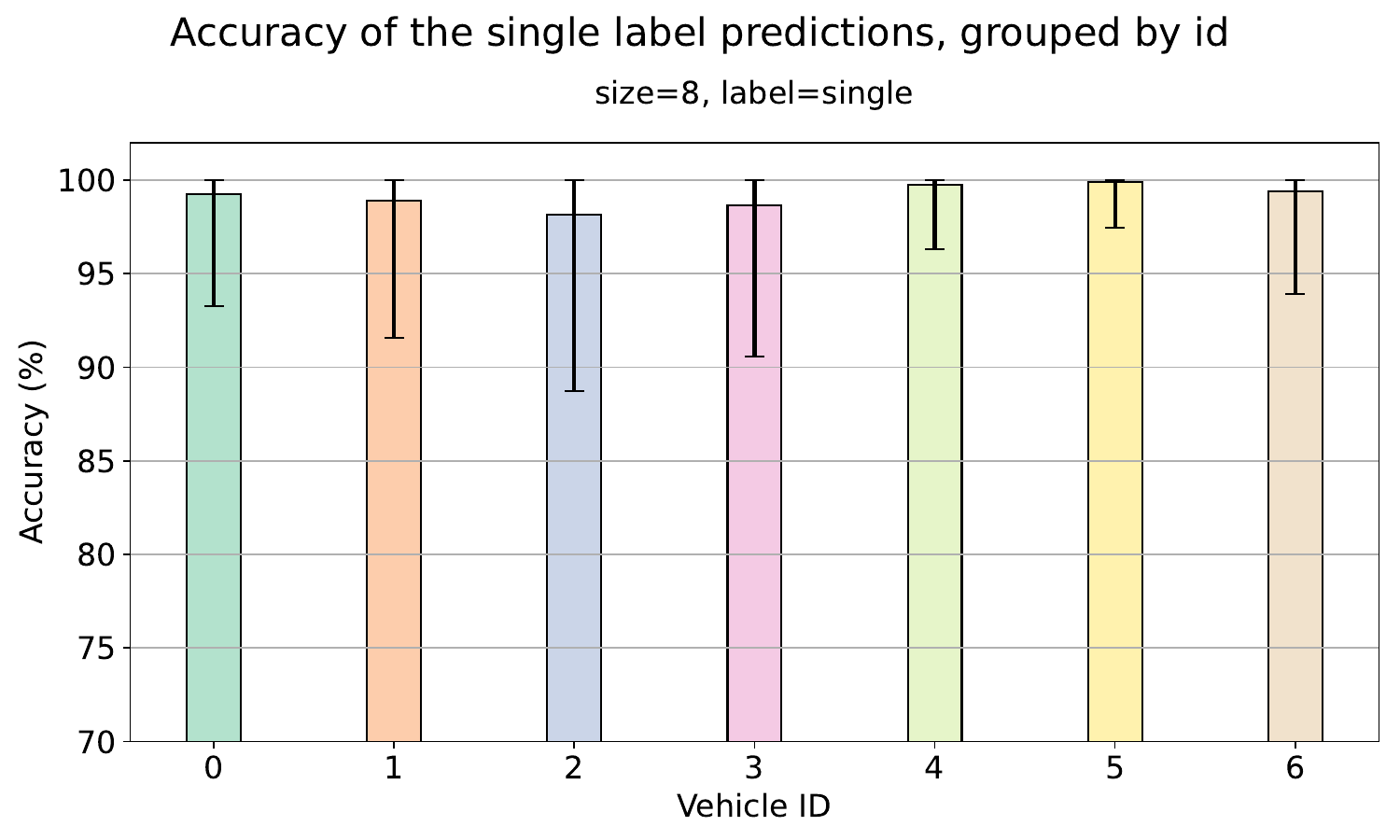}}
    \caption{Prediction accuracy of single label with the confidence interval, evaluated by each vehicle ID in a platoon of size 8}
    \label{fig:av_acc_single_id_8}
\end{figure}

\begin{figure}[H]
    \centering
     \makebox[\textwidth][c]{\includegraphics[width=1\linewidth]{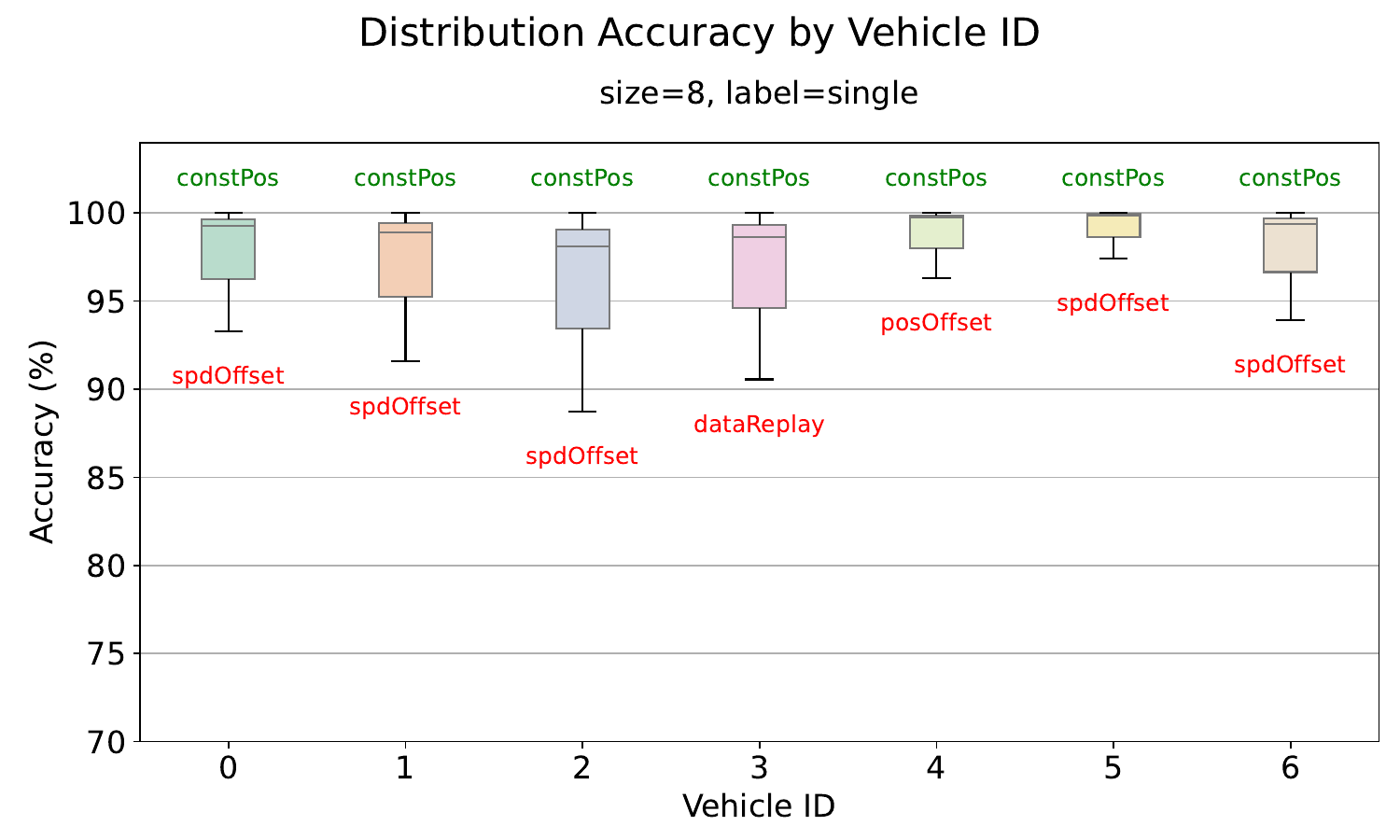}}
    \caption{Prediction accuracy distribution of single label, evaluated by each vehicle ID in a platoon of size 8}
    \label{fig:distr_acc_single_id_8}
\end{figure}

In \cref{fig:av_acc_single_id_4}, it's possible to see the average accuracy in a platoon of size 4, evaluated by each vehicle ID that begins a misbehavior. The average scores are almost the same, close to 100\%, and the confidence interval still reaches a maximum of 100\% for every ID. While in \cref{fig:distr_acc_single_id_4}, it shows the accuracy distribution in a box plot by showing which kind of misbehavior causes the maximum of the interval in green and the minimum in red. It's possible to see how the plots for each ID are almost the same, and the maximums are always caused by constant position, while the minimum is by speed offset. \cref{fig:av_acc_single_id_8} and \cref{fig:distr_acc_single_id_8} show the same accuracy plots, but this time for all the simulations of platoons composed by eight vehicles. It is possible to see how all the averages are still greater than 95\% for all the vehicle's IDs. The maximums of the confidence interval still reach 100\% for all the IDs. Moreover, from the distribution plots, it's possible to see how the maximum is still caused by constant position, while the minimum is from the offsets and data replay, as expected.

\begin{figure}[H]
    \centering
     \makebox[\textwidth][c]{\includegraphics[width=1\linewidth]{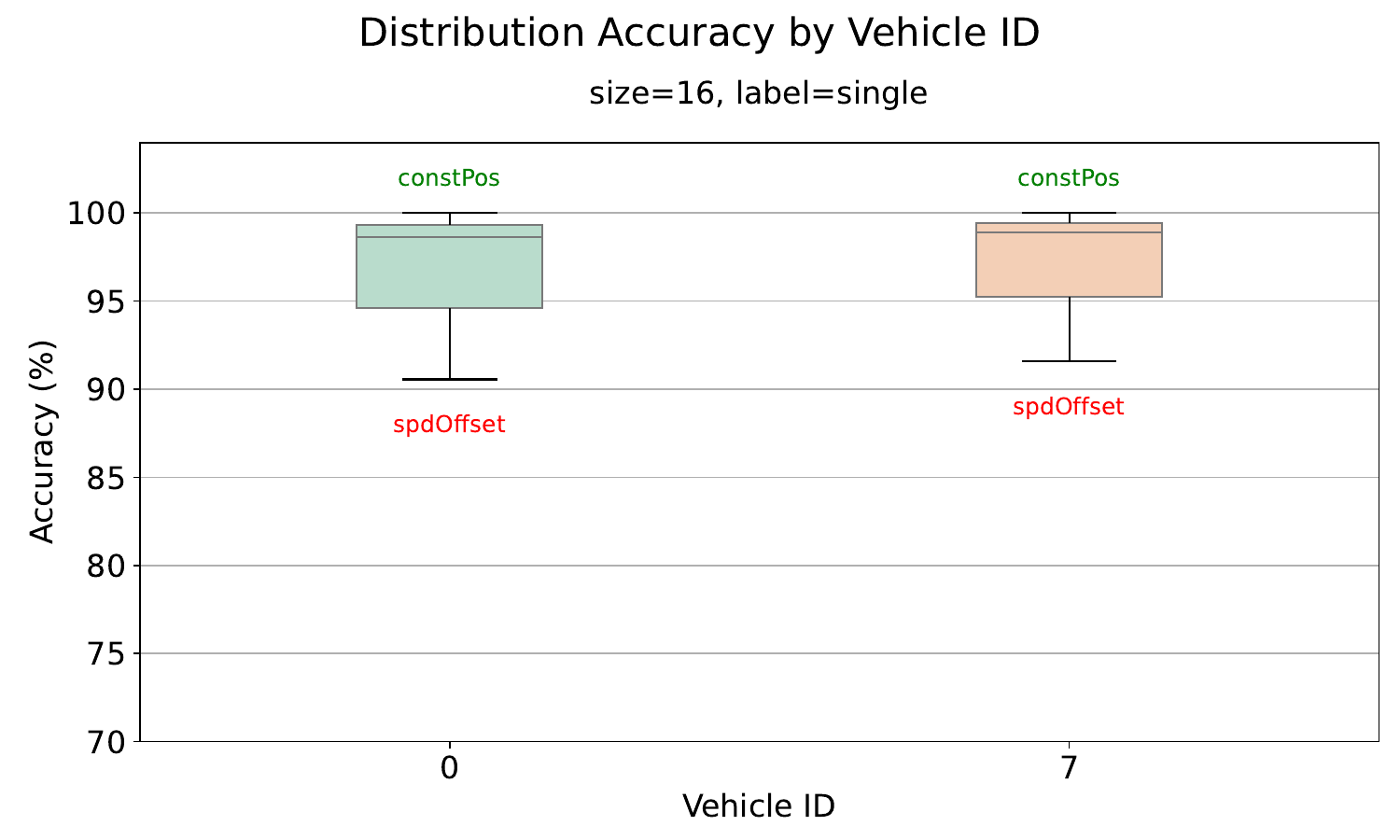}}
    \caption{Prediction accuracy distribution of single label, evaluated by each vehicle ID in a platoon of size 16}
    \label{fig:distr_acc_single_id_16}
\end{figure}

Concerning the platoons of size 16, the distribution of the leader ID and the center vehicle of ID 7 is shown in \cref{fig:distr_acc_single_id_16}, confirming all the considerations of the previous plots. In particular, the average is still greater than 95\% with the maximums that reach 100\%, caused by constant position, and the minimum caused by speed offset. In \cref{tab:mean_single_id} all the punctual values of the accuracy average are resumed for each platoon size.

\begin{table}[H]
\centering
\begin{tabular}{|l|l|l|l|}
\hline
 &
\textbf{Size 4} & 
\textbf{Size 8} & 
\textbf{Size 16}\\ \hline
\textbf{0} & 98.625\% & 99.25\% & 98.625\%\\ \hline
\textbf{1} & 98.625\% & 98.875\% &\\ \hline
\textbf{2} & 98.125\% & 98.125\% &\\ \hline
\textbf{3} & & 98.625\%&\\ \hline
\textbf{4} & & 99.75\%&\\ \hline
\textbf{5} & & 99.875\%&\\ \hline
\textbf{6} & & 99.375\%&\\ \hline
\textbf{7} & & & 98.875\%\\ \hline
\end{tabular}
\caption{Accuracy mean value of the single label grouped by id and for each platoon size simulated}
\label{tab:mean_single_id}
\end{table}

\begin{figure}[H]
    \centering
     \makebox[\textwidth][c]{\includegraphics[width=1\linewidth]{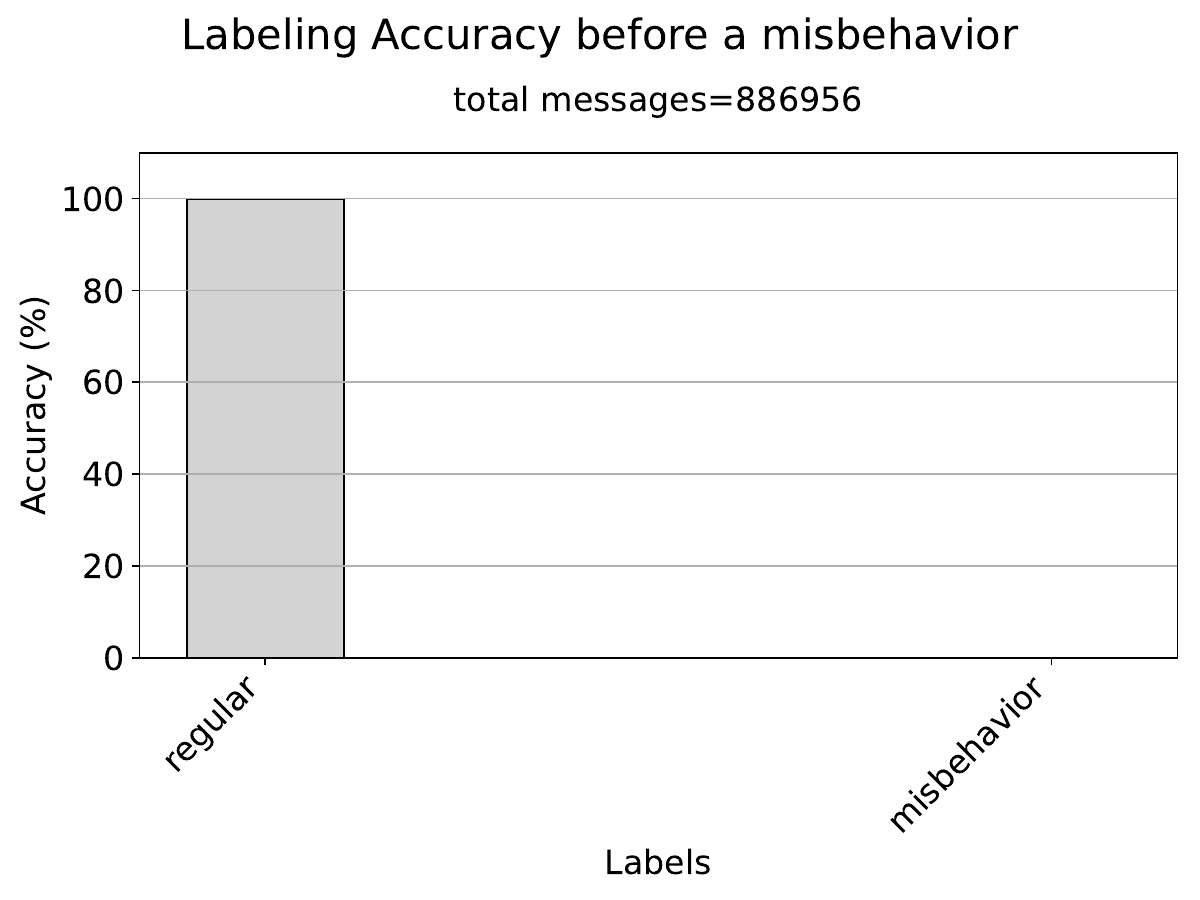}}
    \caption{Labeling accuracy between regular and misbehavior, evaluating how many regular labels are predicted as misbehavior before it begins}
    \label{fig:fp}
\end{figure}

Since the \ac{MDS} seems to work very well on the single label predictions, another important value to be analyzed is the labeling accuracy before a misbehavior starts. In particular, \cref{fig:fp} shows how, on a total of 886956 messages in all the simulations made before the misbehaving time, the regular label is always predicted, indicating how the system is stable and doesn't trigger the protocol when it's not necessary, letting the traffic flow as much as possible. In this case, the labeling choice of doubling the regular label compared to any misbehavior label, presented in \cref{s:training}, seems to bring stability allowing the system not to stumble upon false positives.

\subsection{Multiple label results}
\label{s:res_m}

\begin{figure}[H]
    \centering
    \makebox[\textwidth][c]{\includegraphics[width=1\linewidth]{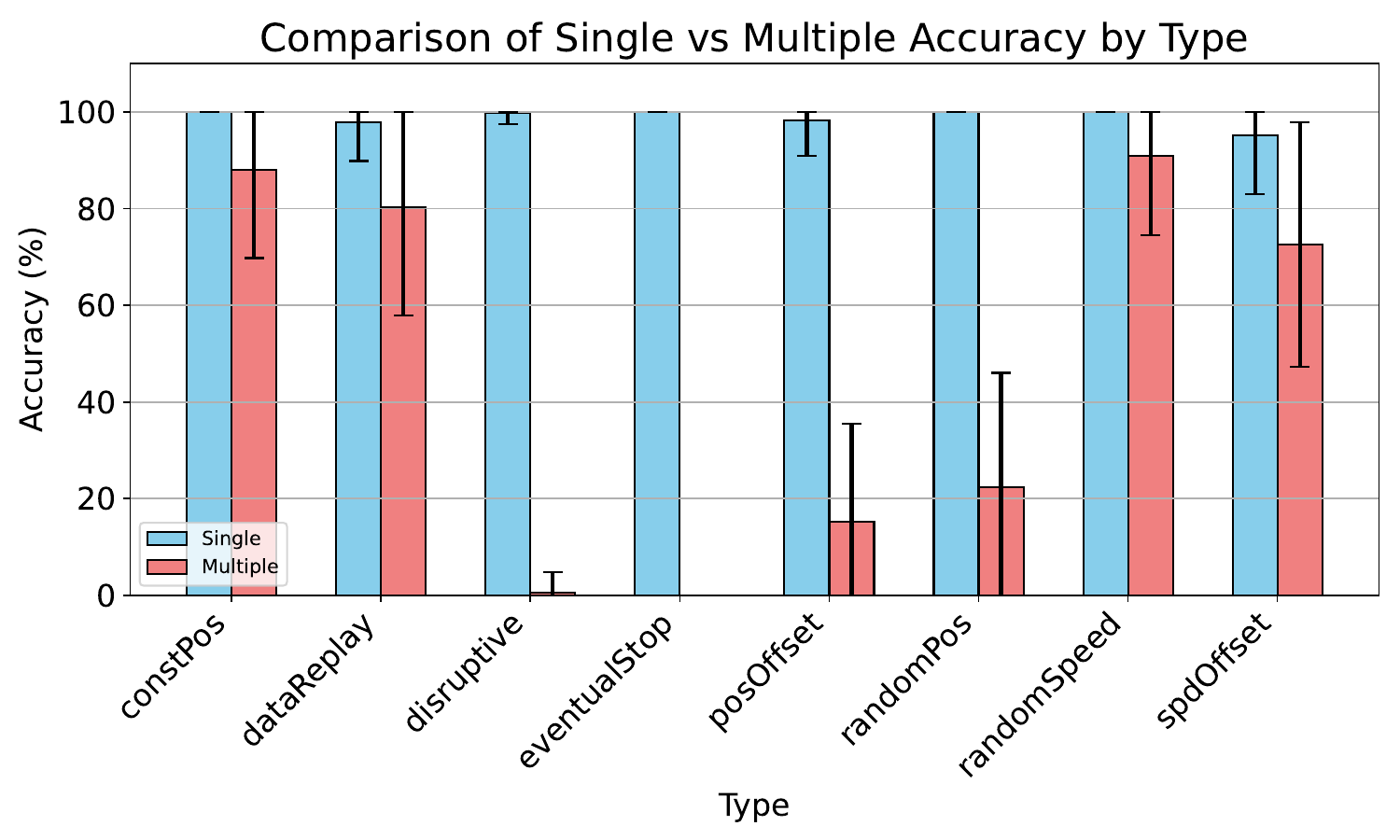}}
    \caption{Prediction accuracy of single label compared with multiple label, with the confidence interval,
    evaluated by each kind of misbehavior}
    \label{fig:av_acc_singlevsmult}
\end{figure}

Without generalizing the unique and single label that includes all the misbehavior, the \ac{MDS} tries to identify each kind of misbehavior from the others. The identification of different misbehaving labels is called multiple label. With good accuracy results on the multiple label, a protocol that makes personalized responses for each misbehavior could be implemented. 
From \cref{fig:av_acc_singlevsmult} it is possible to see immediately that the results do not seem to be good. This figure compares the accuracy levels of the single label, evaluated by each kind of misbehavior, i.e., the same values of \cref{fig:av_acc_single_type}, and the accuracy of multiple label. It's possible to see that the average levels of each misbehavior are very low, and in particular, some of them, like disruptive and eventual stop, are close to zero. 

\begin{figure}[H]
    \centering
    \makebox[\textwidth][c]{\includegraphics[width=1.2\linewidth]{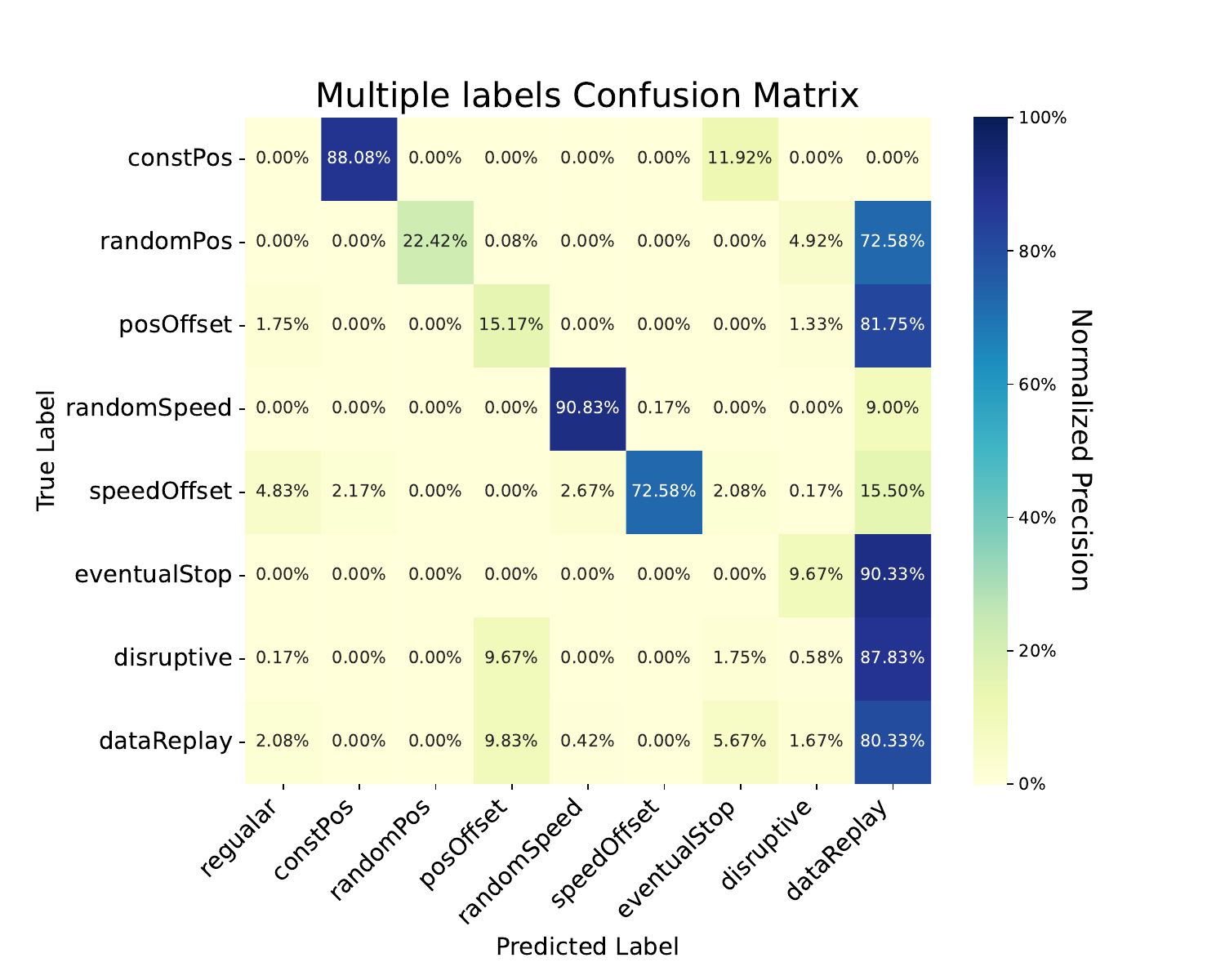}}
    \caption{Confusion matrix of the predictions of the misbehavior on the whole simulation set normalized by row, comparing all the labels }
    \label{fig:cm}
\end{figure}

\noindent In \cref{fig:cm} is presented the confusion matrix of the whole simulation set, as the previous confusion matrix, in \cref{fig:conf-matrix}, this plot
reports a standard confusion matrix with values normalized by the sum of predictions per each category (i.e., per each row), leading to the same metric that we call "Normalized Precision". On this matrix, the main diagonal has to be with the highest value to indicate that the true labels on the y-axis are correctly predicted on the x-axis. The resulting matrix of this simulation shows that the diagonal has good values only for constant position, random speed, and speed offset. In contrast, most of the time, the other values are generally predicted as data replay, with very low values for disruptive and eventual stop. Moreover, the regular label, which has to be zero for every cell because the matrix evaluates results only on misbehavior scenarios, at the misbehaving time, is predicted in some cases for position offset, speed offset and data replay, confirming what is presented in \cref{fig:av_acc_single_type}, that shows that for the single label, those three are the ones with the most significant problems. So, from this matrix, it emerges that data replay creates a lot of noise in the simulation environment, meaning that it could have been replicated differently than the original one, or it could be similar to other misbehavior from the data produced. Furthermore, it could also be the most problematic also for the single label protocol. It has to be noticed that also in the presentation paper of the extended \ac{VeReMi} dataset \cite{kamel2020veremi}, the replay attacks are indicated as the most problematic, especially for the precision.

\begin{figure}[H]
    \centering
    \makebox[\textwidth][c]{\includegraphics[width=1\linewidth]{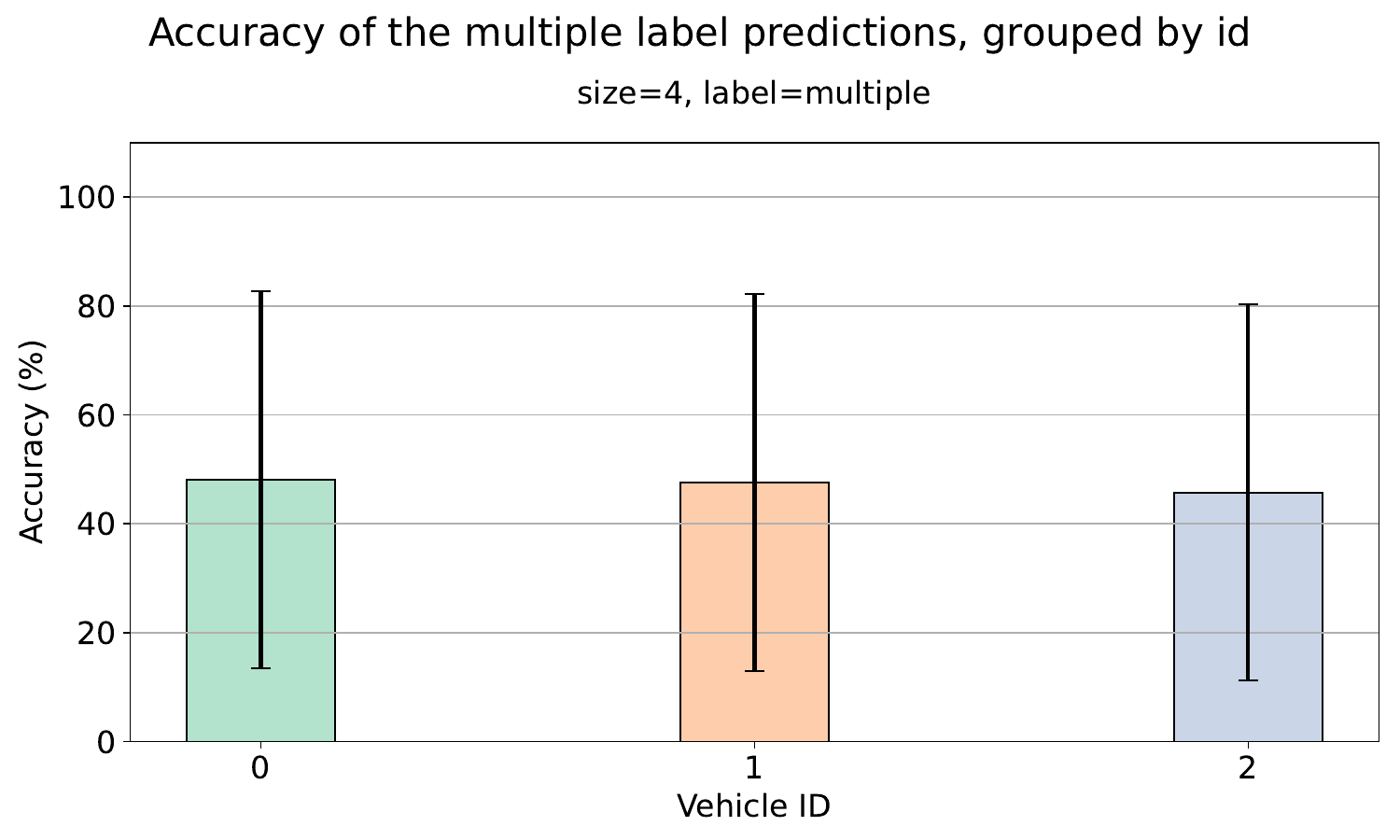}}
    \caption{Prediction accuracy of multiple label with the confidence interval,
    evaluated by each vehicle ID in a platoon of size 4}
    \label{fig:av_acc_mult_id_4}
\end{figure}

\begin{figure}[H]
    \centering
     \makebox[\textwidth][c]{\includegraphics[width=1\linewidth]{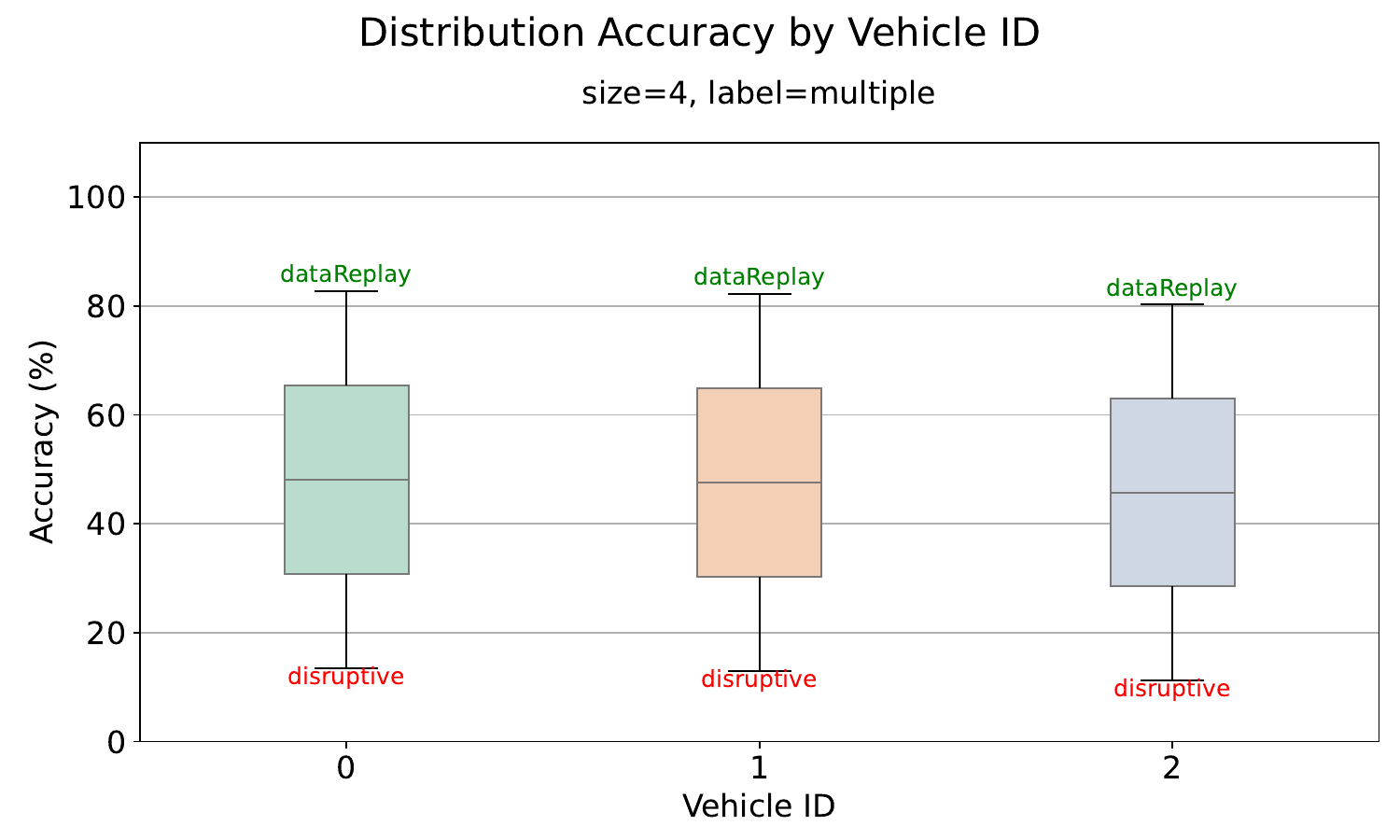}}
    \caption{Prediction accuracy distribution of multiple label, evaluated by each
    vehicle ID in a platoon of size 4}
    \label{fig:distr_acc_mult_id_4}
\end{figure}

\begin{figure}[H]
    \centering
    \makebox[\textwidth][c]{\includegraphics[width=1\linewidth]{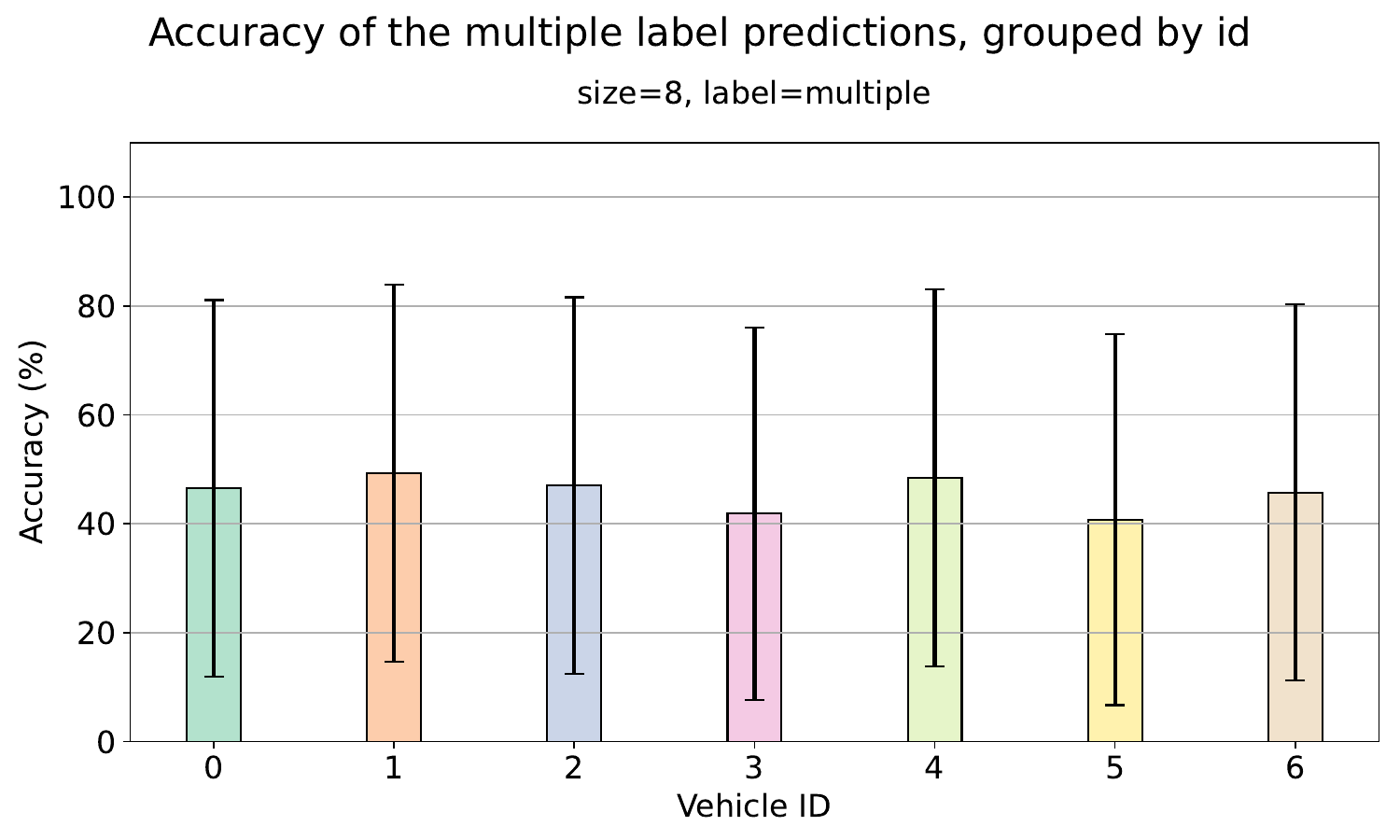}}
    \caption{Prediction accuracy of multiple label with the confidence interval,
    evaluated by each vehicle ID in a platoon of size 8}
    \label{fig:av_acc_mult_id_8}
\end{figure}

\begin{figure}[H]
    \centering
    \makebox[\textwidth][c]{\includegraphics[width=1\linewidth]{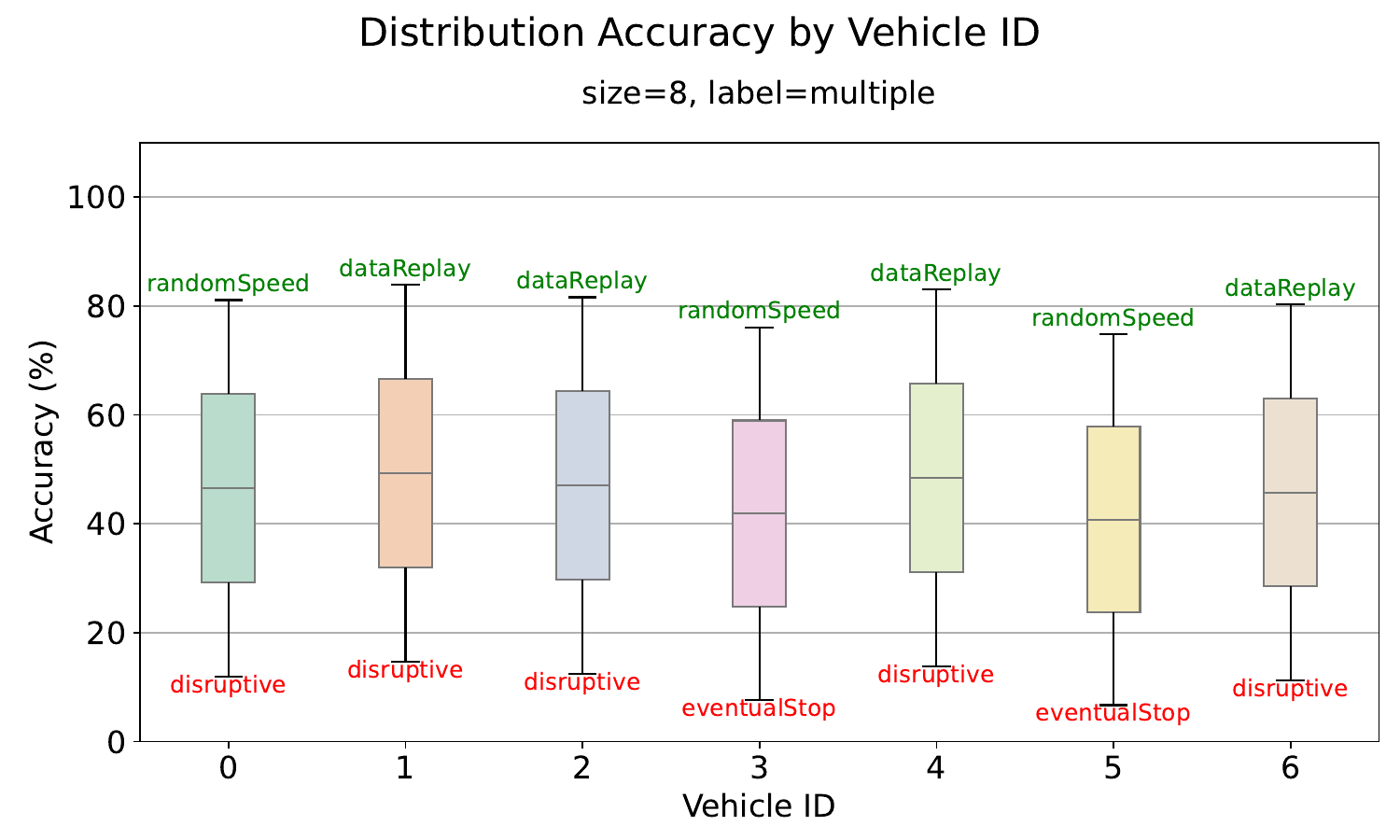}}
    \caption{Prediction accuracy distribution of multiple label, evaluated by each
    vehicle ID in a platoon of size 8}
    \label{fig:distr_acc_mult_id_8}
\end{figure}

\begin{figure}[H]
    \centering
    \makebox[\textwidth][c]{\includegraphics[width=1\linewidth]{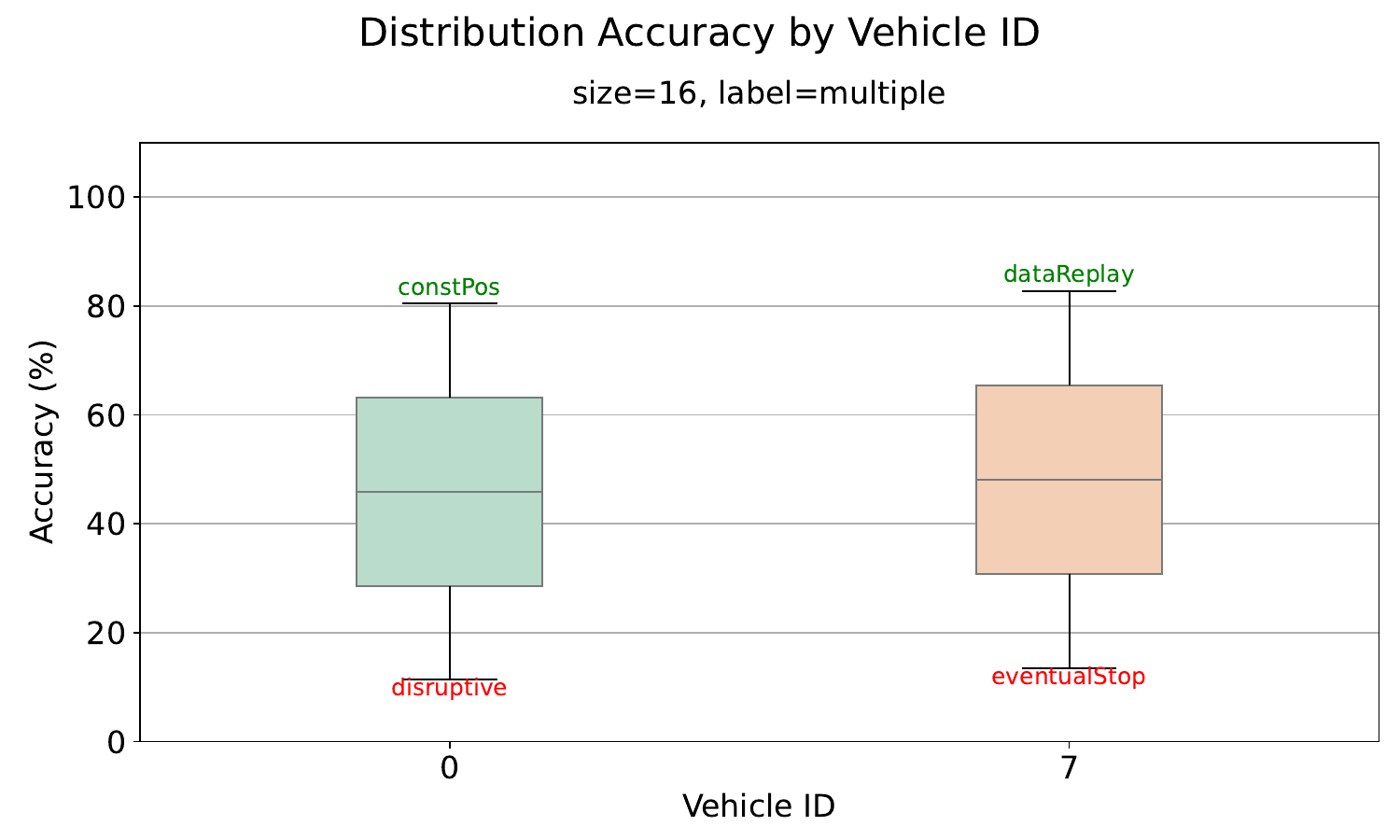}}
    \caption{Prediction accuracy distribution of multiple label, evaluated by each
    vehicle ID in a platoon of size 16}
    \label{fig:distr_acc_mult_id_16}
\end{figure}

In \cref{fig:av_acc_mult_id_4} and \cref{fig:distr_acc_mult_id_4} are shown the average accuracy and the accuracy distribution with the maximum and minimum misbehavior type that causes that value, on a platoon of 4 vehicles, for each ID that begins the misbehavior, while in \cref{fig:av_acc_mult_id_8} and \cref{fig:distr_acc_mult_id_8} the same results are shown for a platoon of size 8. In the end, \cref{fig:distr_acc_mult_id_16} shows the accuracy distribution of a platoon size 16.

The results show average values always around 50\% with distribution values that change between almost 10\% and 80\%. This results' patterns are present for each platoon size and for each vehicle ID that begins a misbehavior, meaning that the position in a platoon is not characterizing for the multiple label detection. Moreover, the \ac{MDS} doesn't work well on the multiple label predictions. This result could be obtained because of the different representation of the misbehavior during the simulations, but also from different data produced from the highway scenario, differently from the urban one. In particular, analyzing the offline performances, in \cref{s:perform}, it is possible to notice that on the offline validation set the scores are good, meaning one more time that the translation at run-time caused some punctual detection problems.

\subsection{Accidents reduction}
\label{s:gain_acc}

Since the single label results, in \cref{s:res_s}, are good, they provide a solid basis for the defense protocol. The purpose of the protocol is save as much vehicles as possible in a platoon. So, the following results present the accidents evolution in a platoon, by activating the defense protocol, based on the single label predictions of the \ac{MDS}.

\begin{figure}[H]
    \centering
    \makebox[\textwidth][c]{\includegraphics[width=1\linewidth]{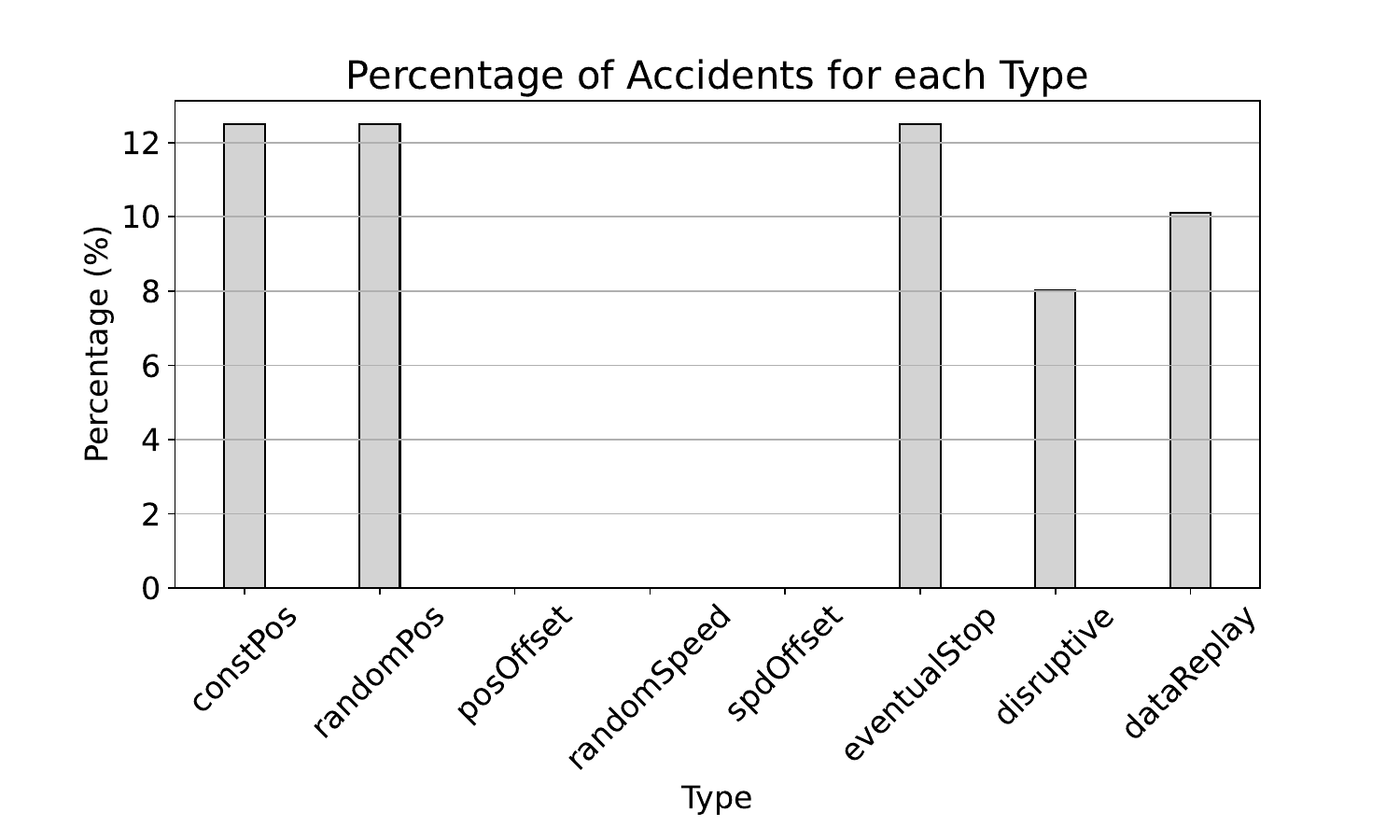}}
    \caption{Fraction of simulations that ended in a crash without the defense protocol. One bar is reported for each kind of simulated attack}
    \label{fig:perc_acc_bytype}
\end{figure}

\begin{figure}[H]
    \centering
    \makebox[\textwidth][c]{\includegraphics[width=1\linewidth]{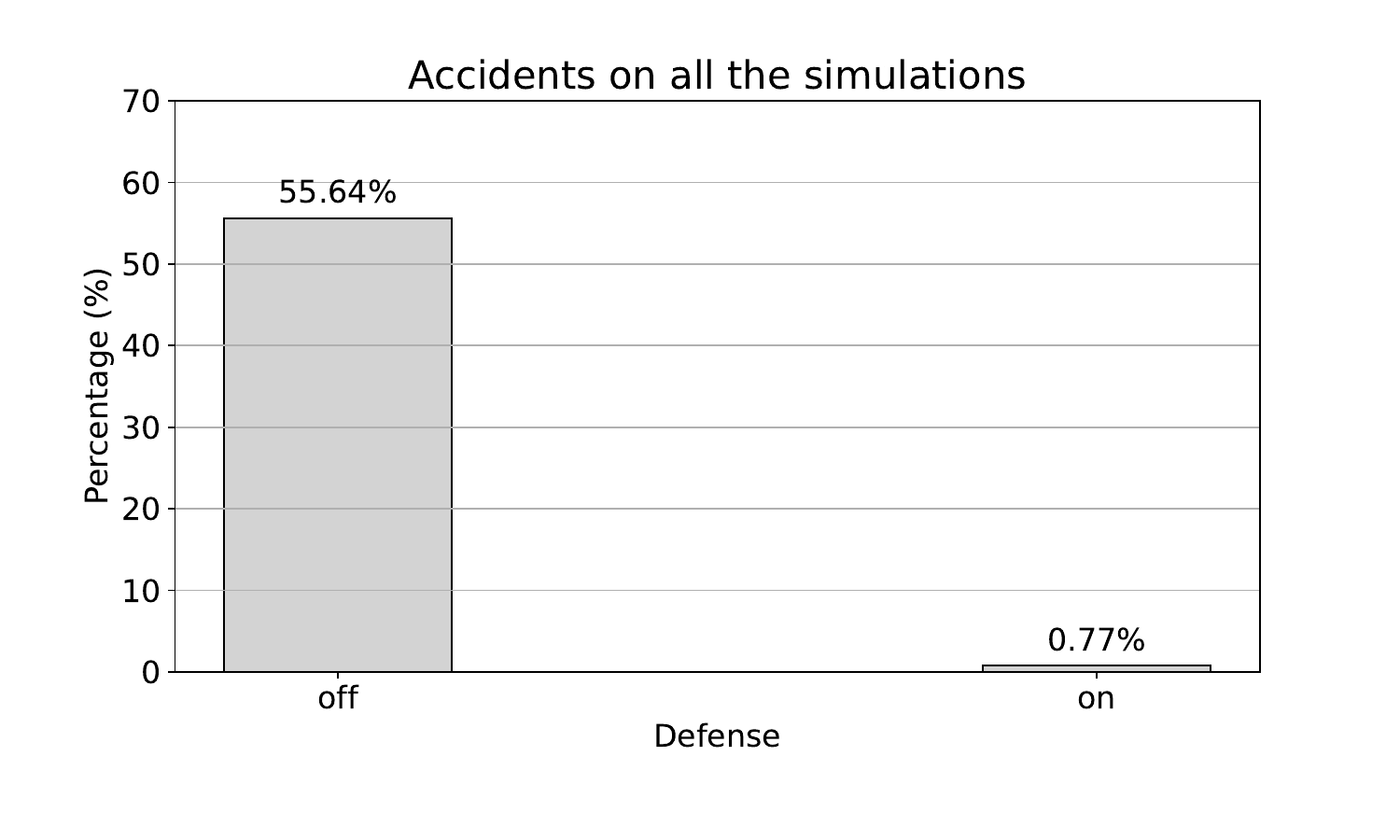}}
    \caption{Percentage of accidents on all the simulations without and with the defense protocol}
    \label{fig:gain_acc}
\end{figure}

\begin{figure}[H]
    \centering
    \makebox[\textwidth][c]{\includegraphics[width=1\linewidth]{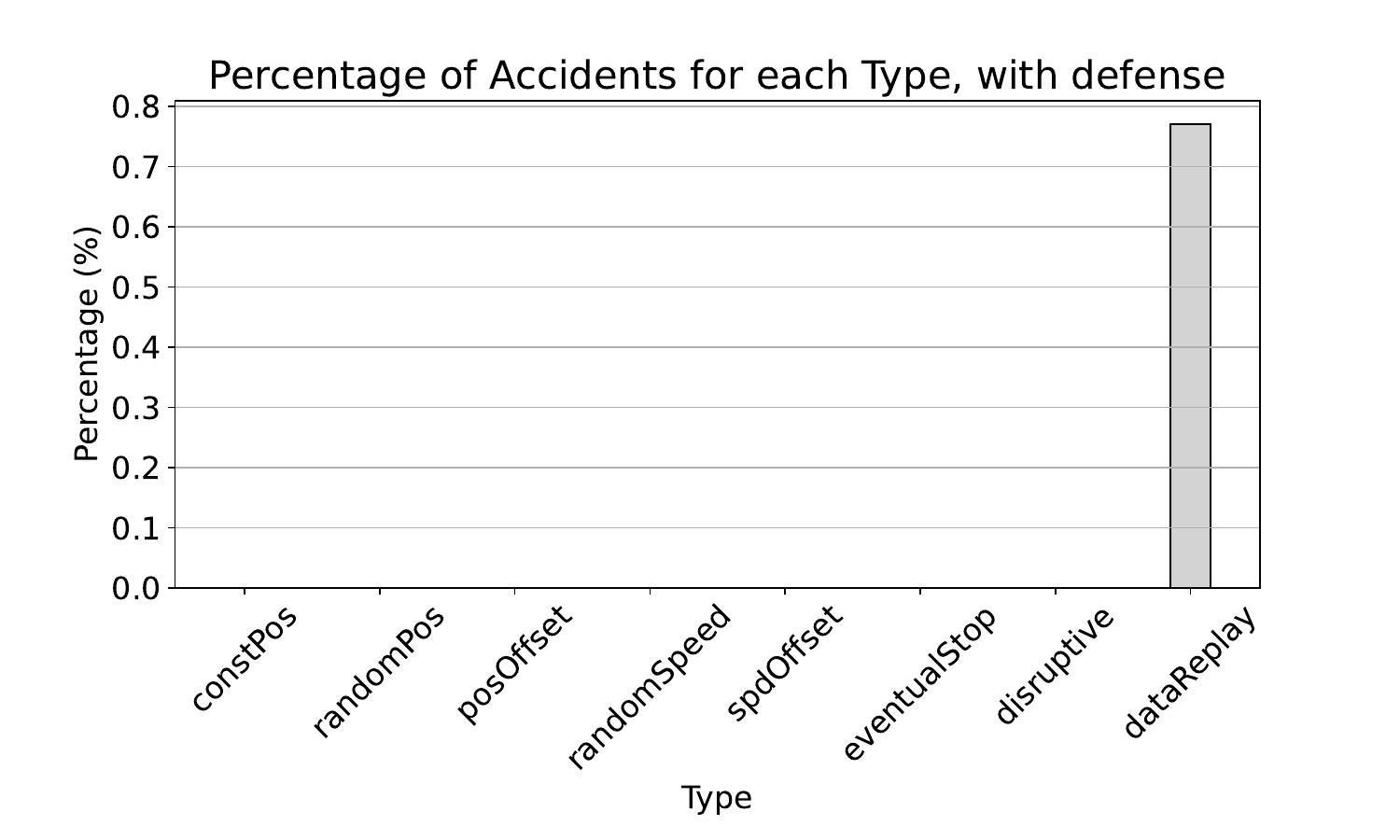}}
    \caption{Fraction of simulations that, despite the active defense protocol, ended in a crash. One bar is reported for each kind of simulated attack}
    \label{fig:perc_acc_def_bytype}
\end{figure}

First of all, in \cref{fig:perc_acc_bytype} it is possible to see the percentages of accidents caused by each misbehavior without the defense being active. In particular, the position and speed offset never cause accidents, meaning that the lower accuracy values of \cref{fig:av_acc_single_type} are not a problem. Also random speed never cause any accident, also if the random speed range is very wide. This means that on a control system like Ploeg a sudden and wide change in the positions transmitted is more significant than a change in the actual speed transmitted. Furthermore, data replay, that is one of the most problematic misbehavior from a detection point of view, cause almost the 10\% of the accidents through all the simulation set without defense. In \cref{fig:gain_acc} is shown how introducing the defense saved many platoons, in particular with a gain in saved platoons of 98.62\%, calculated as $gain = \frac{defense_{off} - defense_{on}}{defense_{off}}$. In the end, \cref{fig:perc_acc_def_bytype} shows how the 0.77\% of remaining accidents with the defense protocol active it's caused by data replay, as expected, since it is the only problematic misbehavior that is sometimes not detected from the \ac{MDS}.

    \chapter{Conclusions}
\label{s:conc}

Through extensive analysis and simulations, the study successfully addressed the creation of a simple \ac{ML} framework, based on \ac{LSTM}, to detect eight different misbehavior taken from the \ac{VeReMi} dataset, a standard in the misbehavior detection field in \ac{VN}. Moreover, the study reached the goal of translating the \ac{MDS} trained offline, in run-time simulations, validated with a platoon simulated on a highway, a profoundly different scenario compared to the training dataset. In the end, this thesis aimed the design of a defense protocol whose purpose is to save the vehicles in a platoon in which a misbehavior, detected by the \ac{MDS}, is happening.

From the results in \cref{s:res_s}, the prediction accuracy in identifying all the anomalies and attacks studied as a unique misbehavior label, i.e., defined as single label through the thesis, is well working. This result is meaningful since it means that training offline a simple \ac{NN} on the \ac{VeReMi} dataset is sufficient to recognize the same set of misbehavior reconstructed in a slightly different way at run-time on every vehicle and in a different scenario. Furthermore, this result provides a solid basis for the defense protocol that, as shown in \cref{s:gain_acc}, has an efficiency close to 99\% in saving accidents all over the simulation set. Although the protocol is built to give the same response to all the misbehavior detected, i.e., the disruption of the platoon, it enhances driving safety without interrupting the cooperative driving when it's not necessary. 

Despite the protocol works on a single misbehavior label, a deep analysis of the \ac{MDS} that tries to identify each specific misbehavior was made in \cref{s:res_m}. Those multiple labels have bad results even though the good metrics value was reached while training offline the model. This result could be easily explained. First of all, the eight anomalies were reconstructed based on the idea of the dataset's misbehavior, but it ended in implementing those misbehavior differently. In particular, the replay attacks are the most different recreated, and in fact, they create noise in both the single and multiple label predictions. Secondly, the data produced through the simulations are different from the urban scenario of the City of Luxembourg. These arguments lead to thinking that, even with a more complex \ac{NN}, an \ac{MDS} that tries to identify every misbehavior could be useless since, on the road, any vehicle can start to misbehave in slightly different ways from the ones learned from the detector.
In the end, the solution adopted by using a defense protocol that dismantles the platoon based on an \ac{MDS} trained offline on the \ac{VeReMi} dataset reaches the goal to enhance driving safety.

\section{Future work}
This study provided a solid basis for exploring the \acp{MDS} in a run-time environment, showing that even if the training is made offline and loaded in a vehicle, the framework can work correctly also under different traffic conditions and with misbehavior reconstructed in different ways.
This achievement opens the doors to future studies. Some potential research areas involve introducing more attacks in the training phase, making the model more complex and efficient while still keeping the required simplicity to be used on board. Also, the designed simple protocol could be made more complex. As presented in the conclusions, we think that building a protocol with a personalized response for each misbehavior is not possible, considering that it is not true that all types of misbehavior are known; for this reason also, a defense protocol like the one presented could be sufficient for the platooning application.

It has to be said that training the model on those simulated attacks and with few data is not sufficient to introduce the \acp{MDS} in the automotive industry. Those systems require millions of data that have to be as much realistic as possible, to achieve the required accuracy to install the \ac{MDS} on a real \ac{OBU} into a vehicle. Consequently, the most interesting branch of studies involves the expansion of the standard \ac{VeReMi} dataset in order to contain millions of data provided from different sources and possibly also from a real-world collection.
But in a future scenario in which the \ac{NN} could be built on a reasonable amount of realistic data, the \ac{MDS} along with the protocol could be introduced in the real world as a new \ac{CITS} technology.

\section*{Acknowledgments}
	The work of this thesis has been carried out within the framework of PNRR-funded research projects granted to the Department of Information Engineering at the University of Brescia.
	In particular, the topics of Cooperative Driving, Cooperative Perception and Secure Vehicular Networks are core of these two projects: 
	\begin{itemize}
		\item National Recovery and Resilience Plan (NRRP), project ``Sustainable Mobility Center (MOST)'', 2022-2026, CUP D83C22000690001, Spoke N$^o$ 7, ``CCAM, Connected networks and Smart Infrastructures;''
		\item  \textit{Parternariati Estesi} program ``Security and Rights in CyberSpace (SERICS),'' (PE00000014), Spoke 7, Project SCAR, Cascade Call ``SCAR: a Privacy enHAnced SEcurity framework (SCARPHASE)'' (CUP C89J24000580008).
	\end{itemize}

 \begin{figure}[!b]
    \vspace{-1cm}
    \hfill \includegraphics[width=0.4\linewidth]{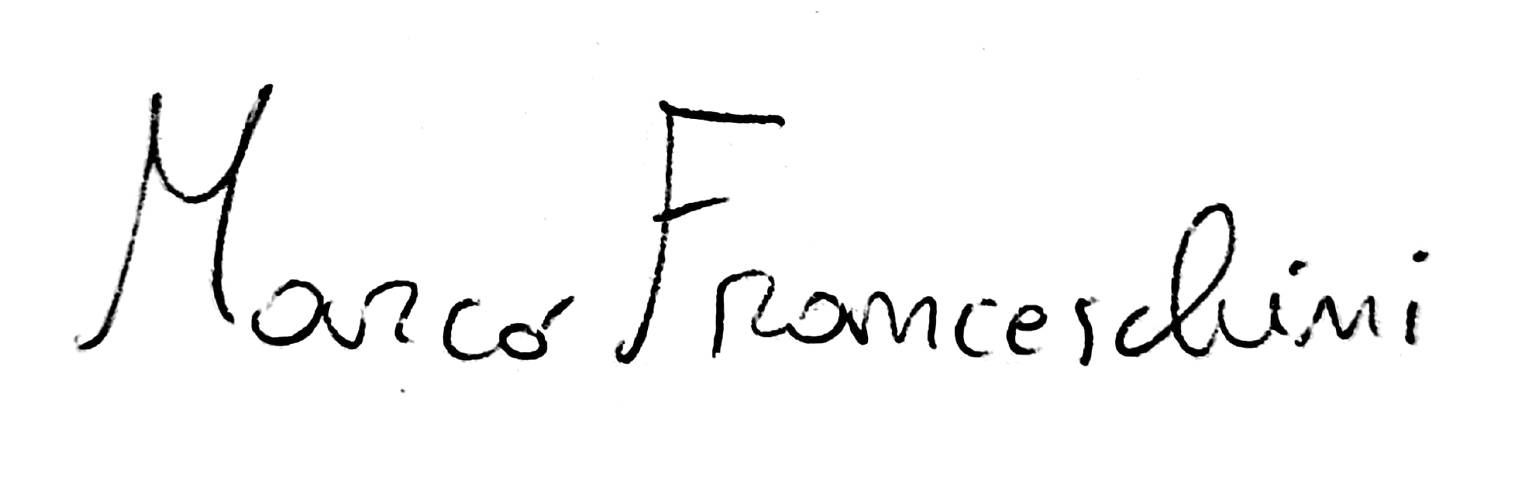}
 \end{figure}

\printbibliography

\newpage

\section*{Ringraziamenti}
Dopo cinque anni il percorso giunge al termine, dopo tante sfide e sacrifici, vorrei ringraziare tutte le persone che hanno caratterizzato questo viaggio speciale. 
Ci terrei a non fare nomi, ma a raccontare episodi perché chiunque di voi mi stia ascoltando o leggendo ha condiviso con me dei momenti speciali, e vorrei che potesse essere in grado di ritrovarsi nelle mie parole. E poi si sa, i ringraziamenti sono sempre un momento delicato, vorrei provare a non dimenticarmi nessuno, perché chiunque sia passato dalla mia vita in questi cinque anni a modo suo me l'ha cambiata.

Vorrei cominciare con qualche piccola eccezione. Ovviamente sto parlando di mamma e papà; non posso neanche quantificare il ringraziamento che dovrei farvi, voi due avete fatto tutti i sacrifici possibili per me, non mi avete mai fatto mancare niente. In questi cinque anni ho sempre ricevuto il supporto necessario nei momenti delicati, i tanti momenti delicati, perché alla fine una casa per stare in piedi ha bisogno di solide fondamenta, voi siete stati le mie. Senza di voi non so se sarei mai arrivato dove sono ora, quindi posso solo dirvi: Grazie.
Oltre a mamma e papà vorrei ringraziare tutta la mia famiglia, a chi è sempre stato disponibile per un passaggio in aeroporto, per una risata a tavola, o per il caffè della domenica mattina, grazie.

Un'altra doverosa eccezione va fatta per chi da ormai quasi tre anni ha creduto in me, affiancandomi sia nella tesi triennale che in quella magistrale e tanti altri lavori negli anni. Vorrei ringraziare il professor Lo Cigno e Lorenzo per avermi fatto crescere al loro fianco in questo percorso.

I momenti seri sono finiti, o quasi, ve lo prometto. E vorrei partire ringraziando i miei amici, e con amici intendo quelli del sabato sera, quelli del sushi organizzato all'ultimo rischiando di pagare tutto, quelli della caviglia slogata nel vialetto di casa, gli amici tra cui non ci ricordiamo mai un compleanno ma alla fine siamo sempre in prima linea alle feste. Quelli che non sai cosa fanno nella vita per mesi, ma poi a tutte le grigliate li rivedi e ci parli per ore. O ancora, a chi tra di loro conosco da quando sono nato, che con poco affetto non mi ha mai fatto una telefonata in sei mesi fuori dall'Italia, ma da un giorno all'altro ha deciso di venire  a trovarmi direttamente. Agli amici con cui ho condiviso i viaggi più belli della mia vita, tra un bicchiere di vino estremamente costoso, senza saperlo, e una partenza senza conoscere la destinazione. Grazie!

Non posso non ringraziare poi tutte le persone che in questi anni sono state parte della mia più grande passione, la musica. E allora vorrei iniziare ringraziando le persone con cui ho iniziato a suonare seriamente, con chi seppur più grande di me ho instaurato un'amicizia forte, facendomi scoprire dopo anni di essere daltonico, oppure a tutti quelli che erano presenti nel momento in cui un pazzo è salito sul palco a bersi una bottiglia di Jack Daniels a goccia, o ancora, a chi era al mio fianco a ridere nel momento in cui una persona mascherata scelse di ribaltare uno spettatore dal pubblico sulla cassa spia del palco. 
Venendo a momenti più recenti, vorrei ringraziare quell'insieme di musicisti per cui dopo il concerto bisogna bere almeno cinque birre, quelli pronti a tutto, anche a sei ore e mezza di macchina per andare fino ad Imola, ma che palco era! Infine vorrei ringraziare la mia seconda famiglia degli ultimi tre anni, coloro con cui ho condiviso forse più ore su un furgone che sul palco, con cui sono cresciuto e da ragazzino sono diventato adulto, suonando in posti che mai mi sarei nemmeno immaginato. Grazie!

Vorrei ringraziare anche tutte le persone che ho conosciuto grazie all'universi-\\tà, tutti quanti diventati grandi amici. A chi mi ha convinto a partire per sei mesi, regalandomi probabilmente una delle esperienze più belle della mia vita. Oppure a chi in quell'esperienza mi è venuto direttamente a trovare, devo confessarvelo, non me lo sarei mai aspettato. Alla compagnia di tutti i giorni, quella delle lezioni e delle piadine a pranzo, quella delle domenica sera al Retroscena. Grazie!

Infine vorrei ringraziare tutte le persone conosciute in Portogallo, a chi è stato la mia famiglia per sei mesi, chi ha fatto la spesa con me per tutti i giorni, chi si è preso cura di me nel momento del bisogno, con chi ho condiviso i viaggi più belli della mia vita, l'alba delle Azzorre, le onde di Nazaré e i tanti tramonti di Porto. Grazie anche all'enorme gruppo di amici che ho conosciuto, come dimenticare la "gita scolastica" ad Aveiro. A chi di loro è stato mio compagno di classe, chi mi ha insegnato la propria cultura, la propria lingua (non ci avrei mai sperato) o anche solo fatto assaggiare un piatto tipico. Potrei andare avanti per ore a elencare le emozioni e i ricordi vissuti in quei sei mesi. Grazie!

\newpage
ENG: I would like to thank all the people I met in Portugal, the ones who became my family for six months, those who shopped with me every day, those who took care of me when I needed it, and those with whom I shared the most beautiful trips of my life: the sunrise in the Azores, the waves of Nazaré, and the sunsets in Porto. I'm also grateful to the amazing group of friends I met, especially those who were my classmates, taught me their culture and language (something I never expected), or even just shared a typical dish with me. I could go on for hours listing the emotions and memories from those six months. Thank you!

PT: Obrigado ao enorme grupo de amigos e pessoas que conheci em seis meses no Portugal. Às quem me ensinaram a sua cultura, a sua língua (nunca teria esperado isso) ou até mesmo me fizeram provar um prato típico. Eu poderia continuar por horas listando as emoções e lembranças que experimentei nesses seis meses. Obrigado!

\end{document}